\documentclass[pdflatex,sn-mathphys-num]{sn-jnl}

\usepackage{graphicx}%
\usepackage{multirow}%
\usepackage{amsmath,amssymb,amsfonts}%
\usepackage{amsthm}%
\usepackage{mathrsfs}%
\usepackage[title]{appendix}%
\usepackage{xcolor}%
\usepackage{textcomp}%
\usepackage{manyfoot}%
\usepackage{booktabs}%
\usepackage{algorithm}%
\usepackage{algorithmicx}%
\usepackage{algpseudocode}%
\usepackage{listings}%

\usepackage{array}%
\usepackage{tabularx}%
\usepackage{pifont}%
\usepackage{bm}%
\usepackage{placeins}%

\newcommand{\methodname}{MS-ECG-FM}

\newcommand{\cmark}{\textcolor{green!60!black}{\ding{51}}}
\newcommand{\xmark}{\textcolor{red!70!black}{\ding{55}}}
\newcommand{\pmark}{\textcolor{orange!85!black}{\raisebox{0.2ex}{$\bm{\sim}$}}}
\newcommand{\StrongEvidence}{\ding{72}}      
\newcommand{\PromisingEvidence}{\ding{115}}  
\newcommand{\ExploratoryEvidence}{\ding{109}}

\theoremstyle{thmstyleone}%
\theoremstyle{thmstyletwo}%

\theoremstyle{thmstylethree}%
\usepackage[capitalize,noabbrev]{cleveref}%
\crefname{appendix}{Supplementary Note}{Supplementary Notes}
\Crefname{appendix}{Supplementary Note}{Supplementary Notes}
\renewcommand{\appendixname}{Supplementary Note}

\begin{document}

\title[\methodname]{\methodname{}: Towards a More Universal Electrocardiogram Foundation Model for Health Monitoring using Multi-source Contrastive Learning}

\author*[1]{\fnm{Robert} \sur{Lewis}}\email{roblewis@mit.edu}

\author[2]{\fnm{I-Min} \sur{Chiu}}

\author[2]{\fnm{Kyle} \sur{Verrier}}

\author[2]{\fnm{Karthik} \sur{Jayaraman Raghuram}}

\author[2,3]{\fnm{Francoise} \sur{Marvel}}

\author[2]{\fnm{Salar} \sur{Abbaspourazad}}

\author[2]{\fnm{Anshuman} \sur{Mishra}}

\author[2,4]{\fnm{Guillermo} \sur{Sapiro}}

\author*[2]{\fnm{Andrew C.} \sur{Miller}}\email{acmiller@apple.com}
\equalcont{These authors contributed equally to this work.}

\author[2]{\fnm{Joseph} \sur{Futoma}}
\equalcont{These authors contributed equally to this work.}

\affil*[1]{\orgname{Massachusetts Institute of Technology} (work done while at Apple)}

\affil[2]{\orgname{Apple, Inc}}

\affil[3]{\orgdiv{Division of Cardiology}, \orgname{Johns Hopkins Medicine}}

\affil[4]{\orgname{Princeton University}}

\abstract{Electrocardiography (ECG) records the electrical activity of the heart, aiding diagnosis by detecting abnormalities in cardiac function. ECG foundation models have demonstrated promising results, but are limited by a reliance on ECG interpretation reports as their sole supervision. Because interpretation reports only capture the subset of waveform information routinely recognized by clinicians, this constrains representation learning to overlook the broader diagnostic signals present in ECG. We introduce a new ECG foundation model --- \methodname{} --- that is trained through contrastive alignment to multiple distinct clinical note types, including ECG, echocardiography, radiology, and discharge reports. We evaluate \methodname{} on an extended set of ECG detection benchmarks, showing that it comprehensively outperforms existing methods on the full span of conditions that ECG can detect, including in reduced-lead configurations. Different reports improve representations for different diagnostic domains, while multi-source alignment captures their complementary information and produces consistently strong representations across clinically diverse tasks.}


\maketitle

\section{Introduction}\label{sec:intro}

Electrocardiography (ECG) is a low-cost technology that captures high-dimensional signatures of cardiac electrical activity. Performed millions of times daily, the standard 12-lead ECG serves as the primary instrument for the initial clinical assessment of cardiac function. Recent data curation efforts have made large-scale, well-characterized clinical ECG datasets available to the academic community (e.g., MIMIC-IV-ECG \cite{gow2023mimic}). This has facilitated new research on improving diagnostic technologies using machine learning (ML) and artificial intelligence (AI) to learn from ECG data at scale \cite{wu2025deep}. 

Deep learning is highly effective on ECG waveforms given the high-dimensional nature of the signal \cite{wu2025deep}. Furthermore, given the sparsely or weakly-labeled nature of many ECG datasets (i.e., many ECG waveforms are not annotated or are only paired with weaker forms of supervision such as unparsed text reports), self-supervised learning (SSL) is the leading deep learning paradigm given its ability to bypass these challenges. In SSL, ECG representations are first \emph{pre-trained} using contrastive or reconstruction-based objectives on large datasets, before transferring the representations (or embeddings) learned in pre-training to \emph{downstream} tasks \cite{kiyasseh2021clocs,MEHARI2022105114,Zhang2023,na2024guiding,tian2024foundation,liu2024zero}. 

Collectively, these data and methods mark the advent of the foundation model era of ECG modeling (ECG-FM), where success is defined by the creation of \emph{universal} ECG representations: in other words, a comprehensive, information-rich way to encode the ECG signal that can be easily adapted to a broad range of downstream tasks such as condition-specific classifiers. Multimodal contrastive ECG-to-text ECG-FMs, where ECG waveforms are aligned to paired ECG machine text reports, currently produce state-of-the-art (SOTA) ECG representations for a number of downstream tasks \cite{hung2025boosting,wang2025from}. Multitask pre-training --- where task labels are derived from machine reports --- has also demonstrated SOTA performance \cite{li_ecgfounder2025}. We survey contemporary ECG-FMs in more detail in \cref{tab:review_ecg_fms}. 

However, progress towards a truly universal ECG-FM has so far been limited by the exclusive reliance of contemporary methods on machine-generated ECG interpretation reports as their sole source of supervision \cite{liu2024zero,hung2025boosting,wang2025from,li_ecgfounder2025}. Historically, these reports have been generated by rules-based systems and they only contain a limited set of statements related to the morphological features of the waveform and the possibility of certain conditions (e.g., rhythm, conduction abnormalities, or ST-segment changes) \cite{Schlapfer2017}. By restricting pre-training to only these reports, there is no incentive for the ECG-FM to learn features that are discriminative of conditions beyond those described within them. More formally, this can be understood from an information theory perspective: contrastive objectives (e.g., InfoNCE) maximize a lower bound on the mutual information between views, guiding the encoder to learn only the features that represent this \cite{oord2018representation,Liang2023}. When this shared information is also discriminative of a downstream outcome, the ECG-FM's representations will transfer well. However, when it is not, the representations can become invariant to task-relevant factors of variation \cite{tian2020makes}, resulting in poor transfer performance on certain tasks (e.g., comparable to using representations from a randomly initialized encoder that has not been trained \cite{berger2026position}). 

Contrary to the narrow information content of ECG interpretation reports, recent research has demonstrated the utility of ECG to detect a far broader set of conditions \cite{attia2021application,siontis2021artificial}. For example, it has been shown that structural heart disease endpoints such as low left ventricular ejection fraction (LVEF) or aortic stenosis (AS) --- which are typically diagnosed with an ECHO --- can also be detected with ECG \cite{yao2021artificial}. Furthermore, emerging evidence shows other non-cardiac diseases such as chronic kidney disease (CKD) and clinically relevant CKD states such as hyperkalemia can also be detected from ECG \cite{hyperkalemia2019,holmstrom2023deep}, as well as pulmonary conditions and obstructive sleep apnea (OSA) \cite{tison2019automated,Wysokinski2024,covassin2025deep}. We synthesize the full range of conditions and outcomes where there is at least exploratory evidence of ECG detection in \cref{tab:review_ecg_outcomes}. 

\begin{figure*}[t]
    \centering
    \includegraphics[width=\textwidth]{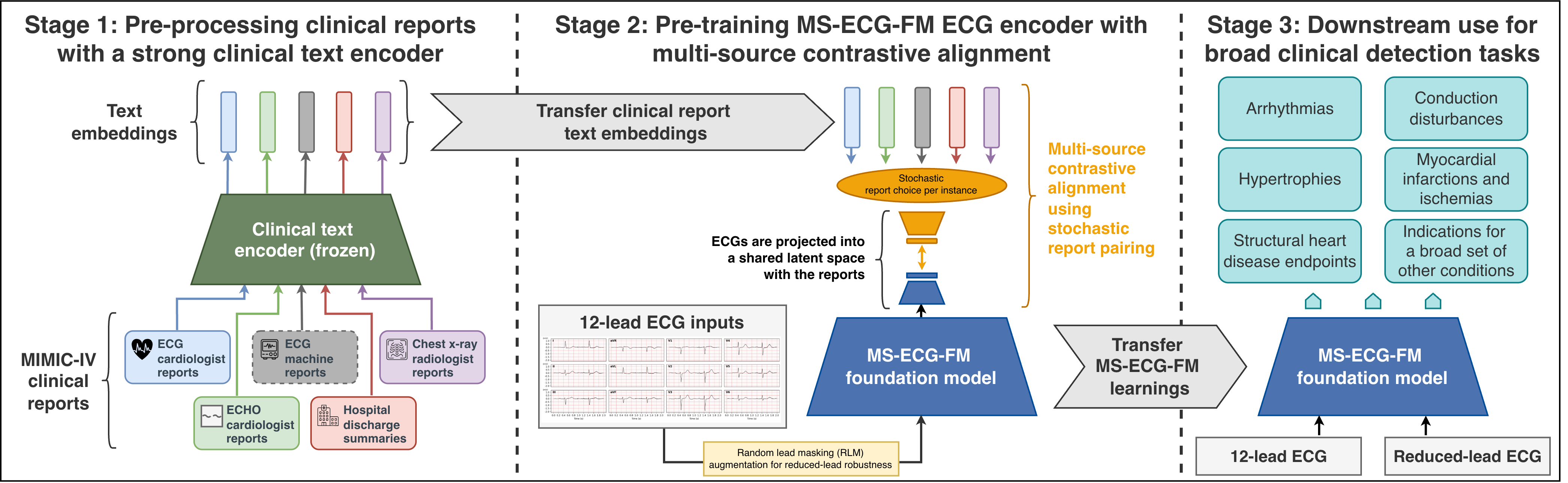}
    \caption{Conceptual overview of \methodname. \methodname{} is an ECG foundation model that can support a broader range of ECG detection tasks. It is pre-trained using concurrent alignment to ECG, ECHO, chest X-ray and discharge reports to improve the universality of its representations.}\label{fig:summary_figure}
\end{figure*}

Therefore, though the content of ECG \textit{reports} is restricted to a subset of diseases determined by current clinical usage, this does not imply that ECG \textit{recordings} are devoid of latent signatures related to many other conditions. Furthermore, by recognizing that an ECG recording is often collected in the context of a broader clinical encounter, notes from other diagnostic tests that assess cardiac function and structure, or indeed broader systemic disease, may be available to jointly supervise the pre-training of an ECG-FM. Used together, these notes can provide broader and more nuanced diagnostic information to learn ECG representations from.

Building from these insights, this work contributes a new ECG foundation model --- \methodname{} --- that is pre-trained using contrastive alignment to a variety of different clinical notes (e.g., ECG, ECHO, chest X-rays, and discharge summaries) that occur sparsely and asynchronously in a patient's medical record. \methodname{} substantially outperforms existing ECG foundation models across well-established ECG benchmark datasets and, importantly, it addresses the significant blind-spot of prior methods on conditions such as structural heart disease \cite{berger2026position}. Beyond a stronger ECG foundation model, we also contribute an extended evaluation framework for ECG-FMs, designed to probe their \emph{universality} by incorporating more labels and reduced-lead evaluations, as well as to provide clearer takeaways about their abilities through new visualizations (\emph{performance profiles}) and logical groupings of condition detection tasks. The overarching goal of this work is to provide insights and reusable artifacts to the clinical research community which can facilitate a new generation of more universal, more accurate ECG foundation models to support real-world prevention and diagnostic applications.

\section{Results}\label{sec:results}

\subsection*{\methodname{} universally outperforms contemporary ECG foundation models at disease detection}

\begin{figure}[t]
    \centering
    \includegraphics[width=\textwidth]{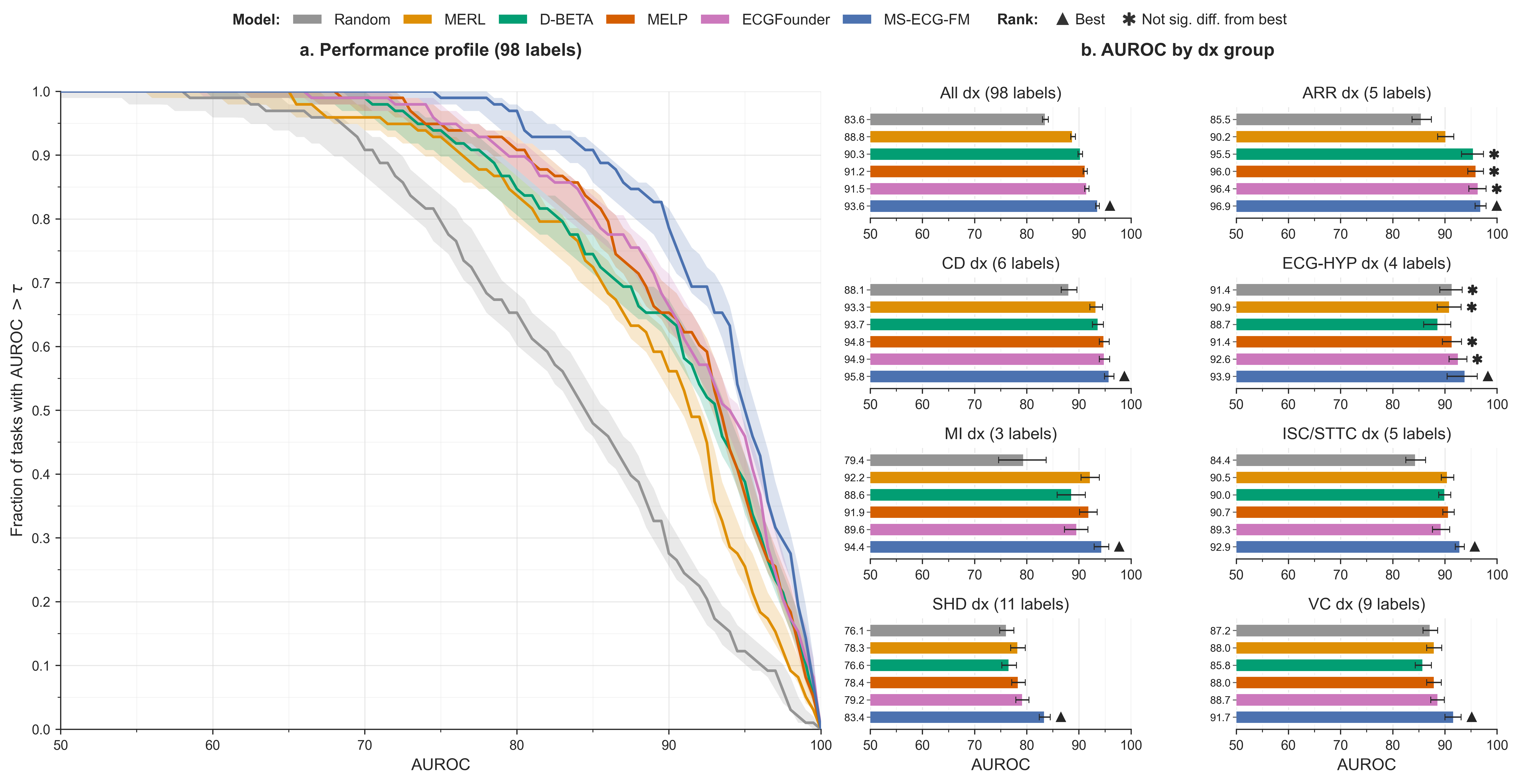}
    \caption{Comparison of \methodname{} to other contemporary ECG foundation models on 12-lead ECG diagnostic tasks using linear probe evaluations. (a): The performance profile shows that \methodname{} universally outperforms the other methods on the 98 downstream labels. (b): \methodname{} is strongest for all diagnostic groups, with large gains on several important diagnosis (dx) groups. Random refers to a randomly initialized ECG encoder. Significance testing is performed using a paired patient-cluster bootstrap procedure (see Methods). 95\% bootstrap confidence intervals are shown. AUROC: area under the receiver operating characteristic curve, ARR: arrhythmia, CD: conduction disturbances, ECG-HYP: hypertrophies detected from ECG, MI: myocardial infarction, ISC/STTC: ischemia and ST/T changes, SHD: structural heart disease, VC: diagnoses that rely on voltage criteria. Conditions by diagnosis group are in \cref{tab:downstream_diagnosis_groups}.}
    \label{fig:perf_profile_external}
\end{figure}

The \methodname{} modeling and evaluation approach is summarized conceptually in \cref{fig:summary_figure}. \cref{fig:perf_profile_external} compares \methodname{}'s performance to that of contemporary ECG-FMs using the area under the receiver operating characteristic curve (AUROC) score on a broad set of downstream disease detection tasks using held-out data from different hospital systems (see Methods and \cref{app:datasets} for dataset details). We compare to MERL \cite{liu2024zero}, D-BETA \cite{hung2025boosting} and MELP \cite{wang2025from} given their state-of-the-art reported performance on ECG benchmarks, and because they are also pre-trained on MIMIC-IV-ECG \cite{gow2023mimic}, facilitating controlled comparison of results. We also compare to ECGFounder \cite{li_ecgfounder2025} given it is pre-trained on an order of magnitude more data --- 10,771,552 ECGs versus at most 800,035 ECGs for MIMIC-IV-ECG.\footnote{\methodname{} pre-trains only on the \emph{train} split of this dataset which contains 720,102 ECGs, while the other methods use all 800,035 ECGs.} Finally, we include a randomly initialized encoder (Random) with the same encoder architecture as \methodname{} to establish an empirical lower bound on performance \cite{berger2026position}.\footnote{The Random method is not pre-trained, so an ECG-FM must outperform it to be considered a useful provider of representations for that task. This is less trivial than might be expected, as some information is preserved in embedding spaces defined by randomly initialized encoder weights \cite{Hermann2020}.} See \cref{app:review_ecg_fms} for further details on these methods. 

\cref{fig:perf_profile_external}.a. shows a \emph{performance profile}, a visualization we introduce to the ECG-FM benchmarking context \cite{dolan2002benchmarking,Agarwal2021}. The core goal of foundation models is to provide \emph{universal} data representations that are consistently strong across a broad set of downstream tasks. A performance profile operationalizes the evaluation of this goal, summarizing the distribution of a model's performance over all binary detection evaluation tasks. Unlike average performance scores which can be overly influenced by outliers (e.g., abnormally strong performance on a small set of tasks) and which discard information about the distribution of performance over tasks, performance profiles are robust and give a more fine-grained view of relative performance on the full evaluation suite. Here we see that \methodname{}'s performance distribution is superior to those of the baselines: at all intermediate AUROC thresholds (0.70 to 1.00), it supports a higher proportion of downstream tasks to perform at or above this level.

The diagnosis groups in \cref{fig:perf_profile_external}.b. provide further insight on the improvements of \methodname{} over contemporary methods. Performance on arrhythmias (ARR) and ECG-detected hypertrophies (ECG-HYP) is in line with the strongest baseline methods. At the same time, \methodname{} has considerably stronger performance on labels in the conduction disturbances (CD), myocardial infarction (MI), ischemia and ST/T changes (ISC/STTC), structural heart disease (SHD), and voltage criteria related (VC) categories. Overall, the macro-averaged AUROC for \methodname{} across all these tasks is 93.6 compared to 91.5 for ECGFounder, the strongest baseline overall, and 91.2 for MELP, the strongest baseline that uses the same pre-training data.

Finally, to complement \cref{fig:perf_profile_external}, we report \methodname{}'s performance on the standard ECG-FM benchmark table in \cref{tab:main_benchmark}. This table compares the performance of each method on each out-of-distribution dataset (the same underlying datasets as \cref{fig:perf_profile_external}, described in Methods) at different fractions of available training labels (1\%, 10\%, 100\%). \methodname{} frequently outperforms other methods by a large margin, and where it is not the strongest model (e.g., PTBXL-Rhythm and some lower label fraction results), its performance is not significantly different from the strongest model with only one exception. Going forwards, we advocate for the diagnostic groupings we propose in \cref{fig:perf_profile_external} as the primary summary of ECG-FM performance because they support intuitive takeaways on an ECG-FM's strengths and weaknesses along clinically meaningful groupings, which in turn can inform prioritization of further data collection or methodological improvements. However, it is important to still report \cref{tab:main_benchmark} here given the de facto status of this table in contemporary literature. 

\begin{sidewaystable}[p]
    \setlength\tabcolsep{2pt}
    \fontsize{7}{8}\selectfont
    \caption{Comparison of \methodname{} to other contemporary ECG foundation models on standard 12-lead ECG benchmark datasets using linear probe evaluations. \methodname{} is frequently the strongest model by a large margin. When it is not the strongest model (e.g., PTBXL-Rhythm and some lower label fraction results), its performance is often not significantly different from the strongest model. The evaluation metric is AUROC: area under the receiver operating characteristic curve.}
    \label{tab:main_benchmark}
    \begin{tabular}{llrrr|rrr|rrr|rrr|rrr|rrr|rrr}
    \toprule
    \multirow{2}{*}{\textbf{Methods}} & & \multicolumn{3}{c}{\textbf{PTBXL-Super}} & \multicolumn{3}{c}{\textbf{PTBXL-Sub}} & \multicolumn{3}{c}{\textbf{PTBXL-Form}} & \multicolumn{3}{c}{\textbf{PTBXL-Rhythm}} & \multicolumn{3}{c}{\textbf{CPSC2018}} & \multicolumn{3}{c}{\textbf{CSN}} & \multicolumn{3}{c}{\textbf{EchoNext-Mini}} \\
    \cmidrule{3-5} \cmidrule(lr){6-8} \cmidrule(lr){9-11} \cmidrule(lr){12-14} \cmidrule(lr){15-17} \cmidrule(lr){18-20} \cmidrule(lr){21-23}
     &  & \textbf{1\%} & \textbf{10\%} & \textbf{100\%} & \textbf{1\%} & \textbf{10\%} & \textbf{100\%} & \textbf{1\%} & \textbf{10\%} & \textbf{100\%} & \textbf{1\%} & \textbf{10\%} & \textbf{100\%} & \textbf{1\%} & \textbf{10\%} & \textbf{100\%} & \textbf{1\%} & \textbf{10\%} & \textbf{100\%} & \textbf{1\%} & \textbf{10\%} & \textbf{100\%} \\
    \midrule
    Random && 82.0 & 85.2 & 87.0 & 71.7 & 83.1 & 85.7 & 56.9 & 73.0 & 72.6 & 70.0 & 84.6 & 87.3 & 69.5 & 81.3 & 83.6 & 65.2 & 81.5 & 86.2 & 67.1 & 71.3 & 76.1 \\ \midrule
    MERL && 86.3 & 88.5 & 89.9 & 81.0 & 88.5 & 92.0 & 59.3 & 73.5 & 80.7 & 79.2 & 90.7 & 90.1 & 78.9 & 87.6 & 90.2 & 76.6 & 87.8 & 92.4 & 65.5 & 74.2 & 78.3 \\
    D-BETA && 86.0 & 88.9 & 89.8 & 81.0 & 86.4 & 90.2 & $^{\ast}$67.7 & 78.5 & 85.0 & \textbf{89.8} & $^{\ast}$95.8 & $^{\ast}$96.8 & \textbf{88.9} & 92.8 & 95.1 & $^{\ast}$82.7 & 90.5 & 94.3 & 69.9 & 74.6 & 76.6 \\
    MELP && 86.4 & 89.3 & 90.3 & 81.9 & 90.2 & 92.6 & \textbf{69.0} & \textbf{81.7} & 85.4 & $^{\ast}$89.5 & \textbf{96.6} & $^{\ast}$96.6 & 87.4 & 92.8 & 94.0 & $^{\ast}$81.8 & 90.9 & 94.8 & 69.6 & 76.1 & 78.4 \\
    ECGFounder && 85.6 & 88.9 & 90.2 & 82.9 & 89.4 & 92.5 & 64.2 & 77.7 & 86.3 & 83.8 & $^{\ast}$96.1 & \textbf{97.0} & 83.4 & 92.1 & $^{\ast}$95.3 & \textbf{83.3} & $^{\ast}$92.5 & 95.8 & 68.3 & 77.2 & 79.2 \\ \midrule
    \textbf{MS-ECG-FM} && \textbf{89.5} & \textbf{92.2} & \textbf{93.2} & \textbf{84.8} & \textbf{92.9} & \textbf{95.0} & $^{\ast}$66.7 & $^{\ast}$81.7 & \textbf{89.7} & 85.6 & $^{\ast}$96.4 & $^{\ast}$96.7 & $^{\ast}$88.4 & \textbf{94.0} & \textbf{95.8} & $^{\ast}$82.7 & \textbf{93.4} & \textbf{96.8} & \textbf{74.9} & \textbf{81.7} & \textbf{83.4} \\
    \bottomrule
\end{tabular}

    \footnotetext{The best model in each column is \textbf{bold}; $\ast$ marks models that are not significantly different from the best per a paired patient-cluster bootstrap significance test (see Methods). Random refers to a randomly initialized ECG encoder. Results are shown for different percentages of available training labels (1\%, 10\%, 100\%).}
    \footnotetext{Each major column header is a different ECG benchmarking dataset, described further in Methods. Each minor column header is a different percentage of available training labels (1\%, 10\%, 100\%).}
\end{sidewaystable}

\subsection*{\methodname{} also supports reduced-lead ECG configurations}

\begin{table}[h!]
    \centering
    \setlength\tabcolsep{2pt}
    \fontsize{5}{6}\selectfont
    \caption{Performance of \methodname{} and contemporary ECG-FMs on detection tasks with common reduced-lead configurations. With the exception of arrhythmias (ARR) and ECG-HYP detected from single lead V2, \methodname{} is consistently the strongest method. AUROC: area under the receiver operating characteristic curve.}
    \label{tab:reduced_lead}
    \begin{tabular}{llrrr|rrrr|r}
    \toprule
    & & \multicolumn{3}{c}{\textbf{Single Lead}} & \multicolumn{4}{c}{\textbf{Lead Subsets}} & \multicolumn{1}{c}{\textbf{12-lead}} \\
    \cmidrule{3-5} \cmidrule(lr){6-9} \cmidrule(lr){10-10}
     &  & \textbf{I (LCX)} & \textbf{II (RCA)} & \textbf{V2 (LAD)} & \textbf{I,II} & \textbf{All limb} & \textbf{I,II,V2} & \textbf{I,II,III,V2} & \textbf{12-lead} \\
    \midrule
    \multirow{6}{*}{All} & Random & 76.4 & 77.7 & 73.7 & 81.1 & 81.7 & 81.0 & 82.3 & 83.6 \\ \cmidrule[0.2pt]{2-10}
     & MERL & 63.5 & 66.6 & 68.5 & 69.7 & 63.5 & 75.8 & 79.0 & 88.8 \\
     & D-BETA & 84.4 & 85.6 & 82.2 & 88.2 & 88.3 & 89.3 & 89.2 & 90.3 \\
     & MELP & 84.9 & 86.0 & 81.9 & 89.1 & 89.2 & 90.6 & 90.7 & 91.2 \\
     & ECGFounder & 86.2 & 86.4 & 83.1 & 89.0 & 88.9 & 90.0 & 90.8 & 91.5 \\ \cmidrule[0.2pt]{2-10}
     & \textbf{MS-ECG-FM} & \textbf{86.9} & \textbf{88.1} & \textbf{85.0} & \textbf{90.8} & \textbf{91.2} & \textbf{92.5} & \textbf{92.7} & \textbf{93.6} \\
    \midrule
    \multirow{6}{*}{ARR} & Random & 79.2 & 82.0 & 75.9 & 86.2 & 87.0 & 82.5 & 84.2 & 85.5 \\ \cmidrule[0.2pt]{2-10}
     & MERL & 61.5 & 64.7 & 71.5 & 59.6 & 58.6 & 76.8 & 79.6 & 90.2 \\
     & D-BETA & $^{\ast}$91.5 & $^{\ast}$95.6 & $^{\ast}$89.2 & $^{\ast}$96.3 & $^{\ast}$96.1 & 95.7 & $^{\ast}$95.8 & $^{\ast}$95.5 \\
     & MELP & 91.8 & $^{\ast}$96.1 & 88.8 & $^{\ast}$96.0 & $^{\ast}$96.0 & $^{\ast}$95.8 & 96.1 & $^{\ast}$96.0 \\
     & ECGFounder & \textbf{93.7} & $^{\ast}$96.3 & \textbf{91.6} & \textbf{96.6} & \textbf{96.5} & \textbf{97.1} & \textbf{97.0} & $^{\ast}$96.4 \\ \cmidrule[0.2pt]{2-10}
     & \textbf{MS-ECG-FM} & $^{\ast}$93.6 & \textbf{96.5} & $^{\ast}$90.7 & $^{\ast}$96.4 & $^{\ast}$96.3 & $^{\ast}$95.9 & 96.0 & \textbf{96.9} \\
    \midrule
    \multirow{6}{*}{CD} & Random & 79.1 & 83.5 & 82.0 & 86.8 & 87.6 & 87.5 & 88.0 & 88.1 \\ \cmidrule[0.2pt]{2-10}
     & MERL & 71.4 & 74.3 & 76.8 & 81.2 & 70.8 & 85.8 & 87.9 & 93.3 \\
     & D-BETA & 85.5 & 87.6 & 87.6 & 90.6 & 91.0 & 92.9 & 92.4 & 93.7 \\
     & MELP & 86.3 & 89.0 & 88.0 & 92.0 & 92.3 & $^{\ast}$94.2 & 94.1 & 94.8 \\
     & ECGFounder & 86.4 & 88.7 & 88.2 & 91.3 & 91.6 & 93.5 & 94.2 & 94.9 \\ \cmidrule[0.2pt]{2-10}
     & \textbf{MS-ECG-FM} & \textbf{87.8} & \textbf{90.8} & \textbf{89.3} & \textbf{93.1} & \textbf{93.3} & \textbf{94.8} & \textbf{94.9} & \textbf{95.8} \\
    \midrule
    \multirow{6}{*}{ECG-HYP} & Random & 79.5 & 83.8 & 75.9 & 87.1 & 88.0 & 84.9 & 86.0 & $^{\ast}$91.4 \\ \cmidrule[0.2pt]{2-10}
     & MERL & 71.4 & 71.9 & 64.9 & 74.0 & 63.7 & 80.2 & 83.8 & $^{\ast}$90.9 \\
     & D-BETA & 83.4 & 83.5 & $^{\ast}$79.7 & 86.4 & 87.0 & 87.5 & 87.1 & 88.7 \\
     & MELP & $^{\ast}$84.6 & 83.3 & $^{\ast}$80.3 & 88.8 & 89.6 & 90.3 & 90.7 & $^{\ast}$91.4 \\
     & ECGFounder & $^{\ast}$85.7 & 85.1 & $^{\ast}$80.7 & 89.0 & 89.4 & 90.5 & 91.4 & $^{\ast}$92.6 \\ \cmidrule[0.2pt]{2-10}
     & \textbf{MS-ECG-FM} & \textbf{87.7} & \textbf{89.2} & \textbf{81.5} & \textbf{91.5} & \textbf{92.7} & \textbf{93.6} & \textbf{93.8} & \textbf{93.9} \\
    \midrule
    \multirow{6}{*}{MI} & Random & 74.5 & 75.8 & 73.7 & 76.5 & 80.5 & 79.6 & 80.2 & 79.4 \\ \cmidrule[0.2pt]{2-10}
     & MERL & 60.1 & 67.6 & 67.1 & 73.5 & 66.8 & 75.4 & 82.5 & 92.2 \\
     & D-BETA & 77.1 & $^{\ast}$78.7 & $^{\ast}$77.7 & 86.6 & 86.8 & $^{\ast}$91.7 & $^{\ast}$92.0 & 88.6 \\
     & MELP & $^{\ast}$78.3 & $^{\ast}$79.9 & $^{\ast}$76.3 & $^{\ast}$89.0 & 89.2 & $^{\ast}$91.8 & 92.0 & 91.9 \\
     & ECGFounder & $^{\ast}$79.0 & $^{\ast}$79.8 & $^{\ast}$77.3 & $^{\ast}$87.7 & 87.9 & 87.8 & 92.2 & 89.6 \\ \cmidrule[0.2pt]{2-10}
     & \textbf{MS-ECG-FM} & \textbf{81.0} & \textbf{80.4} & \textbf{78.2} & \textbf{89.3} & \textbf{90.5} & \textbf{92.7} & \textbf{93.8} & \textbf{94.4} \\
    \midrule
    \multirow{6}{*}{ISC/STTC} & Random & 79.4 & 81.3 & 70.6 & 83.8 & 84.3 & 83.7 & 84.4 & 84.4 \\ \cmidrule[0.2pt]{2-10}
     & MERL & 60.0 & 64.9 & 64.0 & 65.9 & 60.7 & 75.6 & 80.0 & 90.5 \\
     & D-BETA & 82.6 & 84.1 & 73.1 & 89.1 & 89.2 & 88.0 & 87.7 & 90.0 \\
     & MELP & 83.5 & 84.2 & 74.3 & 89.3 & 89.2 & 90.0 & 90.3 & 90.7 \\
     & ECGFounder & $^{\ast}$84.3 & 84.7 & 73.6 & 88.1 & 87.3 & 88.6 & 90.1 & 89.3 \\ \cmidrule[0.2pt]{2-10}
     & \textbf{MS-ECG-FM} & \textbf{85.6} & \textbf{86.5} & \textbf{77.7} & \textbf{90.8} & \textbf{90.8} & \textbf{91.9} & \textbf{91.9} & \textbf{92.9} \\
    \midrule
    \multirow{6}{*}{SHD} & Random & 74.7 & 71.7 & 69.5 & 75.0 & 74.4 & 75.3 & 75.3 & 76.1 \\ \cmidrule[0.2pt]{2-10}
     & MERL & 66.4 & 62.6 & 65.0 & 68.2 & 66.9 & 71.9 & 73.5 & 78.3 \\
     & D-BETA & 75.2 & 72.6 & 72.3 & 74.9 & 75.3 & 75.8 & 75.9 & 76.6 \\
     & MELP & 76.2 & 74.5 & 73.2 & 77.1 & 77.2 & 78.2 & 78.0 & 78.4 \\
     & ECGFounder & 77.1 & 75.0 & 73.7 & 77.8 & 78.0 & 78.1 & 78.1 & 79.2 \\ \cmidrule[0.2pt]{2-10}
     & \textbf{MS-ECG-FM} & \textbf{78.8} & \textbf{77.5} & \textbf{77.1} & \textbf{79.7} & \textbf{80.5} & \textbf{82.1} & \textbf{82.2} & \textbf{83.4} \\
    \midrule
    \multirow{6}{*}{VC} & Random & 77.6 & 79.4 & 70.3 & 84.9 & 85.1 & 83.2 & 84.1 & 87.2 \\ \cmidrule[0.2pt]{2-10}
     & MERL & 67.7 & 71.5 & 62.6 & 73.8 & 63.7 & 76.4 & 79.3 & 88.0 \\
     & D-BETA & 78.7 & 80.4 & 70.6 & 85.1 & 85.2 & 84.5 & 84.1 & 85.8 \\
     & MELP & 79.8 & 80.6 & 71.8 & 86.0 & 86.3 & 86.6 & 86.8 & 88.0 \\
     & ECGFounder & 80.8 & 80.2 & 72.4 & 85.4 & 85.4 & 85.9 & 86.8 & 88.7 \\ \cmidrule[0.2pt]{2-10}
     & \textbf{MS-ECG-FM} & \textbf{83.1} & \textbf{84.8} & \textbf{76.0} & \textbf{89.1} & \textbf{90.0} & \textbf{90.3} & \textbf{90.6} & \textbf{91.7} \\
    \bottomrule
\end{tabular}

    \footnotetext{Linear probe evaluations. Random refers to a randomly initialized ECG encoder. The best model in each column is \textbf{bold}; $\ast$ marks models that are not significantly different from the best per a paired patient-cluster bootstrap significance test (see Methods).}
    \footnotetext{LCX: left circumflex, RCA: right coronary artery, LAD: left anterior descending, ARR: arrhythmia, CD: conduction disturbances, ECG-HYP: hypertrophies detected from ECG, MI: myocardial infarction, ISC/STTC: ischemia and ST/T changes, SHD: structural heart disease, VC: diagnoses that rely on voltage criteria.}
\end{table}

12-lead ECG data is generally better characterized given its collection in clinical settings alongside clinical notes and other metadata, thus making it the primary target for pre-training ECG-FMs. However, reduced-lead applications exist in bedside monitors and are becoming more prominent in ambulatory settings --- e.g., Holter monitors, patches or wrist-worn wearable devices. Thus, ECG-FMs should be able to support these too. The PhysioNet/Computing in Cardiology Challenge 2021 considered the task of ECG diagnosis using common reduced-lead configurations \cite{PhysioNet-challenge-2021-1.0.3}, proposing to study performance with Leads I, II and V2 individually, which are oriented towards the lateral, inferior and anterior walls of the left ventricle, respectively, as well as with several reduced-lead subsets. We make use of these reduced-lead configurations here, reporting results by diagnostic group in \cref{tab:reduced_lead}. 

When reducing the number of ECG leads available for a task, some information about cardiac function --- manifested as electrical potentials in certain directions --- becomes irretrievable if it does not occur in the subspace spanned by the remaining leads. However, performance can still be strong for certain conditions in reduced-lead configurations. \cref{tab:reduced_lead} shows that \methodname{} maintains its strong performance, significantly surpassing the baseline methods on average across all tasks, and generally exceeding them on each diagnostic grouping with the exception of arrhythmia. Several steps were taken to promote \methodname{}'s reduced-lead transfer performance as discussed further in Methods.

It is also noteworthy that performance on the detection of arrhythmias (ARR) systematically drops off less abruptly for single leads I and II compared to the other diagnostic groups (with the exception of MERL, which underperforms in all reduced-lead configurations). This may be explained by how arrhythmias manifest, mediating the temporal structure of many leads simultaneously, compared to other conditions such as infarction and ischemia, which affect the morphological structure of the ECG in a lead-specific manner due to their regional occurrences in the heart. Relatedly, it is interesting that reduced-lead subsets that contain leads viewing the heart from the three key anatomical angles (i.e., ``I, II, V2'' and ``I, II, III, V2'' concurrently view the anterior, inferior and lateral walls) experience only a mild reduction in performance versus the 12-lead setting. \cref{tab:label_level_lead_auroc} and \cref{tab:label_level_lead_auprc} report performance by lead subset at the level of individual labels, providing further insight on these performance trends (e.g., the relative strengths of single leads I, II and V2 on lateral, inferior and anterior myocardial infarction, respectively).

\subsection*{Alignment to multiple types of clinical report grants \methodname{} universally strong performance}

\begin{figure*}[t]
    \centering
    \includegraphics[width=\textwidth]{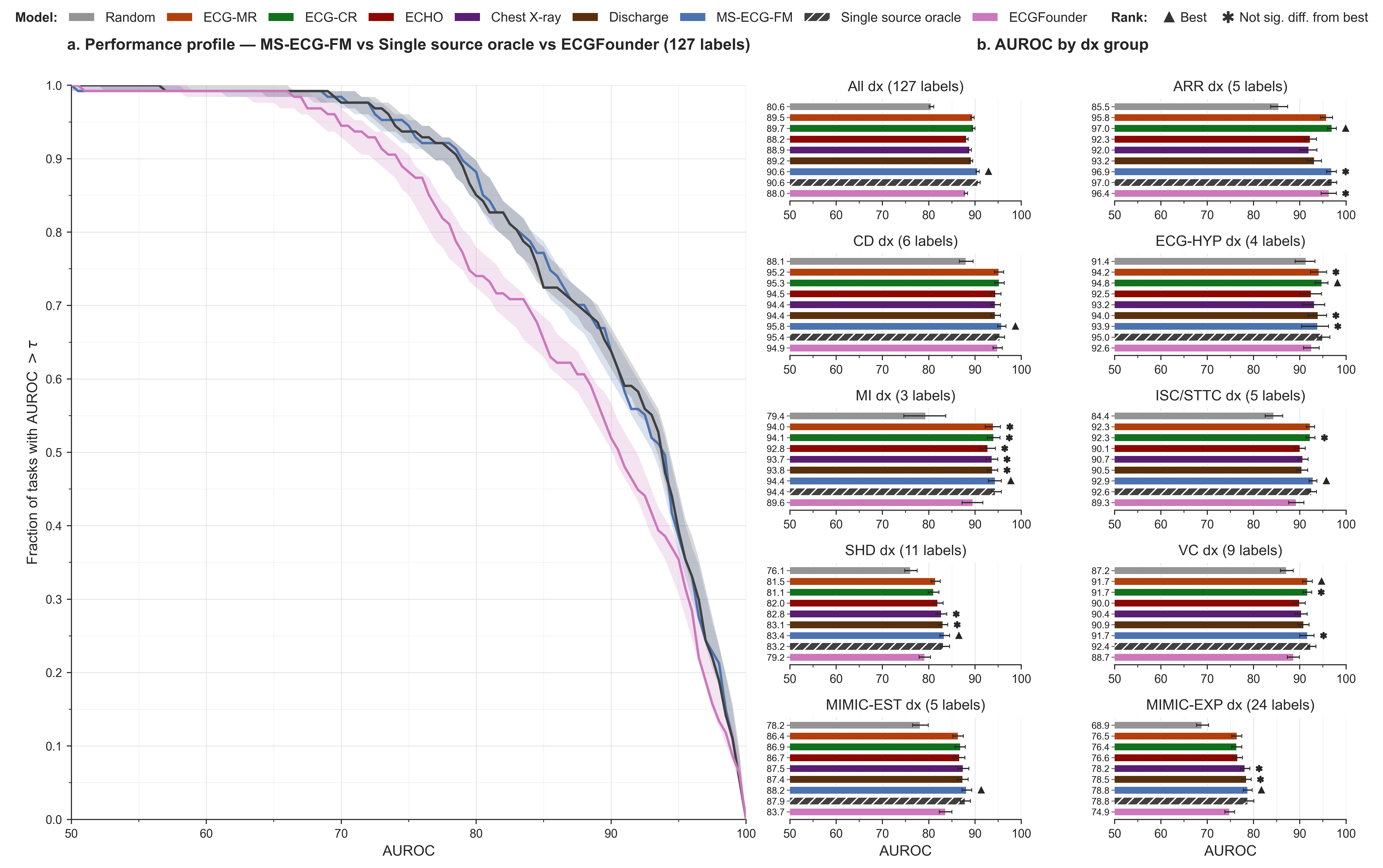}
    \caption{Contribution of different clinical note types to downstream performance. The single source oracle represents the performance when selecting the strongest single source method on each label. (a): \methodname{} matches the performance of the oracle by learning the union of information across clinical report types. (b): Alignment to different clinical note types results in stronger representations for different diagnostic tasks. Random refers to a randomly initialized ECG encoder. Significance testing is performed using a paired patient-cluster bootstrap procedure (see Methods). 95\% bootstrap confidence intervals are shown. AUROC: area under the receiver operating characteristic curve, ARR: arrhythmia, CD: conduction disturbances, ECG-HYP: hypertrophies detected from ECG, MI: myocardial infarction, ISC/STTC: ischemia and ST/T changes, SHD: structural heart disease, VC: diagnoses that rely on voltage criteria, MIMIC-EST: ICD-based task on conditions with established ECG evidence, MIMIC-EXP: ICD-based labels on conditions with exploratory ECG evidence. Conditions by diagnosis group are in \cref{tab:downstream_diagnosis_groups}.}
    \label{fig:perf_profile_internal}
\end{figure*}

The type of clinical report used for contrastive pre-training has an impact on the quality of an ECG foundation model's representations for specific downstream tasks. No single note type is universally strongest on all downstream tasks due to their different clinical focuses. \cref{fig:perf_profile_internal} demonstrates this empirically, where here the number of downstream labels is expanded to 127 to also include International Classification of Diseases (ICD) based labels covering a broader range of conditions for which there is at least exploratory evidence of detection using ECG (see \cref{tab:review_ecg_outcomes}).\footnote{Importantly, these labels are derived from the test set of MIMIC-IV-ECG-Ext-ICD \cite{strodthoff2024data} which is derived from MIMIC-IV-ECG \cite{gow2023mimic}. Because multiple baseline ECG-FMs (MERL, D-BETA, MELP) are pre-trained using \emph{all} 800,035 MIMIC-IV-ECG instances, we only use MIMIC-IV-ECG-Ext-ICD for comparisons between our methods where leakage is strictly controlled by only pre-training on the 720,102 train instances, and ECGFounder which was pre-trained on a different dataset.} First, \cref{fig:perf_profile_internal}.a. shows that \methodname{}'s multi-source contrastive alignment objective allows it to capture much of the information contained within the five distinct report types. The gray line is an \emph{oracle} method representing the performance profile of the best single source method for each label: \methodname{} has a very similar performance distribution, suggesting it encodes the union of information across the reports. 

\cref{fig:perf_profile_internal}.b. provides insights on how different clinical note types influence representation quality. ECG cardiologist reports (ECG-CR, AUROC=89.7) are the strongest overall single source, closely followed by ECG machine reports (ECG-MR, AUROC=89.5). These report types are notably stronger than the others in detection of arrhythmias (ARR), conduction disturbances (CD), ischemia and ST/T changes (ISC/STTC), and voltage criteria related conditions (VC). This is likely explained by the frequent and explicit mention of these conditions in the ECG-CR and ECG-MR notes given the clinical use of ECG to detect them. However, alignment to the other sources clearly provides orthogonal value. For example, discharge summaries and chest X-ray notes are the most useful for structural heart disease (SHD) conditions, as well as for the exploratory ECG outcomes grouped under MIMIC-EXP. The utility of all report types for MI detection is also noteworthy, in particular the ECHO and chest X-ray imaging reports. The PTB-XL MI labels \cite{wagner2020ptb} are stage-agnostic and, where stage is recorded, they predominantly denote prior rather than acute infarction. The finding that imaging reports are useful might thus reflect documentation on visual evidence of prior MI in these reports.

\begin{table}[h!]
    \centering
    \setlength\tabcolsep{1.5pt}
    \fontsize{4.5}{5}\selectfont
    \caption{Contribution of different clinical report types to downstream performance for different condition detection tasks. A glossary of condition codes is in \cref{tab:condition_codes}. AUROC: area under the receiver operating characteristic curve.}
    \label{tab:label_level_main_auroc}
    \begin{tabular}{l>{\raggedright\arraybackslash}p{0.13\linewidth}c|r|rrrrr|r}
    \toprule
    & & \multicolumn{1}{c}{} & \multicolumn{7}{c}{\textbf{AUROC}} \\
    \cmidrule(lr){4-10}
    \textbf{Group} & \textbf{Condition} & \textbf{Prevalence \% (N)} & \textbf{Random} & \textbf{ECG-MR} & \textbf{ECG-CR} & \textbf{ECHO} & \textbf{Chest X-ray} & \textbf{Discharge} & \textbf{MS-ECG-FM} \\
    \midrule
    \multirow{5}{*}{ARR} & AFIB & 7.2\% (152) & 93.2 & $^{\ast}$98.4 & $^{\ast}$98.5 & $^{\ast}$98.0 & 97.9 & 97.9 & \textbf{98.7} \\
     & STACH & 3.9\% (82) & 97.7 & $^{\ast}$99.4 & \textbf{99.5} & 99.2 & 98.8 & 98.9 & $^{\ast}$99.5 \\
     & SARRH & 3.7\% (77) & 58.2 & 89.8 & $^{\ast}$94.1 & 74.0 & 75.0 & 78.7 & \textbf{94.4} \\
     & SBRAD & 3.1\% (64) & 93.5 & $^{\ast}$96.5 & $^{\ast}$96.6 & 95.8 & $^{\ast}$95.4 & $^{\ast}$96.4 & \textbf{96.7} \\
     & SVARR & 0.7\% (14) & 84.8 & $^{\ast}$95.0 & \textbf{96.4} & $^{\ast}$94.5 & 92.6 & $^{\ast}$94.1 & $^{\ast}$95.4 \\
    \midrule
    \multirow{6}{*}{CD} & LAF/LPF Blocks & 8.3\% (179) & 90.4 & $^{\ast}$98.0 & $^{\ast}$97.8 & 96.6 & 96.4 & 96.4 & \textbf{98.2} \\
     & IRBB Block & 5.2\% (112) & 85.2 & 97.6 & $^{\ast}$97.9 & 95.9 & 96.3 & 95.7 & \textbf{98.1} \\
     & CLBB Block & 2.5\% (54) & 99.6 & $^{\ast}$99.8 & $^{\ast}$99.6 & \textbf{99.9} & $^{\ast}$99.6 & $^{\ast}$99.9 & $^{\ast}$99.8 \\
     & CRBB Block & 2.5\% (54) & 98.7 & \textbf{99.8} & $^{\ast}$99.8 & 99.7 & $^{\ast}$99.7 & 99.7 & $^{\ast}$99.8 \\
     & AV Block & 3.8\% (82) & 81.6 & 97.6 & $^{\ast}$97.9 & 97.3 & $^{\ast}$97.5 & 97.3 & \textbf{98.1} \\
     & IVCD & 3.7\% (79) & 73.0 & 78.5 & $^{\ast}$78.9 & 77.5 & 76.9 & 77.6 & \textbf{80.8} \\
    \midrule
    \multirow{4}{*}{ECG-HYP} & LVH & 9.9\% (214) & 92.9 & \textbf{95.3} & 94.8 & $^{\ast}$94.9 & $^{\ast}$94.7 & $^{\ast}$94.7 & $^{\ast}$95.3 \\
     & RVH & 0.6\% (12) & $^{\ast}$94.1 & $^{\ast}$95.1 & \textbf{95.4} & $^{\ast}$92.4 & $^{\ast}$91.4 & $^{\ast}$92.3 & $^{\ast}$91.2 \\
     & LAO/LAE & 1.9\% (42) & 80.8 & 87.2 & $^{\ast}$89.7 & 85.5 & $^{\ast}$87.9 & \textbf{90.0} & $^{\ast}$89.7 \\
     & RAO/RAE & 0.5\% (10) & 97.6 & $^{\ast}$99.1 & $^{\ast}$99.4 & 97.1 & 98.9 & $^{\ast}$99.0 & \textbf{99.5} \\
    \midrule
    \multirow{3}{*}{MI} & Anterior MI & 14.2\% (306) & 88.3 & $^{\ast}$96.9 & \textbf{97.2} & 95.9 & 95.7 & 96.0 & $^{\ast}$97.1 \\
     & Inferior MI & 15.2\% (327) & 77.1 & $^{\ast}$94.2 & $^{\ast}$94.4 & $^{\ast}$93.9 & $^{\ast}$93.9 & $^{\ast}$93.9 & \textbf{94.6} \\
     & Lateral MI & 0.9\% (20) & 72.7 & $^{\ast}$91.0 & $^{\ast}$90.7 & $^{\ast}$88.5 & $^{\ast}$91.4 & \textbf{91.6} & $^{\ast}$91.4 \\
    \midrule
    \multirow{5}{*}{ISC/STTC} & Anterior ISC & 4.3\% (93) & 86.6 & $^{\ast}$93.8 & $^{\ast}$93.8 & 91.0 & 91.4 & 92.3 & \textbf{94.1} \\
     & Inferior ISC & 1.9\% (40) & 79.5 & 95.2 & \textbf{96.4} & 92.0 & 93.3 & 91.5 & $^{\ast}$96.3 \\
     & Non-specific ISC & 5.9\% (128) & 94.3 & \textbf{96.9} & 96.1 & 95.7 & 95.5 & 96.0 & $^{\ast}$96.7 \\
     & STTC & 10.3\% (222) & 84.0 & $^{\ast}$89.6 & $^{\ast}$89.8 & 87.5 & 87.8 & 87.5 & \textbf{90.0} \\
     & Non-specific STC & 3.6\% (77) & 77.9 & $^{\ast}$86.1 & $^{\ast}$85.6 & 84.1 & $^{\ast}$85.5 & $^{\ast}$85.3 & \textbf{87.2} \\
    \midrule
    \multirow{11}{*}{SHD} & Reduced LVEF & 17.7\% (962) & 84.9 & 89.2 & 89.4 & 90.1 & $^{\ast}$90.3 & $^{\ast}$90.6 & \textbf{90.7} \\
     & Increased LVWT & 19.5\% (1,061) & 73.6 & 77.5 & 76.9 & 77.4 & $^{\ast}$77.8 & $^{\ast}$78.0 & \textbf{78.4} \\
     & Aortic Stenosis & 5.3\% (286) & 75.9 & 83.5 & 83.5 & $^{\ast}$84.4 & 83.6 & $^{\ast}$84.7 & \textbf{85.4} \\
     & Aortic Regurg. & 1.2\% (66) & 71.7 & $^{\ast}$74.9 & 71.6 & $^{\ast}$76.7 & \textbf{76.9} & $^{\ast}$76.8 & $^{\ast}$75.0 \\
     & Mitral Regurg. & 6.2\% (337) & 77.7 & 82.4 & 82.0 & 82.9 & 83.3 & $^{\ast}$84.0 & \textbf{84.2} \\
     & Tricuspid Regurg. & 6.5\% (353) & 77.0 & 82.1 & 82.7 & 83.8 & $^{\ast}$84.8 & $^{\ast}$84.9 & \textbf{85.0} \\
     & Pulm. Regurg. & 0.4\% (20) & 80.7 & $^{\ast}$89.3 & $^{\ast}$88.4 & 85.4 & $^{\ast}$87.9 & $^{\ast}$87.9 & \textbf{90.2} \\
     & RV Sys. Dys. & 7.7\% (419) & 84.4 & 88.2 & 88.4 & $^{\ast}$88.8 & $^{\ast}$88.8 & $^{\ast}$89.3 & \textbf{89.5} \\
     & Pericardial Eff. & 1.3\% (69) & 69.6 & 75.2 & 76.0 & $^{\ast}$75.9 & $^{\ast}$77.5 & $^{\ast}$78.6 & \textbf{79.1} \\
     & Elevated PASP & 12.8\% (699) & 72.1 & 78.0 & 77.5 & 78.7 & \textbf{80.3} & $^{\ast}$80.0 & $^{\ast}$80.3 \\
     & Elevated TRV & 6.9\% (375) & 69.9 & 76.1 & 75.9 & 77.8 & $^{\ast}$79.2 & $^{\ast}$79.1 & \textbf{80.1} \\
    \midrule
    \multirow{5}{*}{MIMIC-EST} & I21.0-3: STEMI & 1.0\% (118) & 81.2 & 90.2 & $^{\ast}$92.0 & 90.9 & $^{\ast}$93.3 & $^{\ast}$92.4 & \textbf{93.4} \\
     & I21.4: NSTEMI & 3.2\% (396) & 72.6 & 82.5 & 82.3 & 82.2 & 83.0 & $^{\ast}$83.9 & \textbf{85.1} \\
     & I25.2: Prior MI & 7.1\% (880) & 70.2 & 78.5 & 78.4 & $^{\ast}$79.4 & $^{\ast}$79.6 & $^{\ast}$79.3 & \textbf{79.8} \\
     & I50.2: Sys. HF & 6.1\% (760) & 86.7 & 91.2 & 91.2 & $^{\ast}$91.7 & $^{\ast}$92.0 & $^{\ast}$92.1 & \textbf{92.2} \\
     & I44.1-2: 2/3 AVB & 1.0\% (118) & 80.6 & $^{\ast}$89.7 & $^{\ast}$90.3 & $^{\ast}$89.1 & $^{\ast}$89.6 & 89.0 & \textbf{90.7} \\
    \midrule
    \multirow{25}{*}{MIMIC-EXP} & I25.1: AHD/CAD & 21.4\% (2,641) & 72.4 & 79.2 & 79.4 & 79.7 & $^{\ast}$80.5 & $^{\ast}$80.7 & \textbf{81.1} \\
     & I50.3: Dia. HF & 7.8\% (968) & 74.0 & 80.3 & 80.8 & 81.3 & \textbf{83.4} & $^{\ast}$83.3 & $^{\ast}$83.0 \\
     & I45.81: LongQT & 0.4\% (47) & 69.6 & $^{\ast}$81.8 & $^{\ast}$81.8 & $^{\ast}$81.7 & \textbf{84.3} & $^{\ast}$81.9 & $^{\ast}$82.5 \\
     & I30: Pericarditis & 0.3\% (32) & 70.5 & $^{\ast}$85.2 & $^{\ast}$83.1 & $^{\ast}$86.1 & \textbf{86.9} & $^{\ast}$85.0 & $^{\ast}$86.4 \\
     & I42.0: DCM & 0.1\% (17) & 84.4 & 90.9 & 88.3 & $^{\ast}$92.5 & 90.6 & $^{\ast}$92.4 & \textbf{94.0} \\
     & I42.1-2: HCM & 0.3\% (35) & $^{\ast}$75.4 & $^{\ast}$79.0 & $^{\ast}$76.8 & $^{\ast}$76.3 & $^{\ast}$78.5 & $^{\ast}$78.5 & \textbf{82.1} \\
     & I42.8: OtherCM & 1.9\% (233) & 84.7 & $^{\ast}$87.5 & 87.2 & $^{\ast}$87.4 & $^{\ast}$87.5 & $^{\ast}$87.8 & \textbf{88.4} \\
     & E87.5: HyperK & 4.0\% (490) & 68.1 & 76.9 & 77.8 & 77.7 & $^{\ast}$80.2 & \textbf{80.5} & $^{\ast}$80.3 \\
     & E87.6: HypoK & 2.8\% (348) & 63.3 & 73.1 & 72.8 & 72.3 & $^{\ast}$73.7 & $^{\ast}$73.9 & \textbf{75.2} \\
     & E83.52: HyperCa & 0.6\% (71) & 55.9 & $^{\ast}$67.4 & $^{\ast}$65.8 & 63.1 & $^{\ast}$68.8 & \textbf{69.2} & $^{\ast}$68.8 \\
     & E83.51: HypoCa & 0.7\% (83) & 69.4 & $^{\ast}$76.0 & $^{\ast}$75.5 & 74.2 & 75.4 & $^{\ast}$78.0 & \textbf{79.0} \\
     & E87.0: HyperNa & 2.1\% (254) & 67.7 & 76.6 & 75.8 & 76.9 & \textbf{79.1} & $^{\ast}$79.0 & $^{\ast}$78.9 \\
     & E87.1: HypoNa & 5.8\% (720) & 63.0 & 68.0 & $^{\ast}$68.9 & 68.5 & $^{\ast}$69.6 & $^{\ast}$69.6 & \textbf{70.0} \\
     & D50-D64: Anemia & 23.7\% (2,931) & 64.0 & 70.8 & 70.8 & 71.2 & $^{\ast}$72.8 & \textbf{73.4} & $^{\ast}$72.9 \\
     & G47.33: OSA & 6.9\% (848) & 63.2 & 70.4 & 70.5 & $^{\ast}$71.3 & $^{\ast}$71.4 & $^{\ast}$72.0 & \textbf{72.2} \\
     & I26: PE & 1.3\% (163) & 67.7 & $^{\ast}$73.4 & 72.5 & $^{\ast}$73.0 & $^{\ast}$73.1 & $^{\ast}$73.6 & \textbf{75.1} \\
     & I27.2: PHT & 1.4\% (178) & 75.0 & 82.2 & 82.2 & 81.8 & 83.7 & $^{\ast}$84.4 & \textbf{85.5} \\
     & J44: COPD & 9.3\% (1,150) & 65.5 & 72.8 & 73.9 & 74.1 & $^{\ast}$75.4 & \textbf{75.7} & $^{\ast}$75.7 \\
     & J96.0: ARF & 2.0\% (250) & 74.3 & 82.0 & 80.2 & 78.4 & $^{\ast}$82.9 & 81.9 & \textbf{84.1} \\
     & N18.1-3: CKD1-3 & 4.3\% (533) & 63.8 & 69.4 & 71.1 & 71.0 & 72.2 & \textbf{74.2} & 72.5 \\
     & N18.4-5: CKD4-5 & 1.5\% (185) & 73.8 & 80.0 & 79.6 & 79.8 & $^{\ast}$81.6 & \textbf{82.2} & $^{\ast}$81.4 \\
     & N18.6: ESKD & 4.3\% (534) & 77.8 & 86.1 & 86.6 & 85.8 & $^{\ast}$88.0 & $^{\ast}$88.1 & \textbf{89.0} \\
     & K70/K74: FCL & 3.7\% (455) & 63.5 & 78.6 & 78.7 & 78.1 & $^{\ast}$81.6 & $^{\ast}$82.3 & \textbf{82.6} \\
     & E05: HyperThy & 0.6\% (69) & 46.8 & 47.7 & $^{\ast}$54.5 & \textbf{56.7} & $^{\ast}$55.5 & $^{\ast}$55.7 & 50.2 \\
     & \textcolor{gray}{E85 and I43: CA}$^{\dagger}$ & 0.0\% (6) & $^{\ast}$87.3 & $^{\ast}$82.9 & $^{\ast}$89.3 & $^{\ast}$72.9 & $^{\ast}$83.1 & $^{\ast}$87.8 & \textbf{90.9} \\
    \bottomrule
\end{tabular}

    \footnotetext{Linear probe evaluations. Random refers to a randomly initialized ECG encoder. Prevalence is the proportion of positives in the test set the metrics were computed on. The best model is \textbf{bold}; $\ast$ marks models that are not significantly different from the best per a paired patient-cluster bootstrap significance test (see Methods). By default, only conditions with at least 10 positive and negative cases are shown. $^{\dagger}$Cardiac amyloidosis is excluded from the MIMIC-EXP aggregate score due to low prevalence but shown here given the emerging interest in its detection.}
\end{table}

\cref{tab:label_level_main_auroc} shows results by report type at the level of individual conditions, providing further insight on the strengths of alignment to each source, and further demonstrating the benefit of multi-source alignment. First, we see that the random encoder is only in the not significantly different from best group for right ventricular hypertrophy (RVH), hypertrophic cardiomyopathy (HCM) and cardiac amyloidosis (CA).\footnote{Performance on these conditions might be explained by their very low prevalence in these datasets.} Therefore, this suggests that aligning to all of these report sources --- especially when done concurrently --- can significantly improve the quality of ECG-FM representations for the detection of all other conditions tested here, including the exploratory conditions in MIMIC-EXP, for which robust evidence is still emerging (see \cref{tab:review_ecg_outcomes}). Area under the precision-recall curve (AUPRC) results are reported in \cref{tab:label_level_auprc}.

It is also worth noting here that the ECG-FM pre-trained in \cref{fig:perf_profile_internal} that uses only the ECG machine reports (ECG-MR), which uses \methodname{}'s architecture and recipe (except for the multi-source objective), performs better than MERL \cite{liu2024zero}, D-BETA \cite{hung2025boosting} and MELP \cite{wang2025from} --- methods that were pre-trained with the same machine reports from the MIMIC-IV-ECG dataset \cite{gow2023mimic}. Considering the 98 label evaluation set that can be used to compare to these methods (the same set used in \cref{fig:perf_profile_external}), the ECG-MR method here scores an overall AUROC of 92.8 versus 88.8, 90.3 and 91.2 for MERL, D-BETA and MELP, respectively (see \cref{fig:perf_profile_ecg_mr_external}). This results from several other decisions beyond multi-source alignment that were made in the design of \methodname{} to improve its performance on ECG detection tasks. Briefly, these include the use of pre-processed embeddings from a strong clinical text foundation model during pre-training, the decision not to normalize inputs to preserve amplitude information for voltage criteria based diagnoses (e.g., left ventricular hypertrophy), and a different backbone architecture and pre-training recipe. These are described in detail in Methods, with an experimental ablation study in \cref{app:ablations}.

\section{Discussion}\label{sec:discussion}

Our experiments show that the clinical information used during pre-training meaningfully shapes the downstream capabilities of an ECG foundation model. Different report types provide complementary supervision, with ECG reports preferentially supporting conventional electrocardiographic diagnoses, while other clinical notes contribute information relevant to structural and broader disease phenotypes. By jointly learning from these sources, MS-ECG-FM produces representations that are consistently useful across a wider range of downstream tasks than models trained primarily from ECG machine reports (\cref{fig:perf_profile_external}). In particular, \methodname{} resolves the diagnostic blind-spots of prior methods, for example in the detection of structural heart disease (SHD) related conditions as assessed by the EchoNext-Mini dataset \cite{echonext_mini_2026,berger2026position}. These gains are achieved with fewer model parameters than D-BETA \cite{hung2025boosting} and MELP \cite{wang2025from} (38.28M parameters versus 63.20M and 62.55M, respectively, see \cref{tab:ecg_fm_sizes}), and substantially fewer pre-training ECGs than ECGFounder \cite{li_ecgfounder2025} (720k versus 10.7M). This suggests that supervision diversity may be an important axis of ECG-FM scaling alongside model and dataset size. 

The source-specific experiments provide insight into why multi-source supervision improves ECG representations (\cref{fig:perf_profile_internal}). Different forms of clinical documentation reflect the distinct questions that different diagnostic tests are designed to answer. Therefore, they provide complementary views of cardiovascular health. ECG reports primarily describe electrical abnormalities, whereas other clinical sources capture structural, functional, and broader systemic disease information that may not routinely appear in an ECG interpretation. Consistent with this, discharge summaries and chest X-ray reports provided particularly useful supervision on conditions that were less well supported by the ECG reports, including many structural heart disease endpoints, as well as exploratory endpoints such as electrolyte imbalances, pulmonary and metabolic conditions. The contribution of chest X-ray reports may reflect findings such as cardiomegaly, pulmonary edema, pneumonia or pleural effusion that provide indirect information about cardiac structure and pathophysiological status. Discharge summaries provide an even broader clinical view by synthesizing imaging, laboratory, and other diagnostic findings collected during an encounter. Echocardiography reports, despite their direct relevance to structural heart disease, provided weaker supervision in our experiments, which may partly reflect their substantially smaller sample size (81k notes from only 34k unique patients versus at least 331k notes from at least 102k patients for the other report types, see \cref{tab:mimic_reports}). 

Another important insight from this work is that ECG machine reports (ECG-MR) still provide useful supervision during pre-training, despite their high rates of error and correction by cardiologists \cite{Schlapfer2017,chiu2025}. While not as strong overall as the notes reviewed and edited by cardiologists (ECG-CR), they are the second strongest alignment source overall on these tasks (\cref{fig:perf_profile_internal}), providing useful supervision for conventional ECG-based detection tasks (e.g., arrhythmias and conduction disturbances). Inferior and anterior MI detection is an area where a high rate of cardiologist modification of machine-generated interpretation notes has been reported \cite{chiu2025}. Consistent with these findings, we observe that ECG-CR notes result in better classification performance than ECG-MR on inferior MI (94.4 versus 94.2 AUROC) and anterior MI (97.2 versus 96.9 AUROC, see \cref{tab:label_level_main_auroc}). However, these improvements are slight. We also see at the diagnostic group level that the two are tied on ischemia and ST/T changes (ISC/STTC, 92.3) and voltage criteria (VC, 91.7) conditions, and that ECG-MR is slightly stronger on structural heart disease (SHD, 81.5 versus 81.1) and exploratory MIMIC outcomes (MIMIC-EXP, 76.5 versus 76.4). Thus, while cardiologist-authored or reviewed ECG reports should be used where available, using ECG machine reports still delivers pre-training benefits, and our findings suggest they should be used concurrently. Future machine learning research could investigate why pre-training that aligns to noisy text sources still produces useful ECG representations.

The good performance demonstrated on conditions that are not typically detected by human ECG readers is also encouraging for the use of AI-enhanced ECG as a \emph{digital phenotyping} tool \cite{siontis2021artificial}. We see that \methodname{} is broadly capable of detecting echocardiographically verified structural heart disease endpoints: not just left ventricular systolic dysfunction (tested here as a binary left ventricular ejection fraction label), but also right-sided heart failure, valvular diseases (including aortic stenosis and regurgitation at all major valves), and pulmonary circulation \cite{Poterucha2025}. \methodname{}'s detection performance compares favorably to the official EchoNext-Mini baseline on these same conditions, where the baseline used only supervised learning on the downstream dataset (i.e., a pre-trained foundation model was not used to provide ECG representations) \cite{echonext_mini_2026}. Furthermore, we observed gains over a randomly initialized encoder in the detection of broad exploratory disease endpoints tested using ICD codes from the medical record (see MIMIC-EXP group in \cref{tab:label_level_main_auroc}). These include diastolic heart failure, dilated and other forms of cardiomyopathy (but not hypertrophic cardiomyopathy per ICD10-CM I42.1-2), obstructive sleep apnea, and multiple pulmonary, metabolic, and electrolyte imbalance codes. This corroborates the emerging evidence that ECG can give indications of the presence of even more conditions (as summarized in \cref{tab:review_ecg_outcomes}). These positive findings also hold for many conditions in reduced-lead configurations (see \cref{tab:label_level_lead_auroc,tab:label_level_lead_auprc}) which could facilitate preventative ECG-based screening in ambulatory settings (e.g., using wearables).

A related implication is that using the rich clinical supervision available around hospital 12-lead ECGs may be a useful model building strategy even when downstream deployment involves reduced-lead recordings. Clinical 12-lead ECGs are often collected within well-characterized encounters that include imaging, laboratory measurements, and clinical notes, whereas similarly rich supervision is difficult to obtain for ambulatory ECG data. \methodname{} performs well on many conditions when transferring its representations trained on 12-lead ECG data to reduced-lead configurations (\cref{tab:reduced_lead,tab:label_level_lead_auroc,tab:label_level_lead_auprc}). These findings suggest a practical strategy in which representations are first learned from well-characterized clinical 12-lead ECGs and then subsequently adapted to reduced-lead settings. However, because our reduced-lead evaluations were derived from clinical 12-lead recordings rather than natively acquired wearable or ambulatory ECGs, further validation in real-world reduced-lead datasets is needed.  

There are several limitations to this work. First, the MIMIC-IV-ECG pre-training distribution is skewed towards an older population (median age of 58 with interquartile range of 43–72) and a sicker population (inpatient or emergency room visits) from a single health system. Therefore, some conditions and states may be under- or over-represented in the training data which could impact performance when the model is deployed on different populations. Relatedly, while we evaluated on a broad set of downstream tasks across multiple diagnostic categories, some low prevalence but clinically acute conditions are not adequately represented in the available evaluation datasets. Furthermore, while we evaluated on several out-of-distribution ECG datasets to test the generalization capabilities of \methodname{} to different hospital systems, performance in new systems may still be attenuated depending on how stark the distribution shift is (e.g., if there are large differences in ECG collection or condition diagnosis protocols). Finally, foundation models are not inherently interpretable, meaning they are not the full solution to the adoption gap. 

Beyond addressing these limitations, future work can also consider alternative objectives for multi-source alignment and alignment to other clinical note types (e.g., coronary angiogram reports). The evaluation of \methodname{} on real-world ambulatory reduced-lead datasets would also be valuable. Furthermore, ECG question answering (QA) systems are in the early stages of development with the goal of providing an interactive natural language interface to query what phenomena are present in an ECG waveform \cite{lan2025gem}. These systems are built using a pre-trained ECG encoder and it would be interesting if \methodname{} is able to support more accurate or broader QA given the rich clinical note sources that were used to pre-train it. 

In conclusion, we introduce \methodname{} --- a new ECG foundation model that achieves state-of-the-art detection performance on the full spectrum of diseases where there is strong or emerging evidence of ECG detection capabilities. The results show that the clinical information used during pre-training is an important determinant of what an ECG foundation model learns. We also provide insights on which clinical notes are most informative for the detection of different conditions. Finally, we propose an extended evaluation framework for ECG-FMs, which provides clearer summaries of model capabilities, which in turn can be used to prioritize model improvements. We hope this work supports the research community to build stronger ECG-FMs and to evaluate them more thoroughly, so that the next generation can deliver tangible benefits in real-world clinical settings. Hospital systems already store multi-modal notes at far greater scale than we had access to for this work: using these to pre-train a larger \methodname{} could improve performance even beyond the gains shown here.

\section{Methods}\label{sec:method}

\subsection{\methodname{} data, design and pre-training protocol}

\subsubsection*{Pre-training with MIMIC-IV ECG and clinical report dataset} \label{sec:pretraining_data} 

We use the publicly available MIMIC-IV-ECG dataset to pre-train \methodname{} \citep{gow2023mimic}. This dataset is the largest publicly available ECG dataset with paired clinical notes and it is also used for pre-training by several state-of-the-art ECG foundation models \cite{liu2024zero,hung2025boosting,wang2025from}, allowing for controlled comparison of \methodname{}. MIMIC-IV-ECG provides 800,035 standard 10\,s 12-lead ECG recordings from 161,352 patients. We use them at the native 500\,Hz sampling rate and the only pre-processing step is to replace missing samples with zero. The demographics of the cohort are median age of 58 (interquartile range: 43–72) and biological sex of 51.8\% female, 47.1\% male and 1.1\% unknown.

MIMIC-IV-ECG provides 800,034 ECG machine interpretation reports after cleaning (ECG-MR). It also contains several types of clinician-authored interpretation reports from ECG recordings (ECG-CR; 490,780), ECHO scans (ECHO-CR; 81,821) and chest X-rays (ChXR-RR; 715,465), as well as the overall discharge summary for a hospital stay (DS; 331,793). More details on each report type and how we preprocess them are in \cref{app:reports_info}. For pre-training, we pair reports to ECG recordings with report-dependent logic: ECG reports have a key to ECG recordings; discharge summaries can be linked to all ECG recordings within a hospital visit using a visit identifier; ECHO and ChXR reports are linked by taking the report closest in time to the ECG recording within a 7-day window.  

\subsubsection*{ECG waveform encoder architecture} \label{sec:architecture} 

Three ECG-specific considerations motivated the design of the transformer-based waveform encoder. First, given ECG leads represent cardiac electrical activity along different axes, and meaningful phenomena are frequently lead or location-specific (e.g., myocardial infarction), lead-specific tokens are created. Second, given well-characterized 12-lead ECG data is more abundant for pre-training, but deployment contexts (e.g., wearable) may only have access to reduced-lead subsets (e.g., Lead-I only), the architecture was explicitly designed to support easy transfer to arbitrary reduced-lead subsets. Third, given fine-grained, millisecond-scale local morphology (e.g., QRS widths and ST deviations) is important for ECG interpretation, depth was added to the ECG tokenizer.   

The ECG waveform is first tokenized by a deep, lead-level tokenizer with multiple convolutional layers loosely inspired by \citet{hassani2021escaping}, where the depth is included to promote learning richer millisecond-scale features. While lead-level tokens are produced, the tokenizer weights are shared between leads. Given a 10\,s 12-lead ECG recording and a tokenizer patch size of $P\!=\!250$ samples (0.5\,s at 500\,Hz), the tokenizer produces $12\!\times\!20\!=\!240$ tokens. The tokens are then passed through multiple transformer layers to learn longer-range interactions across time and leads. Temporal order is added using 1D sinusoidal positional encoding, while lead identity is represented using learnable lead embeddings similar to \citet{na2024guiding}. During contrastive pre-training, the final ECG representation is obtained by global average pooling (GAP) over tokens. Downstream GAP is also used for the \emph{linear probe} evaluations. Further details on the architecture are in \cref{app:architecture_details}.

Importantly, as several important ECG diagnoses also depend on absolute signal amplitude rather than waveform morphology alone, the ECG waveform is \emph{not} normalized in \methodname{}. A clear example is left ventricular hypertrophy (LVH), which is identified using voltage criteria --- e.g., Sokolow-Lyon or Cornell --- that sum R- and S-wave amplitudes in specified leads and compare them against a fixed voltage threshold. This information is easily destroyed by many normalization schemes: for example, per-lead or per-recording normalization (e.g., z-scoring, min-max scaling) rescales each waveform by its own statistic, discarding the scale that voltage criteria depend on. Thus, \methodname{} does not normalize the input waveform. Furthermore, to balance training stability (e.g., vanishing gradients) while preserving amplitude information, batch normalization is used after each convolution in the tokenizer \cite{batchnorm2015}. A per-sample normalization in the tokenizer (e.g., layer normalization \cite{layernorm2016}) would not have this property. \cref{tab:other_components_ablation} ablates several of these components.

\subsubsection*{Contrastive multi-source alignment objective} \label{sec:core_loss} 

\methodname{} is pre-trained by contrastive alignment to multiple types of clinical report. It uses a contrastive objective based on the InfoNCE loss and forms multimodal contrastive pairs composed of the ECG waveform and a clinical report \cite{oord2018representation,radford2021learning}. For each instance in any given batch, the report that the waveform is paired with is stochastically chosen from the set of available reports for that instance. The stochastic selection of a paired report was designed as a form of regularization to promote the learning of a broader feature set. The reports are sparsely available with only an arbitrary subset available for a given ECG. Furthermore, some report types will share more information with the paired ECG than others: ECG machine interpretations are generated exclusively from and describe only the ECG and are thus tightly coupled, while chest X-rays or discharge summaries contain diverse information of which only a fraction may be shared with the ECG. Therefore, if all report types are available simultaneously, there is a risk that an ECG encoder only learns to use simpler features to solve the instance-wise discrimination objective. By stochastically selecting reports in each batch, we posit that the model is incentivized to learn a broader set of discriminative features to improve its downstream performance, especially on more difficult tasks where discriminative ECG waveform features may be more subtle (e.g., the MIMIC-EXP group). \cref{tab:ms_objective_ablation} reports an ablation of alternative multi-source alignment objectives with empirical findings to support this design.

Formally, let $\mathcal{R}$ be the set of all report types and $\mathcal{R}_i \subseteq \mathcal{R}$ denote the subset of report types available for instance $i$, with $y_i^{(r)}$ the text of report type $r \in \mathcal{R}_i$. Each time the instance $i$ is sampled during pre-training, a single report type is drawn uniformly at random from those available, $r_i \sim \mathrm{Uniform}(\mathcal{R}_i)$. The ECG waveform is encoded by the backbone $h_i^{\mathrm{wf}} = f_{\theta}(s_i)$ and projected by an ECG projector head $z_i^{\mathrm{wf}} = \pi_{\phi}^{\mathrm{wf}}\bigl(h_i^{\mathrm{wf}}\bigr)$. The sampled report is encoded by a frozen text encoder $g$ (see next section) and projected by a text projector head $z_i^{\mathrm{text}} = \pi_{\psi}^{\mathrm{text}}\bigl(g\bigl(y_i^{(r_i)}\bigr)\bigr)$. Mini-batches $\mathcal{B}$ are stochastically sampled with $|\mathcal{B}|\!=\!K$ and the multi-source (ms) contrastive objective is
\begin{equation}
    L^{\text{ms}} = - \frac{1}{K} \sum_{i \in \mathcal{B}} 
    \log
    \frac{\exp\bigl(\mathrm{sim}(z_i^{\mathrm{wf}},\, z_i^{\mathrm{text}}) / \tau\bigr)}
    {\sum_{j \in \mathcal{B}} \exp\bigl(\mathrm{sim}(z_i^{\mathrm{wf}},\, z_j^{\mathrm{text}}) / \tau\bigr)} ,
\end{equation}
where $\mathrm{sim}(\cdot,\cdot)$ denotes cosine similarity and the temperature, $\tau$, is a learnable parameter as in \cite{radford2021learning}. 

\subsubsection*{Alignment to pre-computed text embeddings to stabilize and simplify computation} \label{sec:embeddings}

A key feature of \methodname{}'s contrastive learning formulation is its use of pre-computed text embeddings for contrastive alignment, rather than jointly learning a text encoder with the ECG encoder during pre-training as in contemporary methods that also use the MIMIC-IV-ECG reports corpus \cite{liu2024zero,hung2025boosting,wang2025from}. By default, MedGemma-27B is used as the text encoder $g^{(r)}$ for all report modalities $r \in \mathcal{R}$ \cite{sellergren2025medgemma}, where report-level text embeddings are extracted in an offline pre-processing step. Details on MedGemma-27B usage are in \cref{app:method_details}. 

We motivate the use of frozen text embeddings to train \methodname{} as follows. First, clinical text encoders such as MedGemma-27B are trained on orders of magnitude more data with broader distributional coverage than the MIMIC-IV reports corpus. Prior work in the image-text domain has shown the benefits of this approach \citep{zhai2022lit}, finding that locking a strong pre-trained encoder in multimodal contrastive pre-training benefits downstream accuracy (especially out-of-distribution). Second, locking the text encoder freezes the disease semantics already encoded in the reports. Given the diseases are described explicitly in the text and MedGemma-27B is already effective at modeling medical language, the pre-training problem reduces to aligning the ECG encoder to fixed text representations, avoiding instability or catastrophic forgetting that could arise from jointly updating both encoders. Finally, there are significant computational benefits to computing text embeddings offline. Fewer activations, parameters and gradients need to be held in memory for text tokens while pre-training the ECG encoder, reducing memory footprint and runtime. These benefits are amplified when multiple long report types must be embedded simultaneously. \cref{tab:mimic_reports} summarizes the median and max token length by report type. Notably, ECHO, chest X-ray and discharge reports would cause significant burden in an online text encoder setup, as they can reach sizes of 1,128, 761 and 15,039 tokens in MIMIC-IV, respectively.

\subsubsection*{Pre-training recipe} \label{sec:pretraining_recipe}

\methodname{} is pre-trained on only the stratified train folds (i.e., folds 0-17) proposed in MIMIC-IV-ECG-Ext-ICD, and uses the validation fold (i.e., fold 18) for model selection \cite{strodthoff2024data, strodthoff2024prospects}. The test fold (i.e., fold 19) is fully held out from pre-training. An unsupervised effective rank metric (\emph{RankMe}) is computed on the backbone embeddings of MIMIC-IV validation set ECGs and is used to select a pre-trained encoder checkpoint for downstream evaluation \cite{pmlr-v202-garrido23a}. This quantifies the entropy of the singular value distribution of the embeddings -- a measure of their effective dimensionality -- and has been found to be a useful proxy for downstream performance. In particular, it is useful for identifying when representations collapse to a degenerate form \cite{pmlr-v202-garrido23a}. Importantly, this metric only requires access to the \emph{in-distribution} pre-training dataset. This differs from recent works that use 100\% of MIMIC-IV ECGs for pre-training, and then select a pre-trained model using zero-shot performance on the validation sets of the \emph{out-of-distribution} (OOD) datasets \cite{liu2024zero,wang2025from}. We chose our approach as it provides a stronger generalization test: no part of the pre-training process is influenced by OOD data.

\methodname{} is pre-trained for 75,000 optimization steps. Random lead masking (RLM) \cite{oh2022lead} and a stochastic time-shift are used for augmentation, with the former introduced to promote stronger reduced-lead performance \cite{oh2022lead}. Full pre-training settings are reported in \cref{app:pretraining_settings}. The single-source methods presented in \cref{fig:perf_profile_internal} and \cref{tab:label_level_main_auroc} use the same architecture and settings as \methodname{}; their only distinction is that they are aligned to just one report type during pre-training, enabling a controlled comparison on the utility of different report types for different downstream condition detection tasks.

\subsection{Extended evaluation of ECG foundation models} \label{sec:evaluation_protocol}

Motivated by the goal of developing a more universal ECG-FM --- where universality is defined by the utility of its representations in broad downstream tasks --- we carefully examined existing ECG-FM benchmarking practices and propose extensions to provide a clearer and more comprehensive assessment of the capabilities of ECG-FM representations.

\subsubsection*{Downstream evaluation protocol and metrics for evaluation} 

We use supervised \emph{linear probing} to evaluate \methodname{} and baseline ECG-FMs. This refers to training a linear classifier on frozen ECG-FM backbone representations (or embeddings). Linear probing is the leading paradigm when the goal is to compare the quality of representations between ECG-FMs with different backbone architectures because it is simple to optimize a single-layer linear classifier on frozen embeddings in a way that is comparable across models. By contrast, fine-tuning foundation models can be challenging given their size and differing architectures: the optimal set of fine-tuning hyperparameters are not known a priori, and it can be compute-intensive to search for the optimal set for each model. We use the standard linear probe settings from the ECG-FM literature, including the same dataset split files to partition instances into training, validation and test sets, and report them in \cref{tab:exp_settings_linear_probe} \cite{wang2025from}. To make evaluation exactly comparable between ECG-FMs, as well as to enable paired statistical tests, we rerun evaluation on the baseline models using their publicly shared pre-trained weights through a seeded pipeline.\footnote{Results on the same benchmarks reported by the authors are summarized in \cref{tab:linear_probe_bms_literature_reported}, with scores broadly aligned with those we generate when re-running evaluation in \cref{tab:main_benchmark}.} The area under the receiver operating characteristic curve (AUROC) is the primary evaluation metric. All labels are binary. When aggregating performance across tasks we macro-average the AUROC. We also report the area under the precision-recall curve (AUPRC) for performance at the level of individual tasks in \cref{app:more_results}.

\subsubsection*{Extended downstream datasets containing the full range of conditions ECG can detect} 

We use standard ECG benchmark datasets \cite{liu2024zero,hung2025boosting,wang2025from}. Furthermore, we incorporate two new benchmark datasets that cover a broader set of diseases and conditions not represented in existing benchmarks but for which there is at least exploratory evidence that ECG can detect. Datasets are marked out-of-distribution (OOD) or in-distribution (ID) relative to the MIMIC-IV-ECG pre-training data.\footnote{In the ID dataset, patients are disjoint from the pre-training and training patients.} The OOD datasets provide a test of how well the ECG-FMs generalize to different hospital systems, while the ID dataset facilitates testing on broader disease categories. The coverage of conditions by dataset is in \cref{tab:review_downstream_datasets}.

\begin{enumerate}

\item \textbf{PTB-XL (OOD):} 21,799 12-lead ECGs annotated with statements across 71 diagnostic, rhythm and morphology classes, where the test set was reviewed and modified by cardiologists \cite{wagner2020ptb}. PTB-XL is often used as a benchmark via four multi-label classification tasks --- super(-diagnostic), sub(-diagnostic), rhythm and form --- where the macro-averaged AUROC across labels is reported.

\item \textbf{CPSC2018 (OOD):} is also a commonly used benchmark dataset \cite{liu2018open}. It contains 6,877 12-lead ECGs with labels for rhythm, conduction and morphology abnormalities.

\item \textbf{CSN (OOD):} is another commonly used benchmark dataset \cite{zheng2022large,zheng2020optimal}. It contains 23,026\footnote{Unlabeled ECGs are excluded.} 12-lead ECGs with labels for rhythm, conduction and morphology abnormalities, as well as a small number of labels from other conditions (e.g., myocardial infarctions).

\item \textbf{EchoNext-Mini (OOD):} 100,000 12-lead ECGs paired with structural heart disease (SHD) labels from paired echocardiographic recordings \cite{elias2025echonext,echonext_mini_2026}. This enables extensive evaluation on SHD conditions --- an area where there is promising clinical evidence on the utility of ECG for detection but which was under-supported by contemporary ML benchmarks \cite{attia2021application,siontis2021artificial}. Labels include left ventricular ejection fractions, right ventricular systolic dysfunction, aortic stenosis and valvular regurgitation severities, and other metrics quantifying heart function or morphology. Consistent with the original work \cite{elias2025echonext,echonext_mini_2026}, we use the released patient-disjoint splits, in which the validation and test sets contain only the most recent ECG per patient, and the training set contains all ECGs from each patient.

\item \textbf{MIMIC-IV-ECG-Ext-ICD (ID):} 468,005 ECGs from MIMIC-IV linked to hospital discharge ICD10-CM codes \cite{strodthoff2024data}. We parse the dataset to create binary ICD10 labels. We include core ischemic and structural heart disease labels, as well as important conditions where there is at least exploratory clinical evidence of detection using ECG (see \cref{tab:review_ecg_outcomes}) but which are not covered by the other datasets (see \cref{tab:review_downstream_datasets}). For example, we include non-ST elevation myocardial infarction (NSTEMI) which is highly acute and important to diagnose and treat expeditiously \cite{al2023machine}. These labels are at the visit-level, and we use only the first ECG per visit in the evaluation set.

\end{enumerate}

\subsubsection*{More informative evaluation of ECG foundation models using diagnostic task groupings and performance profiles}

The common practice in the ECG-FM literature is to report the macro-averaged AUROC across all labels in the dataset for datasets 1-3 (with PTB-XL at the super-diagnostic, sub-diagnostic, rhythm and form levels). We also report this table given its status as the de facto benchmark view for ECG-FM comparison \cite{liu2024zero,hung2025boosting,wang2025from}, adding EchoNext-Mini as an additional dataset. However, this method of aggregation is crude as: (i) it obscures valuable information about the strengths and weaknesses of ECG-FMs by condition type as it averages over unrelated conditions that may manifest very differently in the ECG signal; and, (ii) only reporting the mean across many labels obscures the relative distribution of performance between models (e.g., a method may have a higher mean score because it is consistently better than another method on many labels \emph{or} because it scores very highly on a few isolated labels). Thus, we propose two alternative and complementary ways to report and compare ECG-FM performance which provide greater insight to clinical and technical readers.

First, we use datasets 1, 4 and 5 to group labels into clinically intuitive diagnostic categories --- e.g., all arrhythmias, myocardial infarctions, structural heart disease conditions.\footnote{We select these datasets due to the higher quality of their labels, which are either reviewed by clinicians or derived from reports written by clinicians.} See \cref{tab:downstream_diagnosis_groups} for full documentation on these groups. This facilitates the identification of blind-spots in ECG-FMs by logical diagnostic groupings.

Second, we propose the visualization of \emph{performance profiles} for ECG-FMs. Pioneered first in the optimization \cite{dolan2002benchmarking} and later in the reinforcement learning literature \cite{Agarwal2021}, performance profiles are a complementary cumulative distribution function (or tail distribution) of a method's performance over a set of tasks. In the context of ECG-FMs, we propose that each task is a different diagnostic label and the performance metric is the AUROC. Thus, performance profiles allow us to concisely visualize differences in the performance distributions of different ECG-FMs, in a way that is robust to outliers. Formally, for each ECG-FM, its performance profile is defined by 
\begin{equation}
    \hat{F}(\tau) = \frac{1}{M}\sum_{m=1}^{M}\mathbf{1}[x_m > \tau],
\end{equation}

\noindent where there are $M$ different labels, $x_m$ is the AUROC score for label $m$, $\mathbf{1}$ is the indicator function, and $\tau$ is an arbitrary score value, where $\tau \in [0.5, 1]$. A clear benefit of performance profiles is that they give a notion of \emph{universality}. If the curve for one model is always above that of another model, it is said to stochastically dominate over the set of tasks, and thus it is a \emph{universally} better model. Further details on performance profiles, including how we compute the confidence bands, are in \cref{app:perf_profiles}. 

\subsubsection*{Formally testing differences in foundation model performance}

Finally, to address another limitation in contemporary ECG-FM benchmarking practices, we formally test whether differences in performance between models are statistically significant. We use a paired, patient-level (cluster) bootstrap on the test set of each evaluation dataset, where bootstrap resamples are created by resampling patients with replacement (some patients contribute more than one ECG). The resampled patient indices within each dataset are shared across all labels and all models to enable paired comparisons. We draw $N_{\text{boot}} = 2{,}000$ bootstrap resamples. Each label and model therefore has a bootstrap distribution of $N_{\text{boot}}$ replicates.\footnote{Occasionally, a resample is degenerate because one of the labels has no positive examples. This reduces the effective $N^{\text{eff}}_{\text{boot}}$ but does not invalidate the inference. See \cref{app:statistical_testing}.} To compare two models on a label, we difference their scores within each resample and conclude a significant difference at the $\alpha = 0.05$ level if the 95\% percentile interval of these paired differences excludes zero. Aggregate comparisons --- which we make for AUROC only --- apply the same test after first macro-averaging AUROC across labels within each resample. 

To avoid visual overload when presenting the results of the significance tests, we first mark the strongest model with a special character or in bold. We then highlight methods that are not significantly different from the highest performing method with a distinct special character. Further details on our testing procedure are in \cref{app:statistical_testing}.

\backmatter

\bmhead{Supplementary information} We provide supplementary information in \cref{app:review_ecg_fms,app:review_ecg_clinical,app:datasets,app:more_results,app:ablations,app:reports_info,app:method_details,app:extended_evaluation}.

\bmhead{Data availability} No proprietary data was used in this work, and no new datasets were generated by the authors. All data used during pre-training and evaluation are available from the curators of these datasets to researchers who complete the required human subjects research training and sign the data use agreement for each dataset.

\bmhead{Acknowledgements} The authors wish to thank the following individuals for thoughtful discussion and guidance in support of this work: Achille Nazaret and Matthew Jörke.

\bmhead{Author contributions} R.L., I.C., K.V., A.C.M. and J.F. conceived the initial study. R.L. led the data processing and algorithm development work, with support from I.C., K.V., A.C.M., J.F., A.M. and S.A. All authors including K.J.R., F.M. and G.S. made substantial conceptual and intellectual contributions to the research. R.L. wrote the initial manuscript, which all authors reviewed.

\bmhead{Competing interests} All authors were employed by Apple, Inc. and/or owned Apple, Inc. stock during the preparation of this work. J.F. is currently an employee of Ouraring Inc. His contributions to this work, as well as the statements and opinions expressed, are solely the responsibility of the author and do not represent the official views of Ouraring Inc. J.F. was an employee at Apple Inc. at the time of his contribution to this work, which predates his transition to Ouraring Inc.

\clearpage

\begin{appendices}
\crefalias{section}{appendix}
\crefalias{subsection}{appendix}

\counterwithin*{figure}{section}
\counterwithin*{table}{section}

\section{Review of ECG foundation models} \label{app:review_ecg_fms}

\subsection{Contemporary ECG foundation model methods}

\cref{tab:review_ecg_fms} summarizes contemporary ECG foundation models, providing relevant details about their implementation and usage. Several use the same MIMIC-IV-ECG data that our method is trained on, while others are trained on other ECG datasets. Note: none of those using the MIMIC-IV-ECG data make use of the MIMIC-IV clinical notes during pre-training beyond the ECG machine interpretation reports.

\begin{table}[h]
  \centering
    \setlength\tabcolsep{1pt}
    \fontsize{6}{7}\selectfont
    \caption{Review of prominent electrocardiogram foundation models from 2024-2025. \textsuperscript{$\dagger$} indicates models chosen for our experiments due to their strong benchmark performance. \textsuperscript{$\ddagger$} indicates MIMIC-IV-ECG was used in pre-training pooled with other datasets.}
    \label{tab:review_ecg_fms}
    \begin{tabular}{
    >{\raggedright\arraybackslash}p{0.1\textwidth}
    >{\raggedright\arraybackslash}p{0.45\textwidth}
    >{\raggedleft\arraybackslash}p{0.1\textwidth}
    >{\centering\arraybackslash}p{0.08\textwidth}
    >{\centering\arraybackslash}p{0.085\textwidth}
    >{\centering\arraybackslash}p{0.085\textwidth}
    >{\centering\arraybackslash}p{0.05\textwidth}
    >{\centering\arraybackslash}p{0.05\textwidth}
}
    \toprule
    & & &
    \multicolumn{3}{c}{\textbf{MIMIC-IV data usage}} & 
    \multicolumn{2}{c}{\textbf{Available}}
    \\
    \cmidrule(lr){4-6} \cmidrule(lr){7-8}
    \textbf{Model} &
    \textbf{Pre-training objective(s)} &
    \textbf{\# Pretrain samples} &
    \textbf{ECG waveforms} &
    \textbf{Machine reports} &
    \textbf{Clinical notes} & 
    \textbf{Code} &
    \textbf{Weights}
    \\
    \midrule

    ST-MEM \cite{na2024guiding} &
    Tokenizes 12-lead ECG recordings at spatio-temporal level (i.e., patches within individual leads). Then uses a modified masked-autoencoder (MAE) reconstruction objective. &
    345,779 &
    \xmark & \xmark & \xmark & \cmark & \cmark
    \\ \midrule
    
    \textbf{MERL}\textsuperscript{$\dagger$} \cite{liu2024zero} &
    Multimodal contrastive ECG waveform \& machine report objective (\emph{InfoNCE}). Auxiliary unimodal contrastive ECG w/latent augmentation (\emph{InfoNCE}). &
    800,035 &
    \cmark & \cmark & \xmark & \cmark & \cmark
    \\ \midrule
    
    KED \cite{tian2024foundation} &
    Multimodal contrastive ECG waveform \& LLM-augmented machine report objective, with concurrent use of labels from machine report. &
    800,035 &
    \cmark & \cmark & \xmark & \cmark & \cmark
    \\ \midrule

    ECG-JEPA \cite{kim2024learning} &
    Uses Joint-Embedding Predictive Architecture (\textit{JEPA}) self-distillation objective where masked latent representations of the ECG from the \emph{teacher} network are predicted by the \emph{student} network. &
    174,140 &
    \xmark & \xmark & \xmark & \cmark & \cmark
    \\ \midrule
    
    \textbf{ECG Founder}\textsuperscript{$\dagger$} \cite{li_ecgfounder2025} &
    Supervised multi-label classification with labels sourced from an ECG-machine analysis program. &
    10,771,552 &
    \xmark & \xmark & \xmark & \cmark & \cmark
    \\ \midrule
    
    ECG-FM \cite{mckeen2025ecg} &
    Contrastive objective that combines \textit{wav2vec 2.0} quantized contrastive loss and diversity loss with \emph{CMSC} from \emph{CLOCS}. &
    873,706 &
    \cmark\textsuperscript{$\ddagger$} & \xmark & \xmark & \cmark & \cmark
    \\ \midrule

    HuBERT-ECG \cite{Coppola2025} &
    Follows the \textit{HuBERT} approach to SSL where latent representations are assigned to clusters (computed offline at each iteration) and then cluster IDs are predicted from masked frames and optimized by cross-entropy loss. &
    9,100,000 &
    \cmark\textsuperscript{$\ddagger$} & \xmark & \xmark & \cmark & \cmark
    \\ \midrule

    HeartLang \cite{jin2025reading} &
    A QRS-tokenizer is used, which detects the locations of QRS complexes and then segments the ECG into heartbeat-centered patches by lead. Two objectives are used to train the model: a vector-quantized \emph{beat-level} reconstruction task and a masked ECG \emph{sentence-level} prediction task.
    &
    800,035 &
    \cmark & \xmark & \xmark & \cmark & \cmark
    \\ \midrule

    \textbf{D-BETA}\textsuperscript{$\dagger$} \cite{hung2025boosting} &
    Hybrid multimodal contrastive ECG waveform \& machine report objective (\emph{SigLIP}), and a masked autoencoder loss on ECG and text encoders. &
    800,035 &
    \cmark & \cmark & \xmark & \cmark & \cmark
    \\ \midrule
 
    \textbf{MELP}\textsuperscript{$\dagger$} \cite{wang2025from} &
    Captioning multimodal loss at \emph{token-level} and multimodal contrastive ECG waveform and machine report objectives at the \emph{beat} and \emph{rhythm-level}. &
    800,035 &
    \cmark & \cmark & \xmark & \cmark & \cmark
    \\

\bottomrule
\end{tabular} 
\end{table}

\subsection{Reported foundation model benchmark results in original literature} \label{app:results_literature_source}

We reran evaluation on ECG-FM baselines for our main results in \cref{tab:main_benchmark} and the other main paper figures to ensure (i) the exact same evaluation protocol is used for each method, and (ii) paired bootstrap tests can be performed (which require access to instance-level predictions). 

For completeness, we report performance reported by the authors of these methods on the same benchmarks in \cref{tab:linear_probe_bms_literature_reported}. These are derived from the following sources:
\begin{itemize}
    \item ST-MEM from \cite{na2024guiding}
    \item HeartLang from Table 1 in \cite{jin2025reading}
    \item MERL from Table 2 in \cite{liu2024zero}
    \item D-BETA from Table 2 in \cite{hung2025boosting}
    \item MELP from Table 2 in \cite{wang2025from}
    \item ECGFounder from Table S13 in \cite{li_ecgfounder2025} (ArXiv [v4] Mon, 4 Aug 2025 10:31:10 UTC (7,335 KB))
\end{itemize}

Note: we report a systematic improvement in accuracy for all methods on the PTBXL-Sub dataset because we fixed an error in the benchmarking files and code used by most prior work (e.g., \cite{liu2024zero,hung2025boosting,wang2025from}). Specifically, we found that the column order for 3 codes --- AVB, RAO/RAE, ISC --- was different in the train split file from the order in the validation and test split files, and the columns were accessed by index in the evaluation scripts. By changing the access pattern to look up labels by name, the performance on these three codes is substantially improved, as now the label definition is consistent between training and evaluation.   

\begin{table}[h]
    \centering
    \setlength\tabcolsep{1pt}
    \fontsize{5}{6}\selectfont
    \caption{Linear probe performance of contemporary ECG foundation models as reported in their original papers. Results are shown for different percentages of available training labels (1\%, 10\%, 100\%).}
    \label{tab:linear_probe_bms_literature_reported}
    \begin{tabular}{llccc|ccc|ccc|ccc|ccc|ccc}
    \toprule
    \multirow{2}{*}{\textbf{Methods}} & & \multicolumn{3}{c}{\textbf{PTBXL-Super}} & \multicolumn{3}{c}{\textbf{PTBXL-Sub}} & \multicolumn{3}{c}{\textbf{PTBXL-Form}} & \multicolumn{3}{c}{\textbf{PTBXL-Rhythm}} & \multicolumn{3}{c}{\textbf{CPSC2018}} & \multicolumn{3}{c}{\textbf{CSN}} \\
    \cmidrule{3-5} \cmidrule(lr){6-8} \cmidrule(lr){9-11} \cmidrule(lr){12-14} \cmidrule(lr){15-17} \cmidrule(lr){18-20}
    & & \textbf{1\%} & \textbf{10\%} & \textbf{100\%} & \textbf{1\%} & \textbf{10\%} & \textbf{100\%} & \textbf{1\%} & \textbf{10\%} & \textbf{100\%} & \textbf{1\%} & \textbf{10\%} & \textbf{100\%} & \textbf{1\%} & \textbf{10\%} & \textbf{100\%} & \textbf{1\%} & \textbf{10\%} & \textbf{100\%} \\
    \midrule
    ST-MEM~\cite{na2024guiding} && 61.12 & 66.87 & 71.36 & 54.12 & 57.86 & 63.59 & 55.71 & 59.99 & 66.07 & 51.12 & 65.44 & 74.85 & 56.69 & 63.32 & 70.39 & 59.77 & 66.87 & 71.36 \\
    HeartLang \cite{jin2025reading} && 78.94 & 85.59 & 87.52 & 64.68 & 79.34 & 88.91 & 58.70 & 63.99 & 80.23 & 62.08 & 76.22 & 90.34 & 60.44 & 66.26 & 77.87 & 57.94 & 68.93 & 82.49 \\
    MERL~\cite{liu2024zero} && 82.39 & 86.27 & 88.67 & 64.90 & 80.56 & 84.72 & 58.26 & 72.43 & 79.65 & 53.33 & 82.88 & 88.34 & 70.33 & 85.32 & 90.57 & 66.60 & 82.74 & 87.95 \\
    D-BETA~\cite{hung2025boosting} && 83.15 & 88.36 & 90.11 & 77.74 & 82.92 & 85.15 & 70.10 & 78.91 & 83.98 & 86.61 & 92.83 & 96.71 & 85.46 & 91.35 & 94.92 & 80.04 & 87.36 & 90.71 \\
    MELP~\cite{wang2025from} && 85.82 & 87.61 & 87.87 & 79.22 & 84.40 & 87.46 & 63.41 & 76.71 & 83.30 & 88.83 & 94.65 & 96.91 & 88.54 & 91.75 & 94.32 & 78.25 & 84.83 & 90.17 \\
    ECGFounder~\cite{li_ecgfounder2025} && 79.65 & 87.34 & 91.20 & 75.72 & 81.73 & 88.03 & 61.69 & 68.76 & 82.52 & 70.05 & 88.10 & 93.61 & 64.21 & 79.65 & 83.18 & 64.68 & 75.89 & 84.35 \\
    \bottomrule
\end{tabular}

    \footnotetext{Evaluation metric is AUROC: area under the receiver operating characteristic curve.}
\end{table}

\subsection{Comparison of ECG foundation model capacity} \label{app:ecg_fm_sizes}

\cref{tab:ecg_fm_sizes} reports the number of ECG encoder parameters used to produce the frozen embeddings in our linear-probe evaluations. Counts include only the modules on that forward path: text encoders, unused self-supervised heads (e.g., vector quantizers), and classification heads are excluded. For transformer-based encoders we also report the tokenizer versus transformer split; MELP and D-BETA include a small additional pooling or projection module at probe time.

\begin{table}[h]
  \centering
    \setlength\tabcolsep{2pt}
    \fontsize{7}{8}\selectfont
    \caption{Number of ECG encoder parameters used to produce frozen linear-probe embeddings. Text encoders and unused pre-training or classification heads are excluded. For transformer-based models, tokenizer and transformer counts are listed below the total; MELP and D-BETA additionally use a pooling or projection module at probe time. \methodname{}'s default backbone is \underline{underlined}.}
    \label{tab:ecg_fm_sizes}
    \begin{tabular}{
    >{\raggedright\arraybackslash}p{0.26\textwidth}
    >{\raggedright\arraybackslash}p{0.70\textwidth}
}
\toprule
\textbf{ECG-FM} & \textbf{Number of ECG encoder parameters} \\
\midrule
\multicolumn{2}{l}{\textbf{Baselines}} \\ \cmidrule[0.2pt]{1-2}
MERL \cite{liu2024zero} &
$5.47$M\newline
\textbullet\ tokenizer: $0.12$M\newline
\textbullet\ transformer: $5.35$M \\
D-BETA \cite{hung2025boosting} &
$63.20$M\newline
\textbullet\ tokenizer: $0.60$M\newline
\textbullet\ transformer: $61.42$M\newline
\textbullet\ class embedding, projection, and pooler: $1.18$M \\
MELP \cite{wang2025from} &
$62.55$M\newline
\textbullet\ tokenizer: $0.60$M\newline
\textbullet\ transformer: $61.42$M\newline
\textbullet\ attentional pooler: $0.53$M \\
ECGFounder \cite{li_ecgfounder2025} &
$30.67$M \\
\midrule

\multicolumn{2}{l}{\textbf{\methodname{} ECGViT}} \\ \cmidrule[0.2pt]{1-2}
Tiny &
$5.40$M\newline
\textbullet\ tokenizer: $0.06$M\newline
\textbullet\ transformer: $5.34$M \\
Small &
$21.55$M\newline
\textbullet\ tokenizer: $0.25$M\newline
\textbullet\ transformer: $21.30$M \\
\underline{Middle} &
$38.28$M\newline
\textbullet\ tokenizer: $0.45$M\newline
\textbullet\ transformer: $37.83$M \\
Base &
$86.06$M\newline
\textbullet\ tokenizer: $1.01$M\newline
\textbullet\ transformer: $85.05$M \\

\bottomrule
\end{tabular}

\end{table}

\FloatBarrier \clearpage

\section{Review of clinical ECG machine learning work} \label{app:review_ecg_clinical}

\cref{tab:review_ecg_outcomes} presents how relevant cardiac and non-cardiac conditions and outcomes are diagnosed in clinical practice. Furthermore, it summarizes the weight of evidence for the use of ECG in detecting each condition. From this recent evidence, it is clear that ECG has broader utility -- it can detect many more conditions than those it is used to diagnose in current clinical practice.

\begin{table}[h]
  \centering
    \setlength\tabcolsep{2pt}
    \fontsize{7}{8}\selectfont
    \caption{Summary of diagnostic categories and clinical outcomes and corresponding evidence for their detection or prediction using ECG. Evidence strength is indicated using symbols: \StrongEvidence{} = strong evidence,  \PromisingEvidence{} = promising evidence, \ExploratoryEvidence{} = exploratory evidence.}
    \label{tab:review_ecg_outcomes}
    \begin{tabular}{
    p{0.3\textwidth}
    p{0.45\textwidth}
    p{0.125\textwidth}
    >{\centering\arraybackslash}p{0.075\textwidth}
}
    \toprule
    \textbf{Category} &
    \textbf{Diagnosis in clinical practice} &
    \textbf{ECG detection evidence} &
    \textbf{Evidence strength} \\
    \midrule

    \textbf{Arrhythmias} & & & \\ [0.3em]

    \hangindent=1em\hangafter=0Atrial fibrillation (AFib) &
    Diagnosed from standard ECG (12-lead, 10s) or ambulatory ECG (e.g., with Holter monitors) &
    \cite{hannun2019cardiologist,ribeiro2020automatic} &
    \StrongEvidence \\ 

    \hangindent=1em\hangafter=0Silent or future atrial fibrillation &
    AFib is paroxysmal  / episodic. No prototypical signatures in the ECG when AFib is not active. However, machine learning can be used to screen for AFib (future or latent) from normal sinus rhythm ECGs. &
    \cite{attia2019artificial,noseworthy2022artificial} &
    \StrongEvidence \\ 

    \hangindent=1em\hangafter=0 Other supraventricular and ventricular arrhythmias &
    Diagnosed from standard ECG (12-lead, 10s) or ambulatory ECG (e.g., with Holter monitors) &
    \cite{hannun2019cardiologist,ribeiro2020automatic} &
    \StrongEvidence \\ 

    \midrule

    \textbf{Conduction Disturbances} & 
    ECG-based diagnosis of atrioventricular and intraventricular conduction delays / blocks & 
    \cite{hannun2019cardiologist,ribeiro2020automatic} & 
    \StrongEvidence \\ [0.3em]

    \midrule

    \textbf{Ischemic Heart Disease} & & & \\[0.3em]

    \hangindent=1em\hangafter=0STE-ACS \& STEMI &
    ECG ST-elevations with confirmatory biomarkers (e.g., troponins). Coronary angiography is the gold-standard. &
    \cite{goto2019artificial,al2020machine} &
    \StrongEvidence \\

    \hangindent=1em\hangafter=0NSTE-ACS \& NSTEMI &
    Troponin elevation with supportive ECG and clinical findings. Coronary angiography is the gold-standard. &
    \cite{al2023machine} &
    \PromisingEvidence \\
    
    \hangindent=1em\hangafter=0Chronic coronary syndrome (atherosclerosis, etc.) &
    Stress testing or coronary imaging (e.g., angiography) &
    \cite{awasthi2023identification,park2024artificial,banerjee2025enhancement} &
    \ExploratoryEvidence \\ 
    
    \midrule

    \textbf{Structural Heart Disease} & & & \\ [0.3em]

    \hangindent=1em\hangafter=0 (Systolic) Heart failure with reduced ejection fraction (HFrEF) &
    Echocardiography demonstrating reduced left ventricular ejection fraction &
    \cite{attia2019screening,attia2019prospective} &
    \StrongEvidence \\
    
    \hangindent=1em\hangafter=0 (Diastolic) Heart failure with preserved ejection fraction (HFpEF) &
    Echocardiography with preserved ejection fraction and evidence of diastolic dysfunction or elevated filling pressures &
    \cite{lee2024artificial,Poterucha2025} &
    \ExploratoryEvidence \\
    
    \hangindent=1em\hangafter=0Valvular heart disease  &
    Echocardiography with Doppler flow assessment &
    \cite{elias2022deep,Poterucha2025,attia2021application} &
    \PromisingEvidence \\

    \hangindent=1em\hangafter=0Left ventricular hypertrophy (LVH)  &
    Echocardiography &
    \cite{li2026machine} &
    \PromisingEvidence \\
    
    \hangindent=1em\hangafter=0Cardiomyopathies: e.g., dilated, hypertrophic, arrhythmogenic right ventricular (ARVC) &
    Cardiac imaging (ECHO or MRI), genetic testing, and clinical evaluation &
    \cite{ko2020detection,siontis2021artificial,shrivastava2021artificial,carrick2025improved} &
    \ExploratoryEvidence \\

    \hangindent=1em\hangafter=0Other SHD -- e.g., right ventricular dysfunction &
    Cardiac imaging (ECHO or MRI) &
    \cite{vaid2022using,Poterucha2025} &
    \ExploratoryEvidence \\
    
    \midrule

    \textbf{Other} & & & \\ [0.3em]

    \hangindent=1em\hangafter=0Long QT syndrome (LQTS) &
    ECG, clinical and family history, exclusion of secondary causes, genetic testing &
    \cite{bos2021use} &
    \PromisingEvidence \\
    
    \hangindent=1em\hangafter=0Hyperkalemia &
    Blood chemistry testing (serum potassium) &
    \cite{hyperkalemia2019} &
    \StrongEvidence \\
    
    \hangindent=1em\hangafter=0Other electrolyte imbalances e.g., calcium and sodium &
    Blood chemistry testing &
    \cite{kwon2021artificial} &
    \ExploratoryEvidence \\ 
    
    \hangindent=1em\hangafter=0Cardiac amyloidosis &
    Cardiac imaging (ECHO or MRI) and biopsy &
    \cite{tison2019automated,goto2021artificial} &
    \ExploratoryEvidence \\

    \hangindent=1em\hangafter=0Acute pericarditis &
    ECG, ECHO and clinical examination &
    \cite{liu2022deep} &
    \ExploratoryEvidence \\

    \hangindent=1em\hangafter=0Obstructive sleep apnea (OSA) &
    Polysomnography &
    \cite{covassin2025deep} &
    \ExploratoryEvidence \\
    
    \hangindent=1em\hangafter=0Anemia &
    Blood hemoglobin measurement &
    \cite{kwon2020deep} &
    \ExploratoryEvidence \\

    \hangindent=1em\hangafter=0Chronic kidney disease (CKD) &
    Estimated glomerular filtration rate from lab tests &
    \cite{holmstrom2023deep} &
    \ExploratoryEvidence \\ 

    \hangindent=1em\hangafter=0Pulmonary conditions (e.g., pulmonary embolism, hypertension, etc.) &
    Various chest imaging (e.g., X-Ray, CT), biomarkers, and clinical assessment &
    \cite{tison2019automated,attia2021application,Wysokinski2024} &
    \ExploratoryEvidence \\

    \hangindent=1em\hangafter=0Liver disease &
    Lab testing and abdominal imaging &
    \cite{ahn2022development} &
    \ExploratoryEvidence \\

    \hangindent=1em\hangafter=0Hyperthyroidism &
    Thyroid function tests &
    \cite{choi2022electrocardiographic} &
    \ExploratoryEvidence \\

    \bottomrule
\end{tabular}
\end{table}

\FloatBarrier \clearpage

\section{More information on downstream datasets} \label{app:datasets}

\subsection{Grouping of labels into logical diagnostic groups} \label{app:dx_groups}

\cref{tab:downstream_diagnosis_groups} lists the membership of the diagnosis groups we use in \cref{fig:perf_profile_external} and \cref{fig:perf_profile_internal}.

\begin{table}[h]
  \centering
    \setlength\tabcolsep{2pt}
    \fontsize{6}{8}\selectfont
    \caption{Diagnosis groups used for aggregated evaluation. Each group pools related binary labels from the indicated dataset sources. Only labels with adequate prevalence are included in the aggregate groups (at least 10 positive and negative class instances). \textsuperscript{*}Only for comparison between internal models and ECGFounder as other external baselines pre-train on \textbf{all} of the MIMIC-IV-ECG data.}
    \label{tab:downstream_diagnosis_groups}
    \begin{tabular}{
    >{\centering\arraybackslash}p{0.04\textwidth}
    >{\raggedright\arraybackslash}p{0.28\textwidth}
    >{\raggedright\arraybackslash}p{0.66\textwidth}
}
\toprule
\textbf{\#} & \textbf{Diagnosis group} & \textbf{Dataset source and labels} \\
\midrule
1 & \textbf{All}: all diagnostic labels & External model comparison: all labels from datasets (1) PTB-XL, (2) CPSC2018, (3) CSN, (4) EchoNext-Mini\newline Internal model comparison: all labels from (1) PTB-XL, (2) CPSC2018, (3) CSN, (4) EchoNext-Mini, (5) MIMIC-IV-ECG-Ext-ICD \\
2 & \textbf{ARR}: arrhythmias & PTB-XL Rhythm: atrial fibrillation (AFIB), sinus tachycardia (STACH), sinus arrhythmia (SARRH), sinus bradycardia (SBRAD), supraventricular arrhythmia (SVARR) \\
3 & \textbf{CD}: conduction disturbances & PTB-XL Sub: left anterior/left posterior fascicular block (LAFB/LPFB), incomplete right bundle branch block (IRBBB), complete left bundle branch block (CLBBB), complete right bundle branch block (CRBBB), AV block (AVB), non-specific intraventricular conduction disturbance (block) (IVCD) \\
4 & \textbf{ECG-HYP}: hypertrophy detected from ECG & PTB-XL Sub: left ventricular hypertrophy (LVH), right ventricular hypertrophy (RVH), left atrial overload/enlargement (LAO/LAE), right atrial overload/enlargement (RAO/RAE) \\
5 & \textbf{MI}: myocardial infarction & PTB-XL Sub: anterior myocardial infarction (AMI), inferior myocardial infarction (IMI), lateral myocardial infarction (LMI) \\
6 & \textbf{ISC/STTC}: ischemia and ST/T changes & PTB-XL Sub: ischemic in anterior leads (ISCA), ischemic in inferior leads (ISCI), non-specific ischemic (ISC\_), ST-T changes (STTC), non-specific ST changes (NST\_) \\
7 & \textbf{SHD}: structural heart disease & EchoNext-Mini: reduced left ventricular ejection fraction (LVEF $\leq$ 45\%), increased left ventricular wall thickness (LVWT $\geq$ 13\,mm), moderate or greater aortic stenosis, moderate or greater aortic regurgitation, moderate or greater mitral regurgitation, moderate or greater tricuspid regurgitation, moderate or greater pulmonary regurgitation, moderate or greater right ventricular systolic dysfunction, moderate or large pericardial effusion, elevated pulmonary artery systolic pressure (PASP $\geq$ 45\,mmHg), elevated tricuspid regurgitation maximum velocity ($\geq$ 3.2\,m/s) \\
8 & \textbf{VC}: voltage criteria diagnoses & PTB-XL Sub: left ventricular hypertrophy (LVH), right ventricular hypertrophy (RVH), left atrial overload/enlargement (LAO/LAE), right atrial overload/enlargement (RAO/RAE), left anterior/left posterior fascicular block (LAFB/LPFB) \newline
PTB-XL Form: voltage criteria (QRS) for left ventricular hypertrophy (VCLVH), low QRS voltages in the frontal and horizontal leads (LVOLT), low amplitude T-waves (LOWT) \newline
EchoNext-Mini: left ventricular wall thickness (LVWT $\geq$13\,mm) \\
9 & \textbf{MIMIC-EST}\textsuperscript{*}: MIMIC-IV established ECG diagnoses & MIMIC-IV-ICD: ST-elevation myocardial infarction (I21.0--3), non-ST-elevation myocardial infarction (I21.4), old myocardial infarction (I25.2), systolic heart failure (I50.2), second- or third-degree atrioventricular block (I44.1--2) \\
10 & \textbf{MIMIC-EXP}\textsuperscript{*}: MIMIC-IV exploratory ECG diagnoses & MIMIC-IV-ICD: atherosclerotic heart disease (I25.1), diastolic heart failure (I50.3), long QT syndrome (I45.81), acute pericarditis (I30), dilated cardiomyopathy (I42.0), hypertrophic cardiomyopathy (I42.1--2), other cardiomyopathies (I42.8), hyperkalemia (E87.5), hypokalemia (E87.6), hypercalcemia (E83.52), hypocalcemia (E83.51), hypernatremia (E87.0), hyponatremia (E87.1), anemia (D50--D53, D55--D64), obstructive sleep apnea (G47.33), pulmonary embolism (I26), secondary pulmonary hypertension (I27.2), chronic obstructive pulmonary disease (J44), acute respiratory failure (J96.0), chronic kidney disease stages 1--3 (N18.1--3), chronic kidney disease stages 4--5 (N18.4--5), end-stage renal disease (N18.6), fibrosis and cirrhosis of the liver (K70.2--3 and K74), hyperthyroidism (E05).\newline 
\textbf{Note: cardiac amyloidosis (E85 and I43)} is excluded from the aggregate due to low prevalence but included in the label-level tables. \\
\bottomrule
\end{tabular}

\end{table}

\subsection{Glossary of condition codes} \label{app:condition_codes}

\cref{tab:condition_codes} defines the condition codes used in \cref{tab:label_level_main_auroc,tab:label_level_auprc,tab:label_level_lead_auroc,tab:label_level_lead_auprc}.

\begin{table}[h]
  \centering
    \setlength\tabcolsep{2pt}
    \fontsize{5}{6}\selectfont
    \caption{Glossary of condition codes.}
    \label{tab:condition_codes}
    \begin{tabular}{p{0.28\textwidth}p{0.68\textwidth}}
\toprule
\textbf{Code} & \textbf{Description} \\
\midrule
AFIB & Atrial fibrillation \\
STACH & Sinus tachycardia \\
SARRH & Sinus arrhythmia \\
SBRAD & Sinus bradycardia \\
SVARR & Supraventricular arrhythmia \\
LAF/LPF Blocks & Left anterior/left posterior fascicular block \\
IRBB Block & Incomplete right bundle branch block \\
CLBB Block & Complete left bundle branch block \\
CRBB Block & Complete right bundle branch block \\
AV Block & Atrioventricular block \\
IVCD & Non-specific intraventricular conduction disturbance \\
LVH & Left ventricular hypertrophy \\
RVH & Right ventricular hypertrophy \\
LAO/LAE & Left atrial overload/enlargement \\
RAO/RAE & Right atrial overload/enlargement \\
Anterior MI & Anterior myocardial infarction \\
Inferior MI & Inferior myocardial infarction \\
Lateral MI & Lateral myocardial infarction \\
Anterior ISC & Ischemic in anterior leads \\
Inferior ISC & Ischemic in inferior leads \\
Non-specific ISC & Non-specific ischemic \\
STTC & ST-T changes \\
Non-specific STC & Non-specific ST changes \\
Reduced LVEF & Reduced left ventricular ejection fraction (45 percent or less) \\
Increased LVWT & Increased left ventricular wall thickness (13 mm or more) \\
Aortic Stenosis & Moderate or greater aortic stenosis \\
Aortic Regurg. & Moderate or greater aortic regurgitation \\
Mitral Regurg. & Moderate or greater mitral regurgitation \\
Tricuspid Regurg. & Moderate or greater tricuspid regurgitation \\
Pulm. Regurg. & Moderate or greater pulmonary regurgitation \\
RV Sys. Dys. & Moderate or greater right ventricular systolic dysfunction \\
Pericardial Eff. & Moderate or large pericardial effusion \\
Elevated PASP & Elevated pulmonary artery systolic pressure (45 mmHg or more) \\
Elevated TRV & Elevated tricuspid regurgitation maximum velocity (3.2 m/s or more) \\
I21.0-3: STEMI & I21.0-3: ST elevation myocardial infarction (STEMI) \\
I21.4: NSTEMI & I21.4: Non-ST elevation myocardial infarction (NSTEMI) \\
I25.2: Prior MI & I25.2: Old myocardial infarction \\
I50.2: Sys. HF & I50.2: Systolic (congestive) heart failure \\
I44.1-2: 2/3 AVB & I44.1-2: Atrioventricular block, second or third degree \\
I25.1: AHD/CAD & I25.1: Atherosclerotic heart disease of native coronary artery \\
I50.3: Dia. HF & I50.3: Diastolic (congestive) heart failure \\
I45.81: LongQT & I45.81: Long QT syndrome \\
I30: Pericarditis & I30: Acute pericarditis \\
I42.0: DCM & I42.0: Dilated cardiomyopathy \\
I42.1-2: HCM & I42.1-2: Hypertrophic cardiomyopathy (HCM) \\
I42.8: OtherCM & I42.8: Other cardiomyopathies \\
E87.5: HyperK & E87.5: Hyperkalemia \\
E87.6: HypoK & E87.6: Hypokalemia \\
E83.52: HyperCa & E83.52: Hypercalcemia \\
E83.51: HypoCa & E83.51: Hypocalcemia \\
E87.0: HyperNa & E87.0: Hyperosmolality and hypernatremia \\
E87.1: HypoNa & E87.1: Hypo-osmolality and hyponatremia \\
D50-D64: Anemia & D50-D53, D55-D64: Anemia \\
G47.33: OSA & G47.33: Obstructive sleep apnea \\
I26: PE & I26: Pulmonary embolism \\
I27.2: PHT & I27.2: Other secondary pulmonary hypertension \\
J44: COPD & J44: Other chronic obstructive pulmonary disease \\
J96.0: ARF & J96.0: Acute respiratory failure \\
N18.1-3: CKD1-3 & N18.1-3: Chronic kidney disease, stages 1-3 \\
N18.4-5: CKD4-5 & N18.4-5: Chronic kidney disease, stages 4-5 \\
N18.6: ESKD & N18.6: End stage renal disease \\
K70/K74: FCL & K70.2-3 and K74: Fibrosis and cirrhosis of liver \\
E05: HyperThy & E05: Thyrotoxicosis [hyperthyroidism] \\
E85 and I43: CA & E85 and I43: Cardiac amyloidosis \\
\bottomrule
\end{tabular}

\end{table}

\subsection{Disease coverage by dataset} \label{app:datasets_coverage}

\cref{tab:review_downstream_datasets} summarizes the disease coverage of the different downstream datasets used for evaluation.

\begin{table}[h]
  \centering
    \setlength\tabcolsep{2pt}
    \fontsize{7}{8}\selectfont
    \caption{Disease coverage by downstream evaluation datasets. This expanded evaluation suite has more universal coverage of diseases that are detectable from ECGs by expanding it to include EchoNext-Mini and MIMIC-IV-ECG-Ext-ICD. Number of total instances in the dataset is in parentheses. \cmark{} indicates the dataset supports the condition, \xmark{} indicates it does not, and \pmark{} denotes partial or limited support of the condition as currently used. Labels for MIMIC-IV-ECG-Ext-ICD are at the visit-level.}
    \label{tab:review_downstream_datasets}
    \begin{tabular}{
    p{0.28\textwidth}
    >{\centering\arraybackslash}p{0.12\textwidth}
    >{\centering\arraybackslash}p{0.12\textwidth}
    >{\centering\arraybackslash}p{0.12\textwidth}
    >{\centering\arraybackslash}p{0.12\textwidth}
    >{\centering\arraybackslash}p{0.14\textwidth}
}
\toprule
\vfill \textbf{Category} &
\vfill \textbf{PTB-XL} (21,837) &
\vfill \textbf{CPSC2018} (6,877) &
\vfill \textbf{CSN} (23,026) &
\vfill \textbf{EchoNext-Mini} (100,000) &
\vfill \textbf{MIMIC-IV-ECG-Ext-ICD} (468,005) \\
\midrule

\textbf{Arrhythmias} & 
\cmark & \cmark & \cmark & \xmark & \cmark \\[0.3em]

\midrule

\textbf{Conduction Disturbances} &
\cmark & \cmark & \cmark & \xmark & \cmark \\[0.3em]

\midrule

\textbf{Ischemic Heart Disease} & & & & & \\[0.3em]

\hangindent=1em\hangafter=0STE-ACS / STEMI &
\cmark & \pmark & \pmark & \xmark & \cmark \\

\hangindent=1em\hangafter=0NSTE-ACS / NSTEMI &
\xmark & \xmark & \xmark & \xmark & \cmark \\

\hangindent=1em\hangafter=0Chronic coronary syndrome (e.g., atherosclerosis, etc.) &
\pmark & \xmark & \xmark & \xmark & \cmark \\

\midrule

\textbf{Structural Heart Disease} & & & & & \\[0.3em]

\hangindent=1em\hangafter=0(Sys.) HFrEF &
\xmark & \xmark & \xmark & \cmark & \cmark \\

\hangindent=1em\hangafter=0(Dias.) HFpEF &
\xmark & \xmark & \xmark & \cmark & \cmark \\

\hangindent=1em\hangafter=0Valvular heart disease &
\xmark & \xmark & \xmark & \cmark & \cmark \\

\hangindent=1em\hangafter=0Left ventricular hypertrophy &
\cmark & \xmark & \pmark & \cmark & \cmark \\

\hangindent=1em\hangafter=0Cardiomyopathies &
\xmark & \xmark & \xmark & \pmark & \cmark \\

\hangindent=1em\hangafter=0Other SHD &
\xmark & \xmark & \xmark & \cmark & \cmark \\

\midrule

\textbf{Other} & & & & & \\[0.3em]

\hangindent=1em\hangafter=0Long QT Syndrome &
\xmark & \xmark & \xmark & \xmark & \cmark \\

\hangindent=1em\hangafter=0Hyperkalemia and other electrolyte imbalances &
\xmark & \xmark & \xmark & \xmark & \cmark \\

\hangindent=1em\hangafter=0Acute pericarditis &
\xmark & \xmark & \xmark & \xmark & \cmark \\

\hangindent=1em\hangafter=0Cardiac amyloidosis &
\xmark & \xmark & \xmark & \xmark & \cmark \\

\hangindent=1em\hangafter=0Obstructive sleep apnea (OSA) &
\xmark & \xmark & \xmark & \xmark & \cmark \\

\hangindent=1em\hangafter=0Anemia &
\xmark & \xmark & \xmark & \xmark & \cmark \\

\hangindent=1em\hangafter=0Chronic kidney disease (CKD) &
\xmark & \xmark & \xmark & \xmark & \cmark \\

\hangindent=1em\hangafter=0Pulmonary conditions &
\xmark & \xmark & \xmark & \xmark & \cmark \\

\hangindent=1em\hangafter=0Liver disease &
\xmark & \xmark & \xmark & \xmark & \cmark \\

\hangindent=1em\hangafter=0Hyperthyroidism &
\xmark & \xmark & \xmark & \xmark & \cmark \\

\midrule

\textbf{All ICD10 Diseases} &
\xmark & \xmark & \xmark & \xmark & \cmark \\

\bottomrule
\end{tabular}

\end{table}

\FloatBarrier \clearpage

\section{Supplementary experimental results} \label{app:more_results}

\subsection{Label-level results for \methodname{} by reduced-lead configuration} \label{app:label_level_results_by_lead}

\cref{tab:label_level_lead_auroc,tab:label_level_lead_auprc} report the linear probe AUROC and AUPRC of \methodname{} for the individual labels in \cref{tab:label_level_main_auroc}, at the single-lead, reduced-lead and 12-lead configurations used in \cref{tab:reduced_lead}.

\begin{table}[h!]
    \centering
    \setlength\tabcolsep{2pt}
    \fontsize{4.5}{5}\selectfont
    \caption{Label-level linear probe AUROC of \methodname{} by lead configuration. A glossary of condition codes is in \cref{tab:condition_codes}.}
    \label{tab:label_level_lead_auroc}
    \begin{tabular}{l>{\raggedright\arraybackslash}p{0.13\linewidth}c|rrr|rrrr|r}
    \toprule
    & & \multicolumn{1}{c}{} & \multicolumn{3}{c}{\textbf{Single Lead}} & \multicolumn{4}{c}{\textbf{Lead Subsets}} & \multicolumn{1}{c}{\textbf{12-lead}} \\
    \cmidrule{4-6} \cmidrule(lr){7-10} \cmidrule(lr){11-11}
    \textbf{Group} & \textbf{Condition} & \textbf{Prevalence \% (N)} & \textbf{I (LCX)} & \textbf{II (RCA)} & \textbf{V2 (LAD)} & \textbf{I,II} & \textbf{All limb} & \textbf{I,II,V2} & \textbf{I,II,III,V2} & \textbf{12-lead} \\
    \midrule
    \multirow{5}{*}{ARR} & AFIB & 7.2\% (152) & 97.7 & 98.5 & 97.3 & 98.5 & 98.7 & 98.5 & 98.5 & 98.7 \\
     & STACH & 3.9\% (82) & 99.2 & 99.3 & 99.1 & 99.5 & 99.5 & 99.5 & 99.5 & 99.5 \\
     & SARRH & 3.7\% (77) & 90.0 & 92.7 & 89.3 & 94.1 & 93.4 & 93.5 & 93.4 & 94.4 \\
     & SBRAD & 3.1\% (64) & 96.1 & 96.1 & 95.9 & 96.3 & 96.1 & 96.1 & 96.0 & 96.7 \\
     & SVARR & 0.7\% (14) & 84.9 & 96.1 & 71.9 & 93.7 & 93.7 & 91.8 & 92.4 & 95.4 \\
    \midrule
    \multirow{6}{*}{CD} & LAF/LPF Blocks & 8.3\% (179) & 79.4 & 95.5 & 76.9 & 97.7 & 98.1 & 98.0 & 98.1 & 98.2 \\
     & IRBB Block & 5.2\% (112) & 79.1 & 82.1 & 90.6 & 86.4 & 86.0 & 95.1 & 94.9 & 98.1 \\
     & CLBB Block & 2.5\% (54) & 99.8 & 98.1 & 99.9 & 99.4 & 99.3 & 99.5 & 99.7 & 99.8 \\
     & CRBB Block & 2.5\% (54) & 99.6 & 96.8 & 99.6 & 99.6 & 99.6 & 99.8 & 99.7 & 99.8 \\
     & AV Block & 3.8\% (82) & 95.7 & 97.8 & 95.6 & 97.9 & 98.1 & 97.7 & 97.8 & 98.1 \\
     & IVCD & 3.7\% (79) & 72.9 & 74.4 & 73.1 & 77.4 & 78.8 & 78.5 & 79.0 & 80.8 \\
    \midrule
    \multirow{4}{*}{ECG-HYP} & LVH & 9.9\% (214) & 86.3 & 83.2 & 79.0 & 90.2 & 90.6 & 91.4 & 91.7 & 95.3 \\
     & RVH & 0.6\% (12) & 92.8 & 88.0 & 89.1 & 92.4 & 94.6 & 97.4 & 97.4 & 91.2 \\
     & LAO/LAE & 1.9\% (42) & 79.0 & 87.0 & 75.9 & 86.2 & 87.4 & 88.2 & 88.8 & 89.7 \\
     & RAO/RAE & 0.5\% (10) & 92.6 & 98.5 & 81.9 & 97.4 & 98.2 & 97.2 & 97.4 & 99.5 \\
    \midrule
    \multirow{3}{*}{MI} & Anterior MI & 14.2\% (306) & 84.4 & 80.4 & 95.6 & 86.3 & 86.8 & 96.5 & 96.3 & 97.1 \\
     & Inferior MI & 15.2\% (327) & 72.9 & 88.4 & 70.4 & 93.6 & 94.7 & 93.6 & 94.8 & 94.6 \\
     & Lateral MI & 0.9\% (20) & 85.8 & 72.4 & 68.5 & 87.9 & 90.0 & 88.1 & 90.3 & 91.4 \\
    \midrule
    \multirow{5}{*}{ISC/STTC} & Anterior ISC & 4.3\% (93) & 90.3 & 82.7 & 80.0 & 90.9 & 91.2 & 92.4 & 92.5 & 94.1 \\
     & Inferior ISC & 1.9\% (40) & 77.9 & 91.2 & 71.7 & 95.4 & 95.8 & 95.8 & 96.1 & 96.3 \\
     & Non-specific ISC & 5.9\% (128) & 92.0 & 91.1 & 84.9 & 94.1 & 94.1 & 94.9 & 94.8 & 96.7 \\
     & STTC & 10.3\% (222) & 81.9 & 84.0 & 74.5 & 86.9 & 87.0 & 88.8 & 88.7 & 90.0 \\
     & Non-specific STC & 3.6\% (77) & 86.2 & 83.4 & 77.4 & 86.5 & 86.2 & 87.5 & 87.3 & 87.2 \\
    \midrule
    \multirow{11}{*}{SHD} & Reduced LVEF & 17.7\% (962) & 86.8 & 85.8 & 84.1 & 88.7 & 88.5 & 89.9 & 89.9 & 90.7 \\
     & Increased LVWT & 19.5\% (1,061) & 73.0 & 72.0 & 71.5 & 74.7 & 74.8 & 76.8 & 77.2 & 78.4 \\
     & Aortic Stenosis & 5.3\% (286) & 80.1 & 78.7 & 78.3 & 82.3 & 82.3 & 83.5 & 83.9 & 85.4 \\
     & Aortic Regurg. & 1.2\% (66) & 73.9 & 70.8 & 71.4 & 69.1 & 72.1 & 73.6 & 73.3 & 75.0 \\
     & Mitral Regurg. & 6.2\% (337) & 81.1 & 79.6 & 79.4 & 81.5 & 81.9 & 82.8 & 82.9 & 84.2 \\
     & Tricuspid Regurg. & 6.5\% (353) & 79.8 & 78.6 & 80.4 & 81.3 & 81.4 & 83.3 & 83.4 & 85.0 \\
     & Pulm. Regurg. & 0.4\% (20) & 83.6 & 81.4 & 80.7 & 85.9 & 88.5 & 89.4 & 89.4 & 90.2 \\
     & RV Sys. Dys. & 7.7\% (419) & 85.5 & 85.3 & 85.3 & 87.3 & 87.5 & 88.6 & 88.8 & 89.5 \\
     & Pericardial Eff. & 1.3\% (69) & 74.8 & 74.8 & 67.2 & 76.0 & 77.2 & 79.1 & 79.0 & 79.1 \\
     & Elevated PASP & 12.8\% (699) & 75.1 & 73.5 & 75.8 & 75.8 & 76.4 & 78.3 & 78.5 & 80.3 \\
     & Elevated TRV & 6.9\% (375) & 72.9 & 71.4 & 73.6 & 74.3 & 74.5 & 77.7 & 77.7 & 80.1 \\
    \midrule
    \multirow{5}{*}{MIMIC-EST} & I21.0-3: STEMI & 1.0\% (118) & 81.6 & 84.5 & 82.1 & 91.2 & 91.6 & 92.9 & 93.0 & 93.4 \\
     & I21.4: NSTEMI & 3.2\% (396) & 80.0 & 78.9 & 73.3 & 82.6 & 82.9 & 83.8 & 83.6 & 85.1 \\
     & I25.2: Prior MI & 7.1\% (880) & 74.4 & 75.0 & 71.0 & 78.0 & 78.4 & 78.6 & 79.1 & 79.8 \\
     & I50.2: Sys. HF & 6.1\% (760) & 90.1 & 88.2 & 87.0 & 91.3 & 91.5 & 91.6 & 91.5 & 92.2 \\
     & I44.1-2: 2/3 AVB & 1.0\% (118) & 90.1 & 89.6 & 88.1 & 91.4 & 91.0 & 91.2 & 90.9 & 90.7 \\
    \midrule
    \multirow{25}{*}{MIMIC-EXP} & I25.1: AHD/CAD & 21.4\% (2,641) & 77.3 & 76.3 & 73.2 & 79.5 & 80.0 & 80.2 & 80.6 & 81.1 \\
     & I50.3: Dia. HF & 7.8\% (968) & 78.6 & 78.6 & 77.3 & 80.9 & 81.3 & 81.5 & 82.0 & 83.0 \\
     & I45.81: LongQT & 0.4\% (47) & 82.8 & 81.7 & 78.7 & 82.9 & 81.9 & 83.4 & 82.9 & 82.5 \\
     & I30: Pericarditis & 0.3\% (32) & 77.4 & 85.9 & 75.2 & 87.2 & 86.6 & 87.0 & 86.3 & 86.4 \\
     & I42.0: DCM & 0.1\% (17) & 91.7 & 88.2 & 86.6 & 91.8 & 91.6 & 93.2 & 93.1 & 94.0 \\
     & I42.1-2: HCM & 0.3\% (35) & 74.7 & 75.4 & 78.3 & 78.6 & 79.6 & 78.8 & 82.2 & 82.1 \\
     & I42.8: OtherCM & 1.9\% (233) & 86.3 & 84.4 & 83.9 & 86.3 & 87.4 & 86.9 & 87.5 & 88.4 \\
     & E87.5: HyperK & 4.0\% (490) & 72.6 & 74.4 & 72.0 & 76.7 & 77.1 & 78.4 & 78.4 & 80.3 \\
     & E87.6: HypoK & 2.8\% (348) & 71.4 & 71.8 & 69.7 & 74.3 & 74.6 & 74.4 & 74.7 & 75.2 \\
     & E83.52: HyperCa & 0.6\% (71) & 62.1 & 63.0 & 65.6 & 65.8 & 65.3 & 68.9 & 68.6 & 68.8 \\
     & E83.51: HypoCa & 0.7\% (83) & 75.9 & 74.7 & 74.9 & 74.8 & 77.5 & 76.8 & 77.8 & 79.0 \\
     & E87.0: HyperNa & 2.1\% (254) & 75.5 & 74.1 & 73.0 & 76.1 & 76.8 & 78.3 & 78.4 & 78.9 \\
     & E87.1: HypoNa & 5.8\% (720) & 67.6 & 65.8 & 66.1 & 68.6 & 68.4 & 70.1 & 69.8 & 70.0 \\
     & D50-D64: Anemia & 23.7\% (2,931) & 69.2 & 68.4 & 67.1 & 71.2 & 71.0 & 72.5 & 72.1 & 72.9 \\
     & G47.33: OSA & 6.9\% (848) & 65.9 & 66.2 & 64.8 & 69.2 & 69.9 & 70.1 & 70.8 & 72.2 \\
     & I26: PE & 1.3\% (163) & 69.2 & 71.8 & 71.7 & 71.4 & 73.5 & 73.4 & 74.4 & 75.1 \\
     & I27.2: PHT & 1.4\% (178) & 82.6 & 81.6 & 82.1 & 83.8 & 84.1 & 83.9 & 84.5 & 85.5 \\
     & J44: COPD & 9.3\% (1,150) & 70.8 & 68.5 & 69.0 & 73.0 & 73.8 & 73.7 & 73.9 & 75.7 \\
     & J96.0: ARF & 2.0\% (250) & 80.6 & 81.3 & 79.9 & 81.5 & 82.7 & 82.5 & 83.3 & 84.1 \\
     & N18.1-3: CKD1-3 & 4.3\% (533) & 69.2 & 70.2 & 67.0 & 71.8 & 72.2 & 71.4 & 71.9 & 72.5 \\
     & N18.4-5: CKD4-5 & 1.5\% (185) & 77.6 & 76.0 & 74.9 & 79.0 & 79.4 & 79.1 & 79.4 & 81.4 \\
     & N18.6: ESKD & 4.3\% (534) & 81.5 & 81.3 & 83.5 & 85.2 & 85.3 & 88.0 & 87.7 & 89.0 \\
     & K70/K74: FCL & 3.7\% (455) & 73.2 & 74.8 & 70.5 & 76.5 & 77.3 & 79.1 & 79.6 & 82.6 \\
     & E05: HyperThy & 0.6\% (69) & 59.0 & 52.7 & 52.0 & 53.4 & 54.2 & 56.2 & 54.9 & 50.2 \\
     & \textcolor{gray}{E85 and I43: CA}$^{\dagger}$ & 0.0\% (6) & 73.5 & 86.4 & 90.7 & 84.0 & 89.4 & 89.7 & 91.9 & 90.9 \\
    \bottomrule
\end{tabular}

    \footnotetext{AUROC: area under the receiver operating characteristic curve. Prevalence is the proportion of positives in the test set the metrics were computed on. $^{\dagger}$Cardiac amyloidosis is excluded from the MIMIC-EXP aggregate score due to low prevalence.}
\end{table}

\begin{table}[h!]
    \centering
    \setlength\tabcolsep{2pt}
    \fontsize{4.5}{5}\selectfont
    \caption{Label-level linear probe AUPRC of \methodname{} by lead configuration. A glossary of condition codes is in \cref{tab:condition_codes}.}
    \label{tab:label_level_lead_auprc}
    \begin{tabular}{l>{\raggedright\arraybackslash}p{0.13\linewidth}c|rrr|rrrr|r}
    \toprule
    & & \multicolumn{1}{c}{} & \multicolumn{3}{c}{\textbf{Single Lead}} & \multicolumn{4}{c}{\textbf{Lead Subsets}} & \multicolumn{1}{c}{\textbf{12-lead}} \\
    \cmidrule{4-6} \cmidrule(lr){7-10} \cmidrule(lr){11-11}
    \textbf{Group} & \textbf{Condition} & \textbf{Prevalence \% (N)} & \textbf{I (LCX)} & \textbf{II (RCA)} & \textbf{V2 (LAD)} & \textbf{I,II} & \textbf{All limb} & \textbf{I,II,V2} & \textbf{I,II,III,V2} & \textbf{12-lead} \\
    \midrule
    \multirow{5}{*}{ARR} & AFIB & 7.2\% (152) & 87.4 & 92.3 & 89.3 & 93.5 & 94.5 & 94.0 & 94.4 & 95.6 \\
     & STACH & 3.9\% (82) & 86.5 & 84.8 & 86.2 & 85.7 & 86.7 & 89.0 & 89.8 & 89.5 \\
     & SARRH & 3.7\% (77) & 37.8 & 42.4 & 34.1 & 44.9 & 41.2 & 46.1 & 45.0 & 47.7 \\
     & SBRAD & 3.1\% (64) & 58.0 & 62.5 & 63.7 & 63.4 & 63.5 & 62.0 & 62.7 & 63.3 \\
     & SVARR & 0.7\% (14) & 5.9 & 30.8 & 2.6 & 28.5 & 25.6 & 34.7 & 32.2 & 48.7 \\
    \midrule
    \multirow{6}{*}{CD} & LAF/LPF Blocks & 8.3\% (179) & 35.7 & 77.8 & 31.5 & 82.5 & 83.5 & 83.3 & 83.3 & 83.3 \\
     & IRBB Block & 5.2\% (112) & 22.7 & 23.1 & 56.1 & 27.4 & 25.4 & 63.3 & 61.2 & 74.1 \\
     & CLBB Block & 2.5\% (54) & 93.8 & 85.6 & 97.2 & 94.9 & 95.2 & 96.7 & 96.1 & 96.6 \\
     & CRBB Block & 2.5\% (54) & 81.6 & 54.0 & 87.0 & 84.9 & 86.8 & 88.8 & 88.7 & 90.2 \\
     & AV Block & 3.8\% (82) & 42.4 & 59.1 & 51.7 & 60.5 & 62.2 & 61.1 & 62.9 & 68.1 \\
     & IVCD & 3.7\% (79) & 12.1 & 12.5 & 13.6 & 17.7 & 18.5 & 18.7 & 19.8 & 21.9 \\
    \midrule
    \multirow{4}{*}{ECG-HYP} & LVH & 9.9\% (214) & 51.0 & 43.1 & 36.3 & 58.3 & 59.9 & 60.2 & 61.4 & 76.7 \\
     & RVH & 0.6\% (12) & 27.2 & 8.1 & 26.9 & 14.6 & 14.8 & 18.0 & 22.4 & 22.0 \\
     & LAO/LAE & 1.9\% (42) & 8.1 & 11.2 & 6.4 & 11.7 & 13.1 & 14.4 & 14.7 & 21.4 \\
     & RAO/RAE & 0.5\% (10) & 7.1 & 51.6 & 13.2 & 54.5 & 57.5 & 54.0 & 48.7 & 54.1 \\
    \midrule
    \multirow{3}{*}{MI} & Anterior MI & 14.2\% (306) & 50.1 & 44.3 & 84.4 & 54.2 & 54.4 & 85.5 & 85.1 & 88.4 \\
     & Inferior MI & 15.2\% (327) & 30.8 & 66.3 & 33.0 & 75.8 & 78.7 & 76.4 & 79.7 & 79.1 \\
     & Lateral MI & 0.9\% (20) & 10.0 & 2.1 & 2.2 & 13.2 & 11.8 & 15.4 & 15.8 & 9.5 \\
    \midrule
    \multirow{5}{*}{ISC/STTC} & Anterior ISC & 4.3\% (93) & 27.9 & 18.6 & 13.9 & 26.9 & 26.1 & 26.4 & 26.8 & 34.6 \\
     & Inferior ISC & 1.9\% (40) & 6.2 & 22.7 & 6.3 & 37.6 & 39.8 & 44.6 & 46.1 & 42.7 \\
     & Non-specific ISC & 5.9\% (128) & 50.7 & 46.3 & 33.8 & 59.9 & 60.2 & 60.0 & 60.4 & 69.8 \\
     & STTC & 10.3\% (222) & 32.7 & 37.7 & 27.2 & 43.0 & 42.5 & 47.8 & 47.8 & 52.2 \\
     & Non-specific STC & 3.6\% (77) & 17.6 & 14.7 & 13.6 & 19.3 & 19.5 & 22.3 & 21.1 & 19.9 \\
    \midrule
    \multirow{11}{*}{SHD} & Reduced LVEF & 17.7\% (962) & 64.6 & 61.7 & 59.4 & 69.5 & 69.6 & 72.8 & 73.3 & 75.9 \\
     & Increased LVWT & 19.5\% (1,061) & 38.0 & 37.1 & 37.2 & 40.5 & 40.8 & 43.8 & 44.6 & 46.2 \\
     & Aortic Stenosis & 5.3\% (286) & 20.4 & 17.8 & 19.3 & 22.3 & 23.2 & 24.3 & 25.7 & 28.5 \\
     & Aortic Regurg. & 1.2\% (66) & 3.9 & 4.0 & 4.3 & 3.7 & 3.6 & 5.2 & 5.4 & 4.9 \\
     & Mitral Regurg. & 6.2\% (337) & 22.0 & 21.0 & 22.0 & 24.6 & 25.1 & 26.5 & 27.0 & 29.6 \\
     & Tricuspid Regurg. & 6.5\% (353) & 24.2 & 23.9 & 25.0 & 27.1 & 28.3 & 29.6 & 31.2 & 32.5 \\
     & Pulm. Regurg. & 0.4\% (20) & 10.6 & 1.2 & 2.3 & 3.6 & 5.3 & 6.4 & 6.1 & 6.3 \\
     & RV Sys. Dys. & 7.7\% (419) & 38.4 & 35.4 & 37.9 & 41.8 & 42.8 & 46.4 & 47.1 & 50.6 \\
     & Pericardial Eff. & 1.3\% (69) & 4.3 & 3.9 & 3.3 & 4.1 & 4.0 & 5.1 & 6.4 & 4.6 \\
     & Elevated PASP & 12.8\% (699) & 33.3 & 29.8 & 32.0 & 33.4 & 34.1 & 36.1 & 36.5 & 39.1 \\
     & Elevated TRV & 6.9\% (375) & 18.9 & 16.1 & 17.0 & 19.9 & 20.4 & 21.9 & 22.2 & 25.5 \\
    \midrule
    \multirow{5}{*}{MIMIC-EST} & I21.0-3: STEMI & 1.0\% (118) & 19.9 & 24.3 & 23.8 & 40.0 & 42.6 & 48.3 & 49.9 & 51.5 \\
     & I21.4: NSTEMI & 3.2\% (396) & 14.4 & 11.9 & 11.5 & 16.6 & 16.2 & 18.1 & 18.7 & 22.9 \\
     & I25.2: Prior MI & 7.1\% (880) & 16.8 & 21.7 & 14.5 & 24.8 & 23.7 & 25.0 & 24.7 & 25.1 \\
     & I50.2: Sys. HF & 6.1\% (760) & 39.0 & 36.2 & 33.2 & 43.7 & 43.9 & 44.3 & 43.9 & 47.8 \\
     & I44.1-2: 2/3 AVB & 1.0\% (118) & 26.0 & 26.2 & 18.5 & 28.8 & 27.3 & 26.3 & 26.1 & 26.1 \\
    \midrule
    \multirow{25}{*}{MIMIC-EXP} & I25.1: AHD/CAD & 21.4\% (2,641) & 45.9 & 46.3 & 41.6 & 50.6 & 51.1 & 52.3 & 52.9 & 53.3 \\
     & I50.3: Dia. HF & 7.8\% (968) & 21.5 & 20.8 & 20.1 & 24.2 & 24.8 & 25.4 & 26.4 & 28.9 \\
     & I45.81: LongQT & 0.4\% (47) & 6.2 & 3.8 & 4.0 & 4.5 & 10.7 & 4.3 & 7.0 & 5.7 \\
     & I30: Pericarditis & 0.3\% (32) & 4.7 & 15.6 & 2.0 & 14.0 & 12.2 & 10.3 & 8.7 & 12.5 \\
     & I42.0: DCM & 0.1\% (17) & 4.1 & 2.2 & 1.8 & 1.9 & 1.7 & 2.3 & 2.2 & 3.6 \\
     & I42.1-2: HCM & 0.3\% (35) & 0.6 & 1.9 & 1.4 & 1.2 & 1.2 & 1.3 & 1.3 & 1.5 \\
     & I42.8: OtherCM & 1.9\% (233) & 15.3 & 13.4 & 13.9 & 14.5 & 16.7 & 15.6 & 16.6 & 17.8 \\
     & E87.5: HyperK & 4.0\% (490) & 11.4 & 15.7 & 12.9 & 16.6 & 18.0 & 18.8 & 19.4 & 22.4 \\
     & E87.6: HypoK & 2.8\% (348) & 7.0 & 8.7 & 6.7 & 8.9 & 8.9 & 8.5 & 8.6 & 8.9 \\
     & E83.52: HyperCa & 0.6\% (71) & 1.0 & 1.1 & 1.1 & 1.5 & 1.5 & 2.0 & 1.8 & 2.7 \\
     & E83.51: HypoCa & 0.7\% (83) & 2.4 & 2.3 & 3.0 & 2.2 & 2.3 & 2.5 & 2.4 & 2.9 \\
     & E87.0: HyperNa & 2.1\% (254) & 7.1 & 6.1 & 5.3 & 7.3 & 7.3 & 8.4 & 7.7 & 8.1 \\
     & E87.1: HypoNa & 5.8\% (720) & 10.7 & 10.9 & 9.7 & 12.4 & 11.9 & 13.1 & 12.9 & 12.9 \\
     & D50-D64: Anemia & 23.7\% (2,931) & 38.2 & 39.7 & 37.0 & 41.4 & 41.9 & 43.1 & 43.1 & 43.9 \\
     & G47.33: OSA & 6.9\% (848) & 11.8 & 11.0 & 11.4 & 12.7 & 13.4 & 14.3 & 14.7 & 16.6 \\
     & I26: PE & 1.3\% (163) & 3.4 & 4.5 & 5.7 & 4.6 & 5.1 & 7.0 & 7.1 & 7.5 \\
     & I27.2: PHT & 1.4\% (178) & 10.9 & 7.4 & 7.3 & 10.1 & 10.3 & 11.1 & 9.9 & 10.7 \\
     & J44: COPD & 9.3\% (1,150) & 19.7 & 16.7 & 17.8 & 23.1 & 24.2 & 24.2 & 24.8 & 27.3 \\
     & J96.0: ARF & 2.0\% (250) & 9.0 & 8.4 & 7.7 & 9.7 & 11.6 & 10.2 & 11.7 & 11.7 \\
     & N18.1-3: CKD1-3 & 4.3\% (533) & 8.1 & 8.8 & 8.0 & 9.1 & 9.7 & 8.8 & 9.4 & 9.5 \\
     & N18.4-5: CKD4-5 & 1.5\% (185) & 5.5 & 4.7 & 4.9 & 5.7 & 5.5 & 5.4 & 6.0 & 7.8 \\
     & N18.6: ESKD & 4.3\% (534) & 18.7 & 20.1 & 27.2 & 24.8 & 26.0 & 32.3 & 31.7 & 35.6 \\
     & K70/K74: FCL & 3.7\% (455) & 11.5 & 13.6 & 9.9 & 15.7 & 17.9 & 20.5 & 21.8 & 25.5 \\
     & E05: HyperThy & 0.6\% (69) & 0.8 & 0.6 & 0.6 & 0.6 & 0.6 & 0.6 & 0.7 & 0.5 \\
     & \textcolor{gray}{E85 and I43: CA}$^{\dagger}$ & 0.0\% (6) & 1.0 & 0.8 & 3.2 & 0.6 & 0.7 & 1.0 & 1.3 & 0.6 \\
    \bottomrule
\end{tabular}

    \footnotetext{AUPRC: area under the precision-recall curve. Prevalence is the proportion of positives in the test set the metrics were computed on. $^{\dagger}$Cardiac amyloidosis is excluded from the MIMIC-EXP aggregate score due to low prevalence.}
\end{table}

\clearpage

\subsection{Label-level results for \methodname{} by clinical report supervision source} \label{app:label_level_results_by_source}

\cref{tab:label_level_auprc} reports the linear probe AUPRC counterpart of \cref{tab:label_level_main_auroc}, for the same labels and alignment sources.

\begin{table}[h!]
    \centering
    \setlength\tabcolsep{2pt}
    \fontsize{4.5}{5}\selectfont
    \caption{Contribution of different clinical note types to \methodname{} AUPRC for different condition labels. A glossary of condition codes is in \cref{tab:condition_codes}.}
    \label{tab:label_level_auprc}
    \begin{tabular}{l>{\raggedright\arraybackslash}p{0.13\linewidth}c|r|rrrrr|r}
    \toprule
    & & \multicolumn{1}{c}{} & \multicolumn{7}{c}{\textbf{AUPRC}} \\
    \cmidrule(lr){4-10}
    \textbf{Group} & \textbf{Condition} & \textbf{Prevalence \% (N)} & \textbf{Random} & \textbf{ECG-MR} & \textbf{ECG-CR} & \textbf{ECHO} & \textbf{Chest X-ray} & \textbf{Discharge} & \textbf{MS-ECG-FM} \\
    \midrule
    \multirow{5}{*}{ARR} & AFIB & 7.2\% (152) & 49.6 & 94.8 & $^{\ast}$95.5 & 91.6 & 91.3 & 93.0 & \textbf{95.6} \\
     & STACH & 3.9\% (82) & 61.8 & \textbf{90.0} & $^{\ast}$88.0 & 83.0 & 80.7 & 83.6 & $^{\ast}$89.5 \\
     & SARRH & 3.7\% (77) & 5.3 & 30.8 & $^{\ast}$45.0 & 10.6 & 11.4 & 14.3 & \textbf{47.7} \\
     & SBRAD & 3.1\% (64) & 42.0 & \textbf{65.3} & $^{\ast}$61.5 & 58.4 & $^{\ast}$59.7 & 57.2 & $^{\ast}$63.3 \\
     & SVARR & 0.7\% (14) & 2.4 & $^{\ast}$35.4 & $^{\ast}$32.5 & 28.8 & $^{\ast}$36.7 & 32.6 & \textbf{48.7} \\
    \midrule
    \multirow{6}{*}{CD} & LAF/LPF Blocks & 8.3\% (179) & 52.6 & $^{\ast}$82.7 & \textbf{83.8} & 76.1 & 77.8 & 76.6 & $^{\ast}$83.3 \\
     & IRBB Block & 5.2\% (112) & 26.0 & $^{\ast}$70.1 & $^{\ast}$71.7 & 57.4 & 59.2 & 55.1 & \textbf{74.1} \\
     & CLBB Block & 2.5\% (54) & 82.6 & $^{\ast}$96.7 & \textbf{97.0} & $^{\ast}$96.1 & 96.0 & $^{\ast}$96.2 & $^{\ast}$96.6 \\
     & CRBB Block & 2.5\% (54) & 72.0 & $^{\ast}$91.1 & \textbf{91.8} & $^{\ast}$87.8 & $^{\ast}$88.9 & $^{\ast}$86.8 & $^{\ast}$90.2 \\
     & AV Block & 3.8\% (82) & 13.4 & $^{\ast}$63.2 & $^{\ast}$67.4 & $^{\ast}$62.4 & $^{\ast}$67.6 & $^{\ast}$62.9 & \textbf{68.1} \\
     & IVCD & 3.7\% (79) & 11.1 & $^{\ast}$20.5 & $^{\ast}$21.5 & 16.7 & 16.5 & $^{\ast}$17.8 & \textbf{21.9} \\
    \midrule
    \multirow{4}{*}{ECG-HYP} & LVH & 9.9\% (214) & 61.8 & $^{\ast}$76.4 & 73.5 & 71.6 & 72.2 & 70.5 & \textbf{76.7} \\
     & RVH & 0.6\% (12) & $^{\ast}$13.6 & \textbf{30.0} & $^{\ast}$23.9 & $^{\ast}$15.8 & $^{\ast}$21.4 & $^{\ast}$23.5 & 22.0 \\
     & LAO/LAE & 1.9\% (42) & 5.7 & $^{\ast}$17.4 & $^{\ast}$18.5 & 11.9 & $^{\ast}$19.5 & $^{\ast}$16.1 & \textbf{21.4} \\
     & RAO/RAE & 0.5\% (10) & 14.2 & $^{\ast}$56.6 & \textbf{63.9} & 27.6 & $^{\ast}$46.4 & $^{\ast}$43.6 & $^{\ast}$54.1 \\
    \midrule
    \multirow{3}{*}{MI} & Anterior MI & 14.2\% (306) & 60.9 & $^{\ast}$87.4 & \textbf{88.7} & 83.0 & 84.0 & 83.5 & $^{\ast}$88.4 \\
     & Inferior MI & 15.2\% (327) & 42.8 & $^{\ast}$78.1 & $^{\ast}$77.9 & 75.5 & 75.9 & $^{\ast}$77.0 & \textbf{79.1} \\
     & Lateral MI & 0.9\% (20) & 3.1 & \textbf{12.5} & $^{\ast}$10.4 & $^{\ast}$6.1 & $^{\ast}$11.6 & $^{\ast}$8.4 & $^{\ast}$9.5 \\
    \midrule
    \multirow{5}{*}{ISC/STTC} & Anterior ISC & 4.3\% (93) & 19.5 & $^{\ast}$33.6 & \textbf{37.9} & 29.8 & 31.6 & 30.5 & $^{\ast}$34.6 \\
     & Inferior ISC & 1.9\% (40) & 8.3 & $^{\ast}$43.4 & \textbf{52.6} & 28.8 & 33.2 & 35.0 & $^{\ast}$42.7 \\
     & Non-specific ISC & 5.9\% (128) & 52.5 & \textbf{73.3} & $^{\ast}$71.6 & $^{\ast}$69.1 & 67.0 & 68.1 & 69.8 \\
     & STTC & 10.3\% (222) & 32.6 & $^{\ast}$49.3 & $^{\ast}$48.2 & 44.4 & 43.9 & 44.0 & \textbf{52.2} \\
     & Non-specific STC & 3.6\% (77) & 12.5 & $^{\ast}$19.4 & $^{\ast}$19.4 & $^{\ast}$17.1 & $^{\ast}$18.7 & $^{\ast}$18.2 & \textbf{19.9} \\
    \midrule
    \multirow{11}{*}{SHD} & Reduced LVEF & 17.7\% (962) & 58.1 & 70.7 & 71.5 & 74.3 & 74.7 & $^{\ast}$74.9 & \textbf{75.9} \\
     & Increased LVWT & 19.5\% (1,061) & 41.0 & $^{\ast}$45.4 & 44.0 & $^{\ast}$45.4 & $^{\ast}$45.3 & \textbf{46.5} & $^{\ast}$46.2 \\
     & Aortic Stenosis & 5.3\% (286) & 16.3 & $^{\ast}$26.7 & 24.7 & 25.1 & 24.8 & $^{\ast}$26.6 & \textbf{28.5} \\
     & Aortic Regurg. & 1.2\% (66) & 2.7 & $^{\ast}$3.6 & $^{\ast}$3.7 & $^{\ast}$3.8 & $^{\ast}$4.2 & $^{\ast}$4.4 & \textbf{4.9} \\
     & Mitral Regurg. & 6.2\% (337) & 20.7 & 24.7 & 24.6 & 26.2 & 26.6 & 26.7 & \textbf{29.6} \\
     & Tricuspid Regurg. & 6.5\% (353) & 19.2 & 28.2 & 29.6 & $^{\ast}$30.3 & $^{\ast}$31.5 & $^{\ast}$31.8 & \textbf{32.5} \\
     & Pulm. Regurg. & 0.4\% (20) & 1.8 & $^{\ast}$2.8 & $^{\ast}$2.9 & 1.8 & $^{\ast}$4.0 & $^{\ast}$4.0 & \textbf{6.3} \\
     & RV Sys. Dys. & 7.7\% (419) & 34.3 & 45.6 & 47.1 & $^{\ast}$49.0 & $^{\ast}$49.2 & $^{\ast}$49.7 & \textbf{50.6} \\
     & Pericardial Eff. & 1.3\% (69) & 2.6 & $^{\ast}$4.5 & \textbf{5.1} & $^{\ast}$3.6 & $^{\ast}$5.1 & $^{\ast}$3.9 & $^{\ast}$4.6 \\
     & Elevated PASP & 12.8\% (699) & 26.6 & 33.9 & 35.6 & 35.4 & $^{\ast}$38.5 & $^{\ast}$38.7 & \textbf{39.1} \\
     & Elevated TRV & 6.9\% (375) & 13.5 & 19.3 & 20.5 & 21.9 & $^{\ast}$23.4 & $^{\ast}$24.7 & \textbf{25.5} \\
    \midrule
    \multirow{5}{*}{MIMIC-EST} & I21.0-3: STEMI & 1.0\% (118) & 16.8 & 42.9 & $^{\ast}$48.6 & 45.0 & $^{\ast}$48.7 & 45.9 & \textbf{51.5} \\
     & I21.4: NSTEMI & 3.2\% (396) & 10.1 & 19.0 & 17.3 & 17.7 & 18.9 & $^{\ast}$21.8 & \textbf{22.9} \\
     & I25.2: Prior MI & 7.1\% (880) & 15.4 & $^{\ast}$23.9 & $^{\ast}$23.9 & $^{\ast}$24.6 & $^{\ast}$24.1 & $^{\ast}$24.9 & \textbf{25.1} \\
     & I50.2: Sys. HF & 6.1\% (760) & 31.9 & 44.2 & 42.5 & $^{\ast}$47.3 & $^{\ast}$45.2 & $^{\ast}$46.3 & \textbf{47.8} \\
     & I44.1-2: 2/3 AVB & 1.0\% (118) & 5.7 & $^{\ast}$25.7 & \textbf{26.6} & $^{\ast}$24.2 & $^{\ast}$26.2 & $^{\ast}$24.4 & $^{\ast}$26.1 \\
    \midrule
    \multirow{25}{*}{MIMIC-EXP} & I25.1: AHD/CAD & 21.4\% (2,641) & 40.3 & 50.1 & 50.0 & 50.9 & $^{\ast}$52.7 & $^{\ast}$52.8 & \textbf{53.3} \\
     & I50.3: Dia. HF & 7.8\% (968) & 17.3 & 23.5 & 24.4 & 25.5 & \textbf{29.2} & $^{\ast}$28.5 & $^{\ast}$28.9 \\
     & I45.81: LongQT & 0.4\% (47) & 1.3 & \textbf{9.1} & $^{\ast}$5.8 & $^{\ast}$4.6 & $^{\ast}$7.1 & $^{\ast}$7.2 & $^{\ast}$5.7 \\
     & I30: Pericarditis & 0.3\% (32) & 1.5 & 8.2 & 8.5 & $^{\ast}$11.8 & \textbf{19.1} & 9.1 & $^{\ast}$12.5 \\
     & I42.0: DCM & 0.1\% (17) & 0.8 & $^{\ast}$7.8 & $^{\ast}$7.4 & \textbf{8.2} & $^{\ast}$6.6 & $^{\ast}$6.5 & $^{\ast}$3.6 \\
     & I42.1-2: HCM & 0.3\% (35) & $^{\ast}$0.9 & $^{\ast}$1.2 & $^{\ast}$1.2 & $^{\ast}$1.3 & \textbf{1.6} & $^{\ast}$1.1 & $^{\ast}$1.5 \\
     & I42.8: OtherCM & 1.9\% (233) & 10.2 & $^{\ast}$16.9 & $^{\ast}$16.1 & $^{\ast}$17.7 & 15.5 & \textbf{19.2} & $^{\ast}$17.8 \\
     & E87.5: HyperK & 4.0\% (490) & 9.4 & 15.9 & 18.0 & 16.7 & $^{\ast}$20.9 & $^{\ast}$21.5 & \textbf{22.4} \\
     & E87.6: HypoK & 2.8\% (348) & 5.1 & 8.6 & 8.2 & $^{\ast}$8.6 & $^{\ast}$8.7 & \textbf{10.3} & $^{\ast}$8.9 \\
     & E83.52: HyperCa & 0.6\% (71) & 0.7 & $^{\ast}$1.6 & $^{\ast}$1.5 & $^{\ast}$2.3 & $^{\ast}$2.6 & $^{\ast}$2.0 & \textbf{2.7} \\
     & E83.51: HypoCa & 0.7\% (83) & 1.4 & $^{\ast}$3.2 & 2.0 & $^{\ast}$2.4 & 2.2 & \textbf{3.5} & $^{\ast}$2.9 \\
     & E87.0: HyperNa & 2.1\% (254) & 4.1 & $^{\ast}$6.9 & $^{\ast}$7.6 & $^{\ast}$7.9 & $^{\ast}$7.9 & $^{\ast}$7.6 & \textbf{8.1} \\
     & E87.1: HypoNa & 5.8\% (720) & 9.7 & $^{\ast}$11.9 & $^{\ast}$11.8 & 11.7 & $^{\ast}$12.4 & $^{\ast}$12.8 & \textbf{12.9} \\
     & D50-D64: Anemia & 23.7\% (2,931) & 34.8 & 41.0 & 41.3 & 41.6 & $^{\ast}$43.6 & \textbf{44.2} & $^{\ast}$43.9 \\
     & G47.33: OSA & 6.9\% (848) & 10.6 & 14.5 & 15.1 & $^{\ast}$15.2 & $^{\ast}$15.8 & $^{\ast}$16.2 & \textbf{16.6} \\
     & I26: PE & 1.3\% (163) & 2.6 & $^{\ast}$5.7 & $^{\ast}$5.9 & $^{\ast}$6.0 & $^{\ast}$6.9 & \textbf{8.0} & $^{\ast}$7.5 \\
     & I27.2: PHT & 1.4\% (178) & 4.6 & 8.2 & 8.0 & $^{\ast}$9.3 & $^{\ast}$9.6 & $^{\ast}$9.8 & \textbf{10.7} \\
     & J44: COPD & 9.3\% (1,150) & 15.0 & 22.0 & 24.9 & 24.8 & $^{\ast}$26.5 & $^{\ast}$26.4 & \textbf{27.3} \\
     & J96.0: ARF & 2.0\% (250) & 5.4 & $^{\ast}$10.2 & $^{\ast}$9.3 & 8.0 & 8.9 & $^{\ast}$9.7 & \textbf{11.7} \\
     & N18.1-3: CKD1-3 & 4.3\% (533) & 7.5 & 8.5 & $^{\ast}$9.6 & 8.5 & 8.8 & \textbf{9.9} & $^{\ast}$9.5 \\
     & N18.4-5: CKD4-5 & 1.5\% (185) & 3.9 & 5.2 & 5.0 & 4.9 & \textbf{8.0} & $^{\ast}$6.6 & $^{\ast}$7.8 \\
     & N18.6: ESKD & 4.3\% (534) & 18.6 & 28.5 & 29.6 & 30.4 & 31.5 & 32.4 & \textbf{35.6} \\
     & K70/K74: FCL & 3.7\% (455) & 6.4 & 18.3 & 16.8 & 16.4 & 23.9 & \textbf{27.0} & $^{\ast}$25.5 \\
     & E05: HyperThy & 0.6\% (69) & $^{\ast}$0.6 & $^{\ast}$0.5 & $^{\ast}$0.6 & \textbf{0.7} & $^{\ast}$0.7 & $^{\ast}$0.6 & $^{\ast}$0.5 \\
     & \textcolor{gray}{E85 and I43: CA}$^{\dagger}$ & 0.0\% (6) & $^{\ast}$0.7 & $^{\ast}$1.6 & $^{\ast}$1.1 & \textbf{8.7} & $^{\ast}$1.8 & $^{\ast}$1.8 & $^{\ast}$0.6 \\
    \bottomrule
\end{tabular}

    \footnotetext{AUPRC: area under the precision-recall curve. Prevalence is the proportion of positives in the test set the metrics were computed on. The best model is \textbf{bold}; $\ast$ marks models that are not significantly different from the best per a paired patient-cluster bootstrap significance test (see Methods). $^{\dagger}$Cardiac amyloidosis is excluded from the MIMIC-EXP aggregate score due to low prevalence.}
\end{table}

\FloatBarrier \clearpage

\section{\methodname{} ablations} \label{app:ablations}

\subsection{Ablation of other key components in \methodname{}} \label{app:other_components_ablation}

\cref{tab:other_components_ablation} ablates other components of \methodname{}'s design besides the multi-source alignment objective. \methodname{} does not use input normalization so that amplitude information is preserved for the detection of conditions that rely on voltage criteria. It also uses a deep convolutional tokenizer to promote the learning of stronger local features. Several steps were also taken to improve transfer to reduced-lead configurations. These include using the random lead masking (RLM) augmentation during pre-training \cite{oh2022lead}. Furthermore, lead-level tokens are produced in the tokenizer (rather than tokenizing patches from all leads into a single token) and the tokenizer weights are shared across leads.

\begin{table}[h!]
    \centering
    \setlength\tabcolsep{1pt}
    \fontsize{5}{6}\selectfont
    \caption{Ablation of other \methodname{} components. Results for individual leads are for the \emph{All dx} group. The \methodname{} row is AUROC per diagnostic group. Other rows are $\Delta$AUROC versus \methodname{}. $\uparrow$ and $\downarrow$ mark the sign of the difference. RLM: random lead masking augmentation \cite{oh2022lead}.}
    \label{tab:other_components_ablation}
    \begin{tabular}{lrrrrrrrrrr|rrr}
    \toprule
    & \multicolumn{10}{c}{\textbf{12-lead}} & \multicolumn{3}{c}{\textbf{Reduced-lead}} \\
    \cmidrule{2-11} \cmidrule(lr){12-14}
     & \textbf{All} & \textbf{ARR} & \textbf{CD} & \textbf{ECG-HYP} & \textbf{MI} & \textbf{ISC/STTC} & \textbf{SHD} & \textbf{VC} & \textbf{MIMIC-EST} & \textbf{MIMIC-EXP} & \textbf{I} & \textbf{II} & \textbf{V2} \\
    \midrule
    \textbf{MS-ECG-FM} & 90.6 & 96.9 & 95.8 & 93.9 & 94.4 & 92.9 & 83.4 & 91.7 & 88.2 & 78.8 & 84.5 & 85.3 & 82.6 \\ \midrule
    Z-score waveform-level & $\downarrow$0.5 & $\uparrow$0.1 & $\downarrow$0.2 & $\downarrow$0.5 & 0.0 & $\downarrow$1.0 & $\downarrow$0.6 & $\downarrow$1.1 & $\downarrow$0.2 & $\downarrow$0.4 & $\downarrow$0.6 & $\downarrow$0.7 & $\downarrow$0.5 \\
    Z-score lead-level & $\downarrow$0.7 & $\downarrow$0.2 & $\downarrow$0.2 & $\downarrow$2.4 & $\downarrow$0.8 & $\downarrow$1.1 & $\downarrow$1.3 & $\downarrow$2.8 & $\downarrow$0.3 & $\downarrow$0.4 & $\downarrow$0.6 & $\downarrow$0.8 & $\downarrow$0.2 \\
    Single layer tokenizer & $\downarrow$0.3 & $\uparrow$0.5 & $\downarrow$0.3 & $\uparrow$1.0 & $\downarrow$1.0 & $\downarrow$0.3 & $\downarrow$1.0 & $\uparrow$0.5 & $\downarrow$0.6 & $\downarrow$1.2 & $\downarrow$0.5 & $\downarrow$0.3 & $\downarrow$0.3 \\ \midrule
    No RLM & $\downarrow$0.3 & $\uparrow$0.1 & $\downarrow$0.6 & $\uparrow$0.2 & $\uparrow$0.4 & $\uparrow$0.2 & $\downarrow$0.8 & $\downarrow$0.1 & $\downarrow$0.9 & $\downarrow$0.7 & $\downarrow$2.3 & $\downarrow$2.4 & $\downarrow$2.0 \\
    No lead-level tokens & $\downarrow$0.6 & $\uparrow$0.5 & $\downarrow$0.4 & $\uparrow$0.5 & $\downarrow$0.9 & $\downarrow$0.8 & $\downarrow$1.1 & $\uparrow$0.4 & $\downarrow$0.6 & $\downarrow$1.6 & $\downarrow$0.8 & $\downarrow$0.7 & $\downarrow$1.1 \\
    No tokenizer weight sharing & $\downarrow$0.2 & $\downarrow$0.2 & $\downarrow$0.4 & $\uparrow$1.0 & $\uparrow$0.8 & $\downarrow$0.2 & $\downarrow$0.2 & $\uparrow$0.5 & $\downarrow$0.2 & $\downarrow$0.3 & $\downarrow$0.9 & $\downarrow$0.6 & $\downarrow$0.4 \\
    \bottomrule
\end{tabular}

\end{table}

\subsection{Ablation of multi-source alignment objectives for \methodname{}} \label{app:ms_objective_ablation}

\cref{tab:ms_objective_ablation} shows the performance when pre-training \methodname{}'s backbone with alternative multi-source alignment contrastive learning objectives. \emph{Average embeddings} computes the average embedding over available reports in the batch and aligns the ECG to that. \emph{Report-specific batch} forms batches containing only ECGs that have a certain report type, and alternates between different report types for different batches. \emph{Report-specific projectors} has a pair of projector heads for each report type and forms a hybrid loss function in each batch with a term for each report type. \methodname{} uses \emph{stochastic pairing} which is marginally the strongest approach.

\begin{table}[h]
    \centering
    \setlength\tabcolsep{1.5pt}
    \fontsize{6}{7}\selectfont
    \caption{Ablation of multi-source alignment objectives. Linear probe AUROC by diagnostic group. Default setting for \methodname{} is \underline{underlined} and the best per column is highlighted in \textbf{bold}.}
    \label{tab:ms_objective_ablation}
    \begin{tabular}{lrrrrrrrrrr}
    \toprule
    & \textbf{All} & \textbf{ARR} & \textbf{CD} & \textbf{ECG-HYP} & \textbf{MI} & \textbf{ISC/STTC} & \textbf{SHD} & \textbf{VC} & \textbf{MIMIC-EST} & \textbf{MIMIC-EXP} \\
    \midrule
    \underline{Stochastic pairing} & \textbf{90.6} & 96.9 & \textbf{95.8} & 93.9 & \textbf{94.4} & \textbf{92.9} & 83.4 & 91.7 & \textbf{88.2} & \textbf{78.8} \\
    Average embeddings & 90.4 & 97.2 & 95.5 & \textbf{94.6} & \textbf{94.4} & 92.6 & 83.3 & 91.5 & 88.0 & 78.7 \\
    Report-specific batch & 90.3 & 96.7 & 95.5 & 94.5 & 93.7 & 92.5 & \textbf{83.9} & \textbf{91.8} & 88.0 & 78.5 \\
    Report-specific projectors & 90.3 & \textbf{97.3} & 95.3 & 93.8 & 94.0 & 92.2 & 83.7 & 91.4 & \textbf{88.2} & 78.2 \\
    \bottomrule
\end{tabular}

\end{table}

\subsection{Comparison of ECG machine report (ECG-MR) performance to external baselines pre-trained with same reports} \label{app:ecg_mr_versus_baselines}

\begin{figure}[h]
    \centering
    \includegraphics[width=\textwidth]{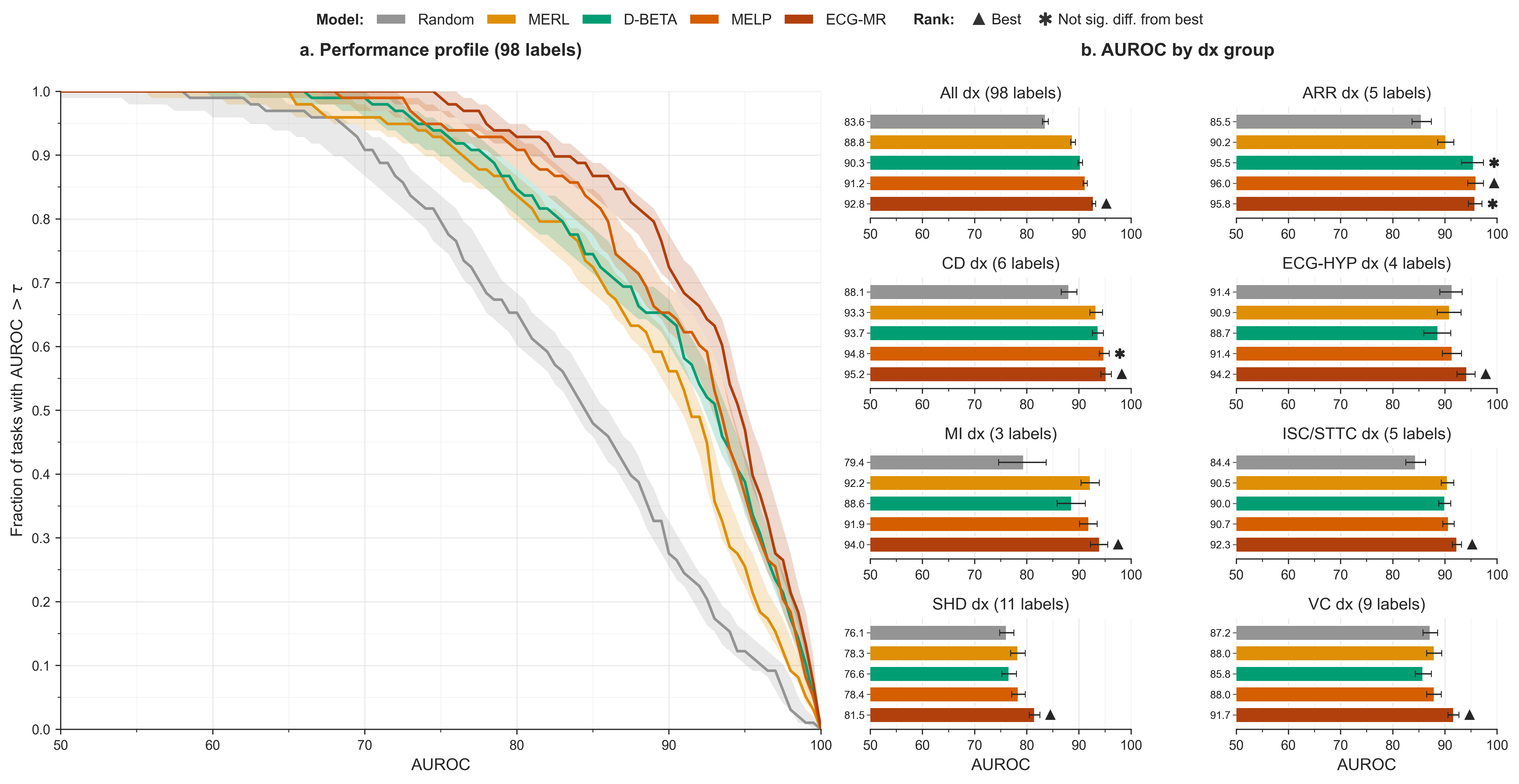}
    \caption{Comparison of a model pre-trained with \methodname{}'s recipe and architecture except that it is only aligned to the ECG machine reports (ECG-MR) to other contemporary ECG foundation models pre-trained with the same MIMIC-IV-ECG machine reports on 12-lead ECG diagnostic tasks using linear probe evaluations. Significance testing is performed using a paired patient-cluster bootstrap procedure (see Methods). 95\% bootstrap confidence intervals are shown. AUROC: area under the receiver operating characteristic curve, ARR: arrhythmia, CD: conduction disturbances, ECG-HYP: hypertrophies detected from ECG, MI: myocardial infarction, ISC/STTC: ischemia and ST/T changes, SHD: structural heart disease, VC: diagnoses that rely on voltage criteria. Conditions by diagnosis group are in \cref{tab:downstream_diagnosis_groups}.}
    \label{fig:perf_profile_ecg_mr_external}
\end{figure}

\cref{fig:perf_profile_ecg_mr_external} compares a model trained using \methodname{}'s architecture and recipe except that it is only aligned to the ECG machine interpretation reports in MIMIC-IV-ECG \cite{gow2023mimic}. These are the same notes that the baselines displayed in \cref{fig:perf_profile_ecg_mr_external} use for pre-training.

\subsection{Ablation of ECG machine report (ECG-MR) pre-training methods} \label{app:ecg_mr_ablation}

\cref{tab:ecg_mr_ablation} shows performance differences when the pre-training recipe is adjusted using only the ECG machine interpretation reports (ECG-MR). The learned representations are strongest when a frozen text encoder is used and the unsupervised effective rank (RankMe) metric is used to select a checkpoint from pre-training. Performance degrades noticeably when using online text encoders, especially when a zero-shot (ZS) model selection criterion is used.\footnote{This is computed on the validation data of some of the \emph{out-of-distribution} downstream datasets as in MERL \cite{liu2024zero} and MELP \cite{wang2025from}. We include it here only to show how it contributes to degraded performance. As mentioned, we also do \textbf{not} advocate for this metric on the basis that we believe the pre-training recipe should be agnostic to out-of-distribution data for a less biased evaluation.} 

\begin{table}[h]
    \centering
    \setlength\tabcolsep{1.5pt}
    \fontsize{5}{6}\selectfont
    \caption{Ablation of pre-training methods when aligning to ECG-MR reports only. Linear probe AUROC by diagnostic group. The frozen-text, ER-checkpoint row is linear-probe AUROC. Other rows are $\Delta$AUROC versus that setting. $\uparrow$ and $\downarrow$ mark the sign of the difference.}
    \label{tab:ecg_mr_ablation}
    \begin{tabular}{lllrrrrrrrrrr}
    \toprule
    \textbf{Report} & \textbf{Text encoder} & \textbf{Checkpoint} & \textbf{All} & \textbf{ARR} & \textbf{CD} & \textbf{ECG-HYP} & \textbf{MI} & \textbf{ISC/STTC} & \textbf{SHD} & \textbf{VC} & \textbf{MIMIC-EST} & \textbf{MIMIC-EXP} \\
    \midrule
    ECG-MR & \underline{Frozen} & \underline{ER} & 89.5 & 95.8 & 95.2 & 94.2 & 94.0 & 92.3 & 81.5 & 91.7 & 86.4 & 76.5 \\ \midrule
    ECG-MR & Frozen & ZS & $\downarrow$0.1 & $\uparrow$0.1 & $\downarrow$0.1 & $\uparrow$0.2 & $\downarrow$0.6 & $\uparrow$0.1 & $\downarrow$0.4 & $\uparrow$0.2 & 0.0 & $\downarrow$0.1 \\
    ECG-MR & Online & ER & $\downarrow$3.9 & $\downarrow$5.8 & $\downarrow$1.4 & $\downarrow$2.6 & $\downarrow$4.3 & $\downarrow$3.3 & $\downarrow$2.5 & $\downarrow$3.3 & $\downarrow$1.9 & $\downarrow$2.2 \\
    ECG-MR & Online & ZS & $\downarrow$24.7 & $\downarrow$26.1 & $\downarrow$24.5 & $\downarrow$20.6 & $\downarrow$26.1 & $\downarrow$24.3 & $\downarrow$21.8 & $\downarrow$24.1 & $\downarrow$20.2 & $\downarrow$15.5 \\
    ECG-MR & Partially online & ER & $\downarrow$0.1 & $\uparrow$0.7 & $\downarrow$0.1 & $\uparrow$0.4 & $\downarrow$0.1 & $\downarrow$0.3 & $\downarrow$0.2 & $\uparrow$0.2 & $\downarrow$0.4 & $\downarrow$0.1 \\
    ECG-MR & Partially online & ZS & $\downarrow$1.5 & $\downarrow$1.4 & $\downarrow$0.4 & $\downarrow$1.6 & $\downarrow$2.3 & $\downarrow$1.0 & $\downarrow$1.3 & $\downarrow$1.0 & $\downarrow$1.1 & $\downarrow$1.6 \\
    \bottomrule
\end{tabular}

    \footnotetext{AUROC: area under the receiver operating characteristic curve. \underline{Underlined} text encoder and checkpoint are the default setting. Frozen uses offline MedGemma-27B embeddings; online jointly updates a MedCPT text encoder (partially online freezes the first 6 layers) \cite{jin2023medcpt}. ER is the RankMe / effective-rank pre-training checkpoint; ZS is the checkpoint selected by out-of-distribution zero-shot AUROC as used by MERL \cite{liu2024zero} and MELP \cite{wang2025from}.}
\end{table}

\FloatBarrier \clearpage

\section{MIMIC-IV clinical reports and waveform preprocessing} \label{app:reports_info}

\subsection{MIMIC-IV-ECG clinical reports}

\cref{tab:mimic_reports} summarizes the different medical reports available in the MIMIC-IV-ECG and MIMIC-IV-Note datasets \cite{gow2023mimic,johnson2023mimic}. It also provides details on how we pre-processed and joined them to the MIMIC-IV-ECG waveform instances.

\begin{table}[h]
  \centering
    \setlength\tabcolsep{1pt}
    \fontsize{4.5}{5.5}\selectfont
    \caption{Summary of clinical reports in the MIMIC-IV-ECG and MIMIC-IV-Note datasets \cite{gow2023mimic,johnson2023mimic}. We join these reports to the 800,035 available ECG recordings (720,102 in the training set). Multiple ECGs from the same visit may be joined to the same ECHO, Chest X-ray, Discharge Summary, Chest CT or Chest CTA report. The \textcolor{red}{$\bullet$} symbol denotes that preliminary experiments looking at performance on the validation set (i.e., not test set) suggested that these reports did not provide useful supervision so we excluded them from further analysis (cf. \cref{app:single_source_all}).}
    \label{tab:mimic_reports}
    \begin{tabular}{
    >{\raggedright\arraybackslash}p{0.1\textwidth}
    >{\raggedright\arraybackslash}p{0.25\textwidth}
    >{\raggedright\arraybackslash}p{0.2\textwidth}
    >{\raggedright\arraybackslash}p{0.15\textwidth}
    >{\centering\arraybackslash}p{0.085\textwidth}
    >{\centering\arraybackslash}p{0.085\textwidth}
    >{\centering\arraybackslash}p{0.085\textwidth}
    >{\centering\arraybackslash}p{0.055\textwidth}
    >{\centering\arraybackslash}p{0.055\textwidth}
}
    \toprule
    & & & & \multicolumn{2}{c}{\textbf{\# Total Notes (Patients)}} & & \multicolumn{2}{c}{\textbf{Length (tokens)}} \\
    \cmidrule(lr){5-6} \cmidrule(lr){8-9}
    \multicolumn{1}{>{\raggedright\arraybackslash}b{0.1\textwidth}}{\textbf{Report}} &
    \multicolumn{1}{>{\raggedright\arraybackslash}b{0.25\textwidth}}{\textbf{Description}} &
    \multicolumn{1}{>{\raggedright\arraybackslash}b{0.2\textwidth}}{\textbf{Pre-processing / Cleaning}} &
    \multicolumn{1}{>{\raggedright\arraybackslash}b{0.15\textwidth}}{\textbf{How to join to ECG}} &
    \multicolumn{1}{>{\centering\arraybackslash}b{0.085\textwidth}}{\textbf{Raw}} &
    \multicolumn{1}{>{\centering\arraybackslash}b{0.085\textwidth}}{\textbf{Clean}} &
    \multicolumn{1}{>{\centering\arraybackslash}b{0.085\textwidth}}{\textbf{\# Train ECGs w/Notes (Patients)}} &
    \multicolumn{1}{>{\centering\arraybackslash}b{0.085\textwidth}}{\textbf{Median}} &
    \multicolumn{1}{>{\centering\arraybackslash}b{0.085\textwidth}}{\textbf{Max}}
    \\ \midrule

    ECG Machine Report (ECG-MR) &
    Reports and summary measures generated by the ECG machine &
    Concatenate all statements together; remove one empty report &
    By \verb|study_id| &
    800,035 (161,352) &
    800,034 (161,352) &
    720,101 (144,545) &
    17 &
    125
    \\ \midrule

    ECG Cardiologist Report (ECG-CR) &
    The ECG is read by a cardiologist who then writes a report. They may reference the machine report or ignore it. &
    (1) Remove clinical indications, (2) Exclude reports with no informative text (e.g., ``ECG interpreted by ordering physician, please see corresponding note") &
    Joined to \verb|waveform_note_links| by \verb|study_id| then to \verb|mimic_iv_note_ecg| by \verb|note_id| &
    623,566 (105,365) &
    490,780 (102,886) &
    441,639 (92,377) &
    33 &
    256
    \\ \midrule

    ECHO Cardiologist Report (ECHO-CR) &
    Ultrasound images of the heart. A cardiologist reviews ECHO images and writes a report. &
    None &
    ECG waveform is joined to the closest ECHO report in a 7-day window &
    81,821 (34,147) &
    81,821 (34,147) &
    129,934 (26,944) &
    442 &
    1,128
    \\ \midrule

    Chest X-ray Radiologist Report (ChXR-RR) &
    X-ray of the chest. Can show cardiomegaly, pulmonary edema, vascular congestion, and other secondary signs of cardiac dysfunction. Radiologist reviews and writes a report. &
    None &
    ECG waveform is joined to the closest ChXR report in a 7-day window &
    715,465 (119,933) &
    715,465 (119,933) &
    415,589 (97,686) &
    108 &
    761
    \\ \midrule

    Discharge Summary Physician Report (DS) &
    Report summarizing a patient's hospitalization including the reason for admission, past medical history, their hospital course (including summaries of various scans and blood/microbiology tests), and any relevant discharge instructions &
    None &
    ECG joined to discharge summary for the corresponding hospital stay by \verb|hosp_hadm_id| or \verb|ed_hadm_id| &
    331,793 (111,647) &
    331,793 (111,647) &
    326,779 (78,754) &
    2,464 &
    15,039
    \\ \midrule \midrule

    Chest CT Radiologist Report (ChCT-RR) \textcolor{red}{$\bullet$} &
    Computed tomography (CT) imaging of the chest. Can show cardiomegaly, pulmonary edema, vascular congestion, and other secondary signs of cardiac dysfunction. Radiologist reviews and writes a report. &
    None &
    ECG waveform is joined to the closest ChCT report in a 7-day window &
    92,513 (33,447) &
    92,513 (33,447) &
    51,994 (18,507) &
    484 &
    2,199
    \\ \midrule

    Chest CTA Radiologist Report (ChCTA-RR) \textcolor{red}{$\bullet$} &
    Computed tomography angiography (CTA) targeted to the pulmonary arteries and thoracic aorta that evaluates acute cardiopulmonary conditions like pulmonary embolism or aortic dissection and can show secondary cardiac findings. Radiologist reviews and writes a report. &
    None &
    ECG waveform is joined to the closest ChCTA report in a 7-day window &
    33,308 (22,125) &
    33,308 (22,125) &
    49,517 (17,005) &
    469 &
    2,214
    \\

\bottomrule
\end{tabular}

\end{table}

\subsection{Single source performance of all report types} \label{app:single_source_all}

\cref{tab:linear_probe_old_bms_all_sources_val_error} shows the validation set (i.e., not test set) AUROC of all individual clinical note sources. We found that generally the chest CT (ChCT-RR) and chest CTA (ChCTA-RR) reports did not provide good supervision in the single source contrastive learning models. We thus excluded them. Their poor performance might be explained by low mutual information with paired ECGs (given these reports are more focused on pulmonary findings), but also potentially due to the small training set size after joining to an ECG (see \cref{tab:mimic_reports}). 

\begin{table*}[h]
    \centering
    \setlength\tabcolsep{1.25pt}
    \fontsize{6}{6}\selectfont
    \caption{Linear probe AUROC \textbf{on the validation set} of existing out-of-distribution (OOD) benchmarks. The results are shown for different percentages of available training labels (1\%, 10\%, 100\%). \textcolor{red}{$\bullet$} denotes exclusion from further analysis. Note: these experiments were preliminary with an earlier version of the model, so deviations with the other results in this manuscript are expected.}
    \label{tab:linear_probe_old_bms_all_sources_val_error}
    \newlength{\valcolwidth}
\setlength{\valcolwidth}{20pt}

\newcommand{\srcgrp}[1]{%
    \multicolumn{3}{c}{\makebox[\dimexpr3\valcolwidth+4\tabcolsep\relax]{#1}}%
}
\resizebox{\textwidth}{!}{%
\begin{tabular}{ll *{3}{w{r}{\valcolwidth}}|*{3}{w{r}{\valcolwidth}}|*{3}{w{r}{\valcolwidth}}|*{3}{w{r}{\valcolwidth}}|*{3}{w{r}{\valcolwidth}}|*{3}{w{r}{\valcolwidth}}}
    \toprule
    \multirow{2}{*}{\textbf{Methods}} & & \srcgrp{\textbf{PTBXL-Super\textsuperscript{$\dagger$}}} & \srcgrp{\textbf{PTBXL-Sub\textsuperscript{$\dagger$}}} & \srcgrp{\textbf{PTBXL-Form\textsuperscript{$\dagger$}}} & \srcgrp{\textbf{PTBXL-Rhythm\textsuperscript{$\dagger$}}} & \srcgrp{\textbf{CPSC2018\textsuperscript{$\dagger$}}} & \srcgrp{\textbf{CSN\textsuperscript{$\dagger$}}} \\
    \cmidrule{3-5} \cmidrule(lr){6-8} \cmidrule(lr){9-11} \cmidrule(lr){12-14} \cmidrule(lr){15-17} \cmidrule(lr){18-20}
    & & \textbf{1\%} & \textbf{10\%} & \textbf{100\%} & \textbf{1\%} & \textbf{10\%} & \textbf{100\%} & \textbf{1\%} & \textbf{10\%} & \textbf{100\%} & \textbf{1\%} & \textbf{10\%} & \textbf{100\%} & \textbf{1\%} & \textbf{10\%} & \textbf{100\%} & \textbf{1\%} & \textbf{10\%} & \textbf{100\%} \\
    \midrule
    ECG-MR && 88.7 & 91.5 & 92.3 & 77.0 & 87.4 & 88.9 & 64.9 & 79.1 & 86.7 & 81.5 & 96.3 & 97.3 & 83.9 & 90.8 & 92.9 & 81.5 & 89.8 & 94.5 \\
    ECG-CR && 88.9 & 92.0 & 93.1 & 80.3 & 87.8 & 90.3 & 67.5 & 80.4 & 88.3 & 91.2 & 96.4 & 98.3 & 86.9 & 92.3 & 94.0 & 79.9 & 91.8 & 97.0 \\
    ECHO-CR && 87.3 & 90.0 & 91.7 & 79.2 & 85.6 & 88.6 & 59.0 & 71.6 & 79.1 & 78.9 & 86.2 & 93.1 & 78.3 & 88.6 & 91.7 & 74.0 & 86.7 & 93.2 \\
    ChXR-RR && 87.1 & 89.7 & 91.2 & 77.8 & 84.2 & 87.9 & 62.7 & 74.0 & 82.2 & 77.0 & 87.8 & 94.1 & 75.7 & 88.4 & 92.3 & 74.2 & 86.3 & 92.0 \\
    DS && 86.5 & 89.9 & 91.6 & 78.3 & 85.7 & 89.1 & 62.9 & 75.9 & 81.8 & 79.8 & 91.5 & 95.7 & 76.7 & 88.8 & 93.3 & 74.3 & 86.4 & 93.5 \\ \midrule
    ChCT-RR \textcolor{red}{$\bullet$} && 85.8 & 88.7 & 90.2 & 77.6 & 84.1 & 86.3 & 56.9 & 71.0 & 78.0 & 80.5 & 87.2 & 92.0 & 71.4 & 84.9 & 88.0 & 70.9 & 83.8 & 87.3 \\
    ChCTA-RR \textcolor{red}{$\bullet$} && 82.7 & 85.8 & 87.1 & 72.7 & 79.1 & 81.0 & 56.3 & 66.0 & 73.7 & 75.5 & 81.3 & 87.4 & 67.2 & 79.3 & 82.2 & 66.7 & 78.2 & 82.5 \\
    \bottomrule
\end{tabular}%
}%

\end{table*}

\subsection{MIMIC-IV-ECG and downstream ECG waveform pre-processing}
\label{app:ecg_preprocessing}

We deliberately apply minimal pre-processing to the ECG waveforms for \methodname{} pre-training. Each MIMIC-IV-ECG record is 12 leads, 10\,s at 500\,Hz, in physical units (mV), and the only cleaning step is to replace missing samples with zero. 

The downstream benchmark datasets are stored in heterogeneous formats, so we harmonize them to match the pre-training data: lead order is remapped to the MIMIC-IV canonical order, each recording is brought to 10\,s at 500\,Hz, and amplitudes are converted to millivolts. As with the pre-training data, we apply no filtering or normalization for \methodname{}. For the external baseline methods, we apply the normalization and filtering approach they apply in pre-training so that evaluation is fair. For the reduced-lead evaluations, the retained leads are left unchanged and the absent leads are zero-filled, so the encoder receives its input in the same form as under random lead masking during pre-training.

\FloatBarrier \clearpage

\section{Further details about \methodname{} design and implementation} \label{app:method_details}

\subsection{ECG waveform encoder architecture} \label{app:architecture_details}

\subsubsection{Lead-level tokenizer} 

Given a 12-lead ECG recording $s \in \mathbb{R}^{\textrm{L}\times\textrm{T}}$ with $\textrm{L}$ leads and $\textrm{T}$ samples, each lead is tokenized independently using a stack of four convolutional blocks whose weights are shared across leads. Each block applies a 1-dimensional convolution (kernel $k\!=\!3$, stride 1) followed by batch normalization, a GELU nonlinearity, and an overlapping max-pooling layer that downsamples time by a factor of 2, 5, 5 and 5 in successive blocks. Channel width grows across blocks (32, 64, 256 and 512) as temporal resolution falls. The tokenizer returns $\textrm{L}\!\times\!\textrm{M}$ tokens of dimension $D\!=\!512$, where $\textrm{M}\!=\!\textrm{T}/P$ and each token summarizes a patch of $P=250$ samples (0.5\,s at 500\,Hz); neighboring tokens' receptive fields overlap by roughly one patch. The convolutional weights are shared across leads.

\subsubsection{Transformer layers}

The $\textrm{L}\!\times\!\textrm{M}$ lead-level tokens are flattened into a single sequence of $\textrm{LM}$ tokens --- 240 for a 10\,s 12-lead recording --- and passed through 12 transformer blocks of width $D=512$, with 8 attention heads of dimension 64 and a feed-forward dimension of 2048. Attention is over the joint lead-time sequence, so any token can attend both across time within its own lead and across leads at any time offset; this is the first point in the encoder at which information crosses leads. Temporal order is supplied by a fixed 1-dimensional sinusoidal positional encoding that is shared across leads and generated at the observed token count, so recordings of different durations require no interpolation or re-learning of the encoding. Lead identity is supplied by a learned embedding table holding one $D$-dimensional vector per lead, which is added to every token of that lead, similar to \citet{na2024guiding}. The two are additive, so each token carries both its time position and its lead identity. 

Layer normalization \cite{layernorm2016} is applied in the transformer block before the attention and feed-forward sublayers. This is compatible with the amplitude-preservation philosophy: by this stage the tokenizer has already mapped absolute voltage into a $D$-dimensional token. The final ECG representation is obtained by global average pooling over the $\textrm{LM}$ output tokens, both during pre-training and for the linear probe evaluations.

\subsection{Contrastive multi-source alignment objective}
\label{app:ms_objective}

\textbf{Projector heads.} Both projector heads map into a shared 256-dimensional alignment space and are reused across report types. The ECG projector $\pi_{\phi}^{\mathrm{wf}}$ is a two-layer MLP with hidden dimension of 512 and output dimension of 256. The text projector $\pi_{\psi}^{\mathrm{text}}$ is a single linear layer from the 5,376-dimensional frozen MedGemma-27B embedding to the same 256-dimensional space, with weights initialized from an orthogonal matrix. Projector heads are discarded after pre-training: downstream linear probing operates on the backbone embeddings $h_i^{\mathrm{wf}}$. Projector head embeddings are $L_2$-normalized before computing the loss.

\subsection{MedGemma-27B embedding extraction} \label{app:medgemma}

To extract report text embeddings, we use \textit{vLLM} to wrap the MedGemma-27B model and extract embeddings efficiently for each set of reports \citep{kwon2023efficient,sellergren2025medgemma}. To get a report-level embedding, we mean pool report tokens. The raw text embedding dimension is 5,376. We used version \texttt{google/medgemma-27b-text-it} from HuggingFace -- \url{https://huggingface.co/google/medgemma-27b-text-it}.

\subsection{Pre-training recipe} \label{app:pretraining_settings}

\cref{tab:exp_settings_pretrain} summarizes the pre-training configuration. A single NVIDIA H200 GPU was used for pre-training.

\begin{table}[h] 
\centering
    \caption{Pre-training settings for \methodname{} and single-source models.}
    \label{tab:exp_settings_pretrain} 
    \begin{tabular}{ll} 
        \toprule
        \textbf{Config} & \textbf{Value} \\
        \midrule
            Optimizer & AdamW \\  
            Batch size & 512 \\
            Learning rate & 2e-4 \\
            Weight decay & 0.2 \\
            Optimizer momentum & $\beta_1, \beta_2{=}0.9, 0.999$ \\  
            Learning rate schedule & Cosine annealing w/linear warm-up \\ 
            Linear warm-up (start lr) & 5\% of steps ($2\times10^{-7}$) \\
            Final / cosine min LR & 1e-5 \\
            Training steps & 75,000 \\
            Checkpoint selection metric & RankMe \cite{pmlr-v202-garrido23a} (every 5,000 steps) \\
            Precision & bf16-mixed \\
            Seed & 42 \\
            Temperature init (floor) & 0.07 (0.01) \\
            Augmentations & RLM~\cite{oh2022lead} and time-shift \\
        \bottomrule
    \end{tabular}
\end{table}

\FloatBarrier \clearpage

\section{Extended evaluation of ECG foundation models} \label{app:extended_evaluation}

\subsection{Downstream evaluation settings} \label{app:downstream_evaluation_protocol}

\subsubsection{Linear probing} 

\cref{tab:exp_settings_linear_probe} summarizes the training configuration for downstream linear-probing.

\begin{table}[h] 
    \centering
    \caption{Linear probe settings for all ECG foundation models.}
    \label{tab:exp_settings_linear_probe} 
    \begin{tabular}{ll} 
        \toprule
        \textbf{Config} & \textbf{Value} \\
        \midrule
            Optimizer & AdamW \\
            Batch size EchoNext-Mini & 1024 \\
            Batch size MIMIC-IV-ECG-Ext-ICD & 1024 \\
            Batch size PTBXL / CPSC2018 / CSN 1--10\% & 16 \\
            Batch size PTBXL / CPSC2018 / CSN 100\% & 128 \\
            Learning rate & 0.001 \\
            Weight decay & 1e-4 \\
            Optimizer momentum & $\beta_1, \beta_2{=}0.9, 0.999$ \\
            Learning rate schedule & Cosine annealing w/linear warm-up \\
            Linear warm-up (start lr) & 5 steps ($1\times10^{-6}$) \\
            Final / cosine min LR & 1e-5 \\
            Training epochs & 100 \\
            Early stopping & val.~AUROC (off for PTBXL/CPSC2018/CSN at 1--10\%) \\
            Augmentations & None \\
        \bottomrule
    \end{tabular}
\end{table}

\subsubsection{Random encoder baseline} 
In line with \citet{berger2026position}, we include a randomly initialized encoder as a linear-probe baseline. They recommend this comparison because randomly initialized ECG encoders are often competitive with pre-trained models, so a foundation model should outperform it before its representations are considered useful. The random encoder baseline uses the same ECG encoder architecture as \methodname{}. In the random encoder, each feature is standardized to zero mean and unit variance with a non-affine batch normalization layer before being passed to the linear classifier.

\subsection{More details on statistical testing}
\label{app:statistical_testing}

Resampling a rare label can produce a draw in which one class is absent, leaving the AUROC undefined. We therefore fix the set of labels entering any aggregate before resampling: a label is retained only if it has at least 10 positives \emph{and} 10 negatives in the full test set. Retained labels can still occasionally have an undefined AUROC in a resample because the patient-level bootstrap does not stratify on prevalence. We therefore exclude any resample in which at least one retained label is undefined, and use this same effective replicate set for all aggregate confidence intervals and tests (this pattern is shared across models, so pairing is preserved). This retains $N^{\text{eff}}_{\text{boot}} = 1999$ of the $2000$ resamples (where the single loss is due to a low-prevalence CSN label), and ensures each replicate's macro-average is taken over the identical label set. 

\cref{tab:main_benchmark} is a deliberate exception, because we want the macro-averaged AUROC point estimates there to be consistent with those reported in the prior literature. We therefore average over every label of each dataset, including the low-prevalence ones. We can still formally test differences between methods with the same paired bootstrap: within each replicate we difference the two models' scores label-wise and average over the labels that are non-degenerate in that replicate. A degenerate label drops out of both models simultaneously. This lets us report the de facto benchmark view on its usual terms while adding the formal comparisons that are absent from the literature.

\subsection{More details on performance profiles}
\label{app:perf_profiles}

To add confidence intervals to the performance profiles we compute the estimator for each bootstrap resample $b$ as
\begin{equation}
    \hat{F}^{(b)}(\tau) = \frac{1}{M}\sum_{m=1}^{M}\mathbf{1}[x^{(b)}_m > \tau].
\end{equation}
The bands in \cref{fig:perf_profile_external,fig:perf_profile_internal,fig:perf_profile_ecg_mr_external} are bounded by the pointwise 2.5th and 97.5th percentiles of $\{\hat{F}^{(b)}(\tau)\}_{b=1}^{N^{\text{eff}}_{\text{boot}}}$ over the effective replicates.

The \emph{single source oracle} method in \cref{fig:perf_profile_internal} takes, on every label, the best score achieved by any single-source model. Writing $x_{r,m}$ for the score of the model aligned to report type $r$ on label $m$, its profile is the same function evaluated on the per-label maxima,
\begin{equation}
    \hat{F}^{\mathrm{oracle}}(\tau) = \frac{1}{M}\sum_{m=1}^{M}\mathbf{1}\Big[\max_{r \in \mathcal{R}}[x_{r,m}] > \tau\Big].
\end{equation}
Naturally, this is not a deployable method as it uses information from the test set to select the best method per label. It is intended as a device to characterize the ability of \methodname{} to simultaneously capture the performance benefits of different report types.

\FloatBarrier \clearpage

\end{appendices} 

\bibliography{sn-bibliography}

\end{document}